%% file: bare_jrnl_compsoc.tex
\documentclass[10pt,journal,compsoc]{IEEEtran}
\usepackage{microtype}
\usepackage{booktabs}
\usepackage{enumitem}
\usepackage{amsmath}
\usepackage{amssymb}
\usepackage{algorithmic}
\usepackage{array}
\usepackage{subfigure}
\usepackage{epsfig}
\usepackage{graphicx}
\usepackage{multirow}
\usepackage{color}

\newcommand{\tinypm}[1]{\tiny{$\pm${#1}}}

\usepackage[framemethod=tikz]{mdframed}
\usepackage{url}

\usepackage{subfigure}
\usepackage[ruled]{algorithm2e}

\ifCLASSOPTIONcompsoc
  \usepackage[nocompress]{cite}
\else
  \usepackage{cite}
\fi
\ifCLASSINFOpdf
\else
\fi
\begin{document}
%
\title{General Greedy De-bias Learning}
%
%
%
%

\author{Xinzhe~Han,~\IEEEmembership{Student~Member,~IEEE,}
        Shuhui~Wang,~\IEEEmembership{Member,~IEEE,}
        Chi~Su,
        Qingming~Huang,~\IEEEmembership{Fellow,~IEEE,}
        and~Qi~Tian,~\IEEEmembership{Fellow,~IEEE}

\IEEEcompsocitemizethanks{
	\IEEEcompsocthanksitem Corresponding author: Shuhui Wang.
	\IEEEcompsocthanksitem  X. Han and Q. Huang are with the School of Computer Science and Technology, University  of  Chinese  Academy  of  Sciences, Beijing 101408, China, and with the Key Laboratory of Intelligent Information Processing, Institute of Computing Technology, Chinese Academy of Sciences, Beijing 100190, China. Q. Huang is also with Peng Cheng Laboratory, Shenzhen 518066, China. \protect\\
	E-mail:  xinzhe.han@vipl.ict.ac.cn, qmhuang@ucas.ac.cn.
	\IEEEcompsocthanksitem  S. Wang is with the Key Laboratory  of Intelligent Information Processing, Institute of Computing Technology, Chinese Academy of Sciences, Beijing 100190, China, and with Peng Cheng Laboratory, Shenzhen 518066, China. \protect\\
	E-mail: wangshuhui@ict.ac.cn.
	\IEEEcompsocthanksitem C. Su is with SmartMore, Beijing, 100085.\protect\\
	Email: chi.su@smartmore.com
	\IEEEcompsocthanksitem  Q. Tian is with Huawei Cloud \& AI, Shenzhen 518129, China. \protect\\
	E-mail: tian.qi1@huawei.com. \protect}
} 
\markboth{Journal of \LaTeX\ Class Files,~Vol.~14, No.~8, August~2015}%
{Shell \MakeLowercase{\textit{et al.}}: Bare Demo of IEEEtran.cls for Computer Society Journals}
%



\IEEEtitleabstractindextext{%
\begin{abstract}
Neural networks often make predictions relying on the spurious correlations from the datasets rather than the intrinsic properties of the task of interest, facing with sharp degradation on out-of-distribution (OOD) test data. 
Existing de-bias learning frameworks try to capture specific dataset bias by annotations but they fail to handle complicated OOD scenarios. Others implicitly identify the dataset bias by special design low capability biased models or losses, but they degrade when the training and testing data are from the same distribution.
In this paper, we propose a General Greedy De-bias learning framework (GGD), which greedily trains the biased models and base model.
The base model is encouraged to focus on examples that are hard to solve with biased models, thus remaining robust against spurious correlations in the test stage. GGD largely improves models' OOD generalization ability on various tasks, but sometimes over-estimates the bias level and degrades on the in-distribution test.
We further re-analyze the ensemble process of GGD and introduce the Curriculum Regularization inspired by curriculum learning, which achieves a good trade-off between in-distribution (ID) and out-of-distribution performance. Extensive experiments on image classification, adversarial question answering, and visual question answering demonstrate the effectiveness of our method. GGD can learn a more robust base model under the settings of both task-specific biased models with prior knowledge and self-ensemble biased model without prior knowledge. Codes are available at \url{https://github.com/GeraldHan/GGD}.
\end{abstract}

}

\maketitle

\IEEEdisplaynontitleabstractindextext

%
\IEEEpeerreviewmaketitle

\input{Intro}

\input{related_work}

\input{method}
\input{experiment}

\ifCLASSOPTIONcompsoc
\section*{Acknowledgments}
\else
\section*{Acknowledgment}
\fi

This work was supported in part by the National Key R\&D Program of China under Grant 2018AAA0102000, in part by National Natural Science Foundation of China: 62022083, 62236008 and 61931008, and in part by the Beijing Nova Program under Grant Z201100006820023.

%

\ifCLASSOPTIONcaptionsoff
  \newpage
\fi



\bibliographystyle{IEEEtran}
\bibliography{IEEEabrv,jrnl_bib}

\begin{IEEEbiography}[{\includegraphics[width=1in,height=1.25in,clip,keepaspectratio]{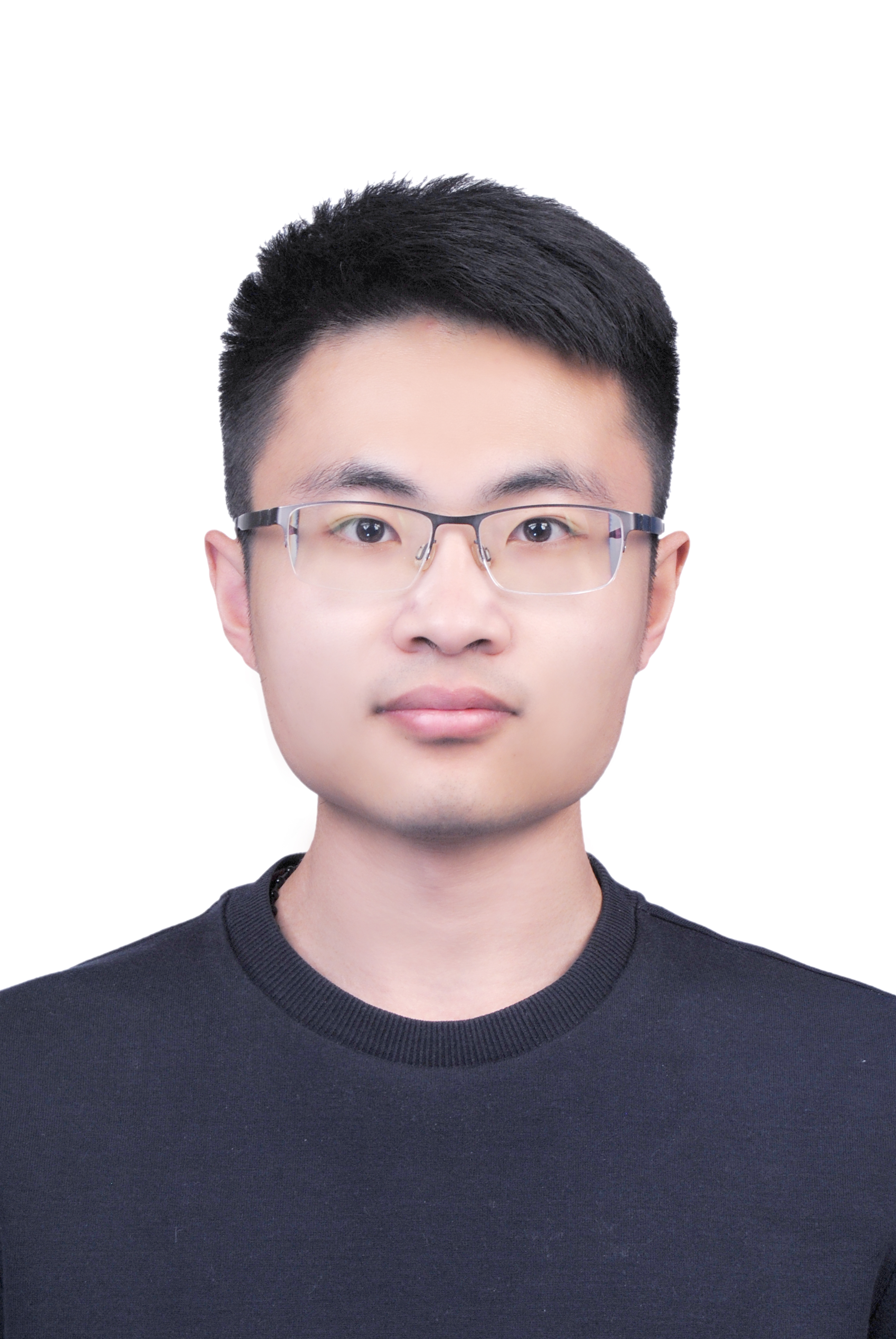}}]{Xinzhe Han} received the B.S. degree from Xidian University in 2017. He is currently pursuing the Ph.D. degree in the School of Computer Science and Technology, University of Chinese Academy of Sciences. His current research interests include visual question answering and trustable machine learning.
\end{IEEEbiography}

\begin{IEEEbiography}[{\includegraphics[width=1in,height=1.25in,clip,keepaspectratio]{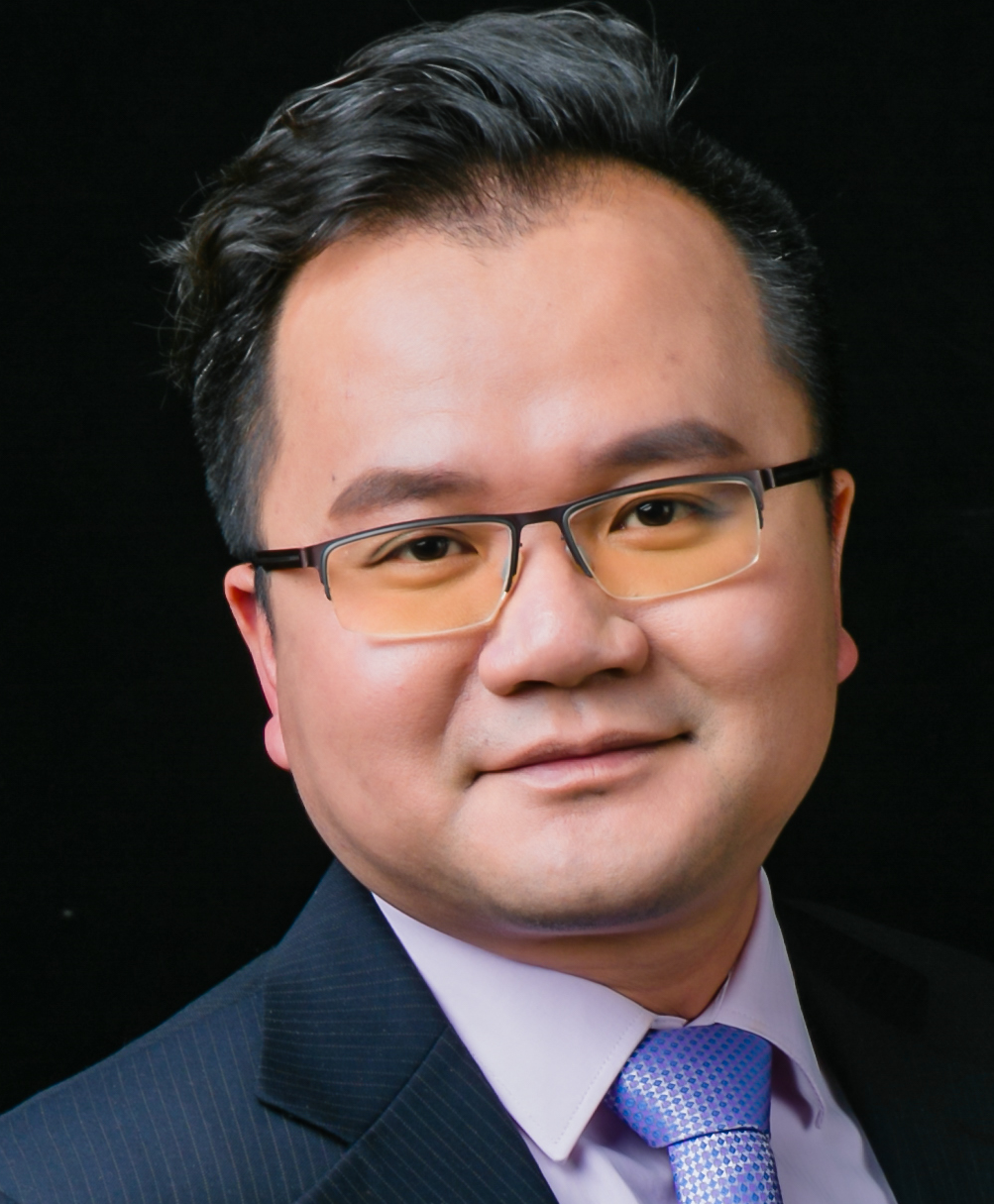}}]{Shuhui Wang} received the B.S. degree in electronics
	engineering from Tsinghua University, Beijing, China, in 2006, and the Ph.D. degree from the Institute of Computing Technology, Chinese Academy
	of Sciences, Beijing, China, in 2012. He is currently a Full Professor with the Key Laboratory of Intelligent Information Processing~(CAS), Institute of Computing Technology, Chinese Academy of Sciences.
	He is also with Pengcheng Laboratory, Shenzhen. His research interests include image/video understanding/retrieval, cross-media analysis and visual-textual knowledge extraction.
\end{IEEEbiography}

\begin{IEEEbiography}[{\includegraphics[width=1in,height=1.25in,clip,keepaspectratio]{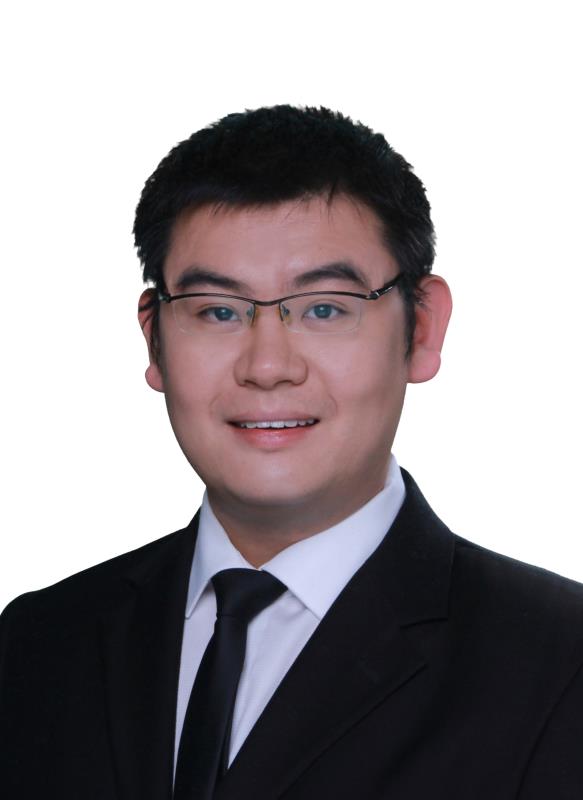}}]{Chi Su} is currently a General Manager of SmartMore, Beijing. He received the PhD degree in the Institute of Digital Media, EECS, Peking University. His research include computer vision and machine learning, with focus on object detection, object tracking, and human identification and recognition.
\end{IEEEbiography}

\begin{IEEEbiography}[{\includegraphics[width=1in,height=1.25in,clip,keepaspectratio]{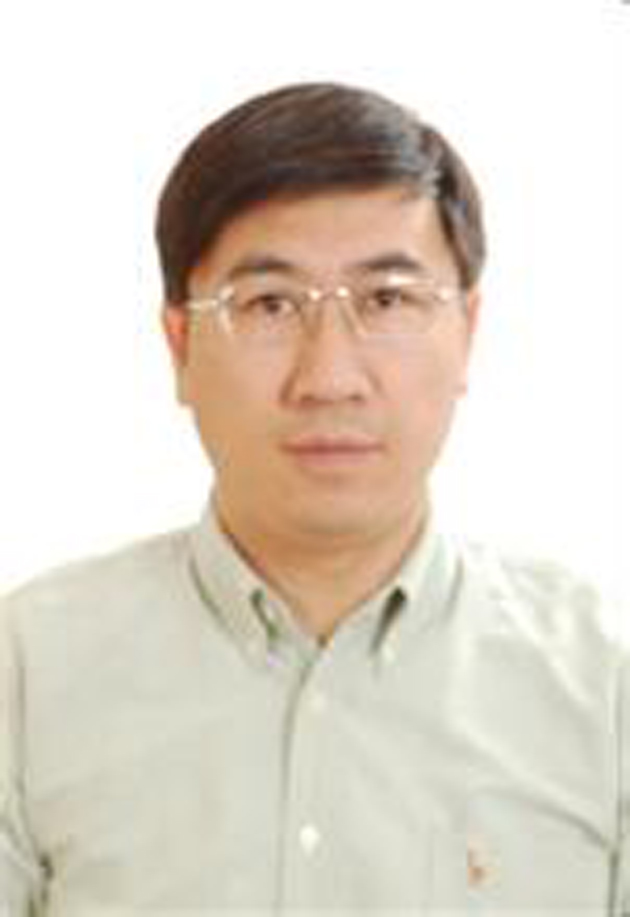}}]{Qingming Huang} 
	received the B.S. degree in computer science and Ph.D. degree in computer engineering from the Harbin Institute of Technology, Harbin, China, in 1988 and 1994, respectively.
	He is currently a Chair Professor with the School of Computer Science and Technology,
	University of Chinese Academy of Sciences. He has published over 500 academic papers in international journals, such as IEEE Transactions on Pattern Analysis and Machine Intelligence,
	IEEE Transactions on Image Processing, IEEE Transactions on Multimedia, IEEE Transactions on Circuits and Systems for Video Technology, and top level international conferences, including
	the ACM Multimedia, ICCV, CVPR, ECCV, VLDB, and IJCAI. He was the Associate Editor of IEEE Transactions on Circuits and Systems for Video Technology and the Associate Editor of Acta Automatica Sinica. His research interests include multimedia computing, image/video processing, pattern recognition, and computer vision.
\end{IEEEbiography}

\begin{IEEEbiography}[{\includegraphics[width=1in,height=1.25in,clip,keepaspectratio]{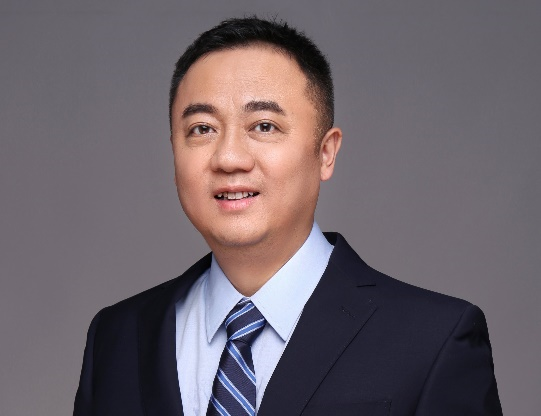}}]{Qi Tian} is currently a Chief Scientist in Artificial Intelligence at Cloud BU, Huawei. From 2018-2020, he was the Chief Scientist in Computer Vision at Huawei Noah's Ark Lab.  He was also a Full Professor in the Department of Computer Science, the University of Texas at San Antonio (UTSA) from 2002 to 2019. During 2008-2009, he took one-year Faculty Leave at Microsoft Research Asia (MSRA). 
Dr. Tian received his Ph.D. in ECE from University of Illinois at Urbana-Champaign (UIUC) and received his B.E. in Electronic Engineering from Tsinghua University and M.S. in ECE from Drexel University, respectively. Dr. Tian's research interests include computer vision, multimedia information retrieval and machine learning and published 600+ refereed journal and conference papers. His Google citation is over 44700+ with H-index 97. He was the co-author of best papers including IEEE ICME 2019, ACM CIKM 2018, ACM ICMR 2015, PCM 2013, MMM 2013, ACM ICIMCS 2012, a Top 10\% Paper Award in MMSP 2011, a Student Contest Paper in ICASSP 2006, and co-author of a Best Paper/Student Paper Candidate in ACM Multimedia 2019, ICME 2015 and PCM 2007.
Dr. Tian research projects are funded by ARO, NSF, DHS, Google, FXPAL, NEC, SALSI, CIAS, Akiira Media Systems, HP, Blippar and UTSA. He received 2017 UTSA President’s Distinguished Award for Research Achievement, 2016 UTSA Innovation Award, 2014 Research Achievement Awards from College of Science, UTSA, 2010 Google Faculty Award, and 2010 ACM Service Award. He is the associate editor of IEEE TMM, IEEE TCSVT, ACM TOMM, MMSJ, and in the Editorial Board of Journal of Multimedia (JMM) and Journal of MVA.  Dr. Tian is the Guest Editor of IEEE TMM, Journal of CVIU, etc. Dr. Tian is a Fellow of IEEE.
\end{IEEEbiography}

\clearpage

\appendices  
\input{appendix}

%
\end{document}

%% file: Intro.tex
\IEEEraisesectionheading{\section{Introduction}\label{sec:introduction}}

%
%
%
%
\IEEEPARstart{D}{eep} learning have been used in a wide range of tasks that involves vision and/or language~\cite{9356220}. 
Most of the current approaches are data-driven and heavily rely on the assumption that the training and testing data are drawn from the same distribution. They are usually susceptible to poor generalization on out-of-distribution or biased settings~\cite{2011unbiased}. 
This limitation partially arises because supervised training only identifies the correlations between given examples and their labels~\cite{2017generalization},  which may reflect the dataset-specific bias rather than intrinsic properties of the task of interests~\cite{2020shortcut,2020overpara}. In general, under the supervised objective function fitting paradigm, if the bias is sufficient to make the model achieve high accuracy, there is less motivation for models to further learn those true instrinsic factors of the task. For example, QA models trained on SQuAD~\cite{2016squad} tend to select the text near question-words as answers regardless of the context~\cite{2017adqa,2019-compositional}, and
VQA models usually leverage superficial correlations between questions and answers without considering the vision information~\cite{2017mfh,2017analysis}. 
When it comes with the more common situation that the distribution of test data deviates from that of training data, models exploiting the biases in training data are prone to show poor generalization and hardly provide proper evidence for their predictions.
 

Being aware of this problem, researchers re-examine many popular datasets, resulting in the discovery of a wide variety of biases on different tasks, such as language bias in VQA~\cite{2018vqacp}, color bias in Biased-MNIST~\cite{2020rebias}, gender/background bias in image classification~\cite{2015CelebA,2021NICO}, and the ubiquitous long-tailed distribution~\cite{2019LDAM,2019class}. 
Built on these findings, explicit de-bias methods~\cite{2019don,2019rubi,2019GDRO,2019lnl,2020rebias,2021end} assume that bias variables are explicitly annotated, then the out-of-distribution performance can be directly improved by preventing the model from using the known biases or bias-related data augmentation~\cite{2020counterfactual,2020counterfactualgs}. 
Although these methods achieve remarkable improvement on typical diagnosing datasets, they can only mitigate one specific bias, which is inconsistent with the real world datasets with compositional biases~\cite{2021investigation}. For example, in VQA, biases may stem from unbalanced answer distribution, spurious language correlations, and object contexts.
Even when all bias variables are identified, the explicit de-bias methods still cannot well handle multiple types of biases.
Some recent works, {\it i.e.}, the implicit methods~\cite{2020lff,2020GS,2020rsc,2021biaswap}, try to discover the compositional biases without explicit task-related prior knowledge.
They are somehow overcomplicated and usually perform worse than explicit methods under well-defined circumstances with known biases.

\begin{figure}[h]
	\centering
	\includegraphics[width=\linewidth]{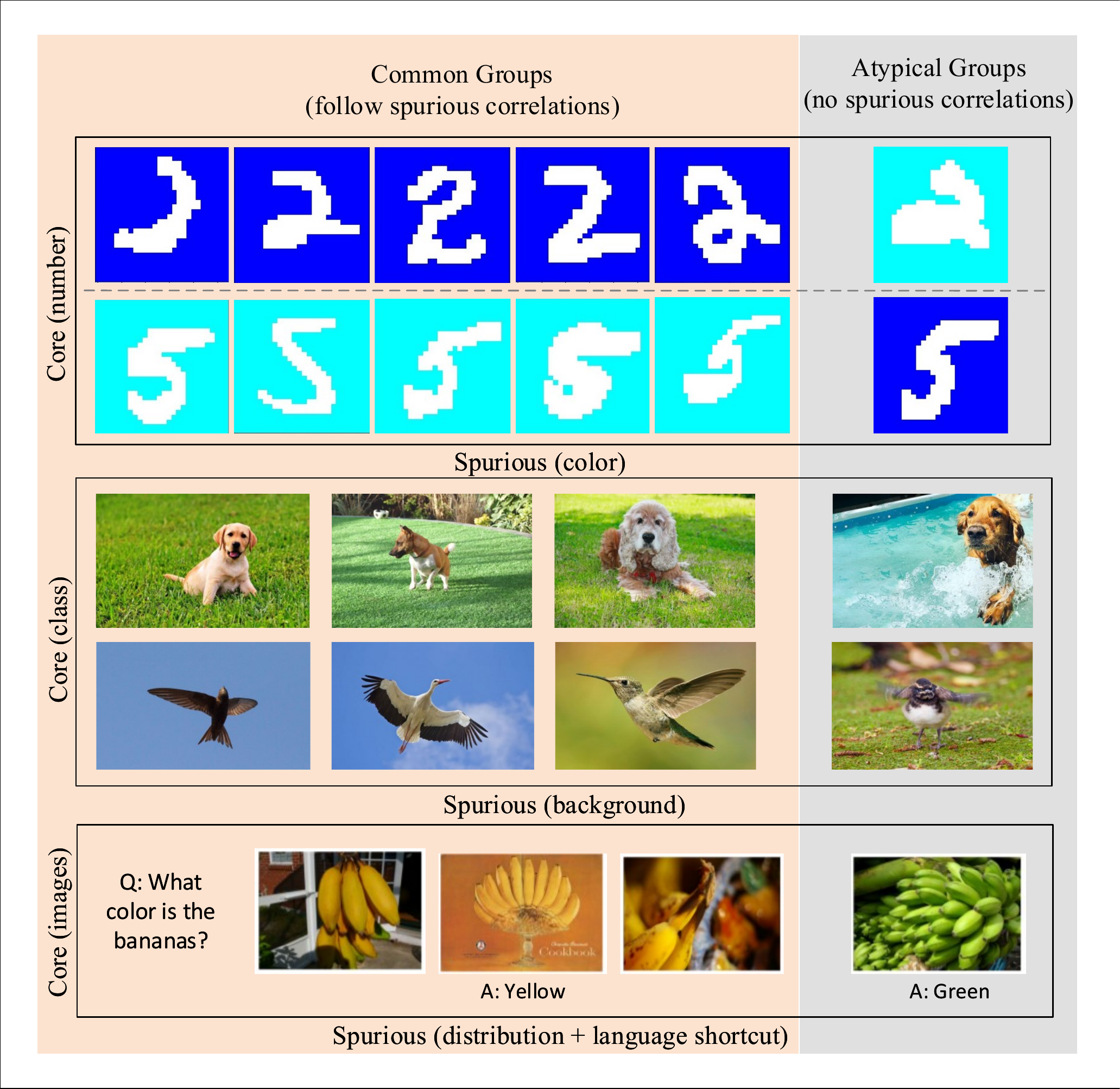}
	\caption{Examples of dataset biases in different tasks. Models tend to capture spurious correlations between the inputs and the labels instead of the task of interest. From top to bottom, we illustrate the color bias in Biased-MNIST, background bias in image classification, and compositional language bias in VQA.}
	\label{sup-corr}
\end{figure}

In fact, the dataset biases can be reduced in a more straightforward manner. As shown in Fig.~\ref{sup-corr}, features learned from biases are thought to be ``spurious" because they can only generalize to the majority groups of samples in the dataset. 
Although the model may incur high training error on the minority groups where the spurious correlation does not hold, the overall loss will still be trapped in a local minimum due to the low average training error dominated by the majority groups. Compared with the core features (\emph{e.g.}, the semantics of objects), it is relatively easier to identify the biases brought by distractive information (\emph{e.g.}, the backgrounds).
If it is possible to know ahead of time that which subsets of instances are irrelevant to spurious features, we can encourage the model to focus on these samples and then reduce the unexpected correlations.
In our preliminary work~\cite{han2021greedy}, we propose a de-bias framework Greedy Gradient Ensemble (GGE) to mitigate the language biases in VQA, achieving great improvement on the biased dataset VQA-CP~\cite{2018vqacp}. 
GGE greedily learns a series of biased models and then ensembles the biased models and the base model like gradient descent in the functional space. The gradient of biased model naturally indicates the difficulty of a sample with certain spurious correlation.




However, without explicit labels for bias variables, even when the prior knowledge for dataset bias is given, disentangling the biased part from the whole feature still remains ill-posed~\cite{2021ebd}. Degradation on in-distribution test is a common problems in exiting de-bias method~\cite{2019rubi,2020counterfactual,2020lff} including GGE.
Greedily emphasizing the bias reduction would also lead to the overreach of the learning objectives.
Ever worse, this bias overestimation brings harm to the model generalizablity under more general cases.
Few works have considered this issue for de-bias learning~\cite{2021introspective, kumar2022calibrated}, relying on extra effort in constructing a biased model separately for ensemble or distillation, but they appear to be less flexible in dealing with complex real applications.
In this paper, we first re-analyse the cause of bias overestimation in GGE. 
We empirical find that if a large amount of data can be correctly predicted via the biased model with high confidence, they will be excluded in the training of base model. 
As a result, the base model may be under-fitted to some labels due to inadequate training data. 
Decomposing the negative gradient of cross-entropy loss, we further find that the cross-entropy between the base prediction and biased prediction measures the difficulty of samples in GGE. 

Base on this finding, we transform the negative gradient supervision to a flexible regularization and formulate a more general framework, {\it i.e.}, General Greedy De-bias~(GGD), to tackle the bias over-estimation problem more appropriately.
Inspired by the curriculum learning~\cite{2009curriculum}, we treat the regularization term as a difficulty metric for the curriculum selection function. In this way, all data can participate in base model training in the early stage and gradually focus on hard examples along with the training procedure. This treatment endows our model with more flexibility, and demonstrates robustness on both out-of-distribution test data and general datasets like ImageNet~\cite{2009imagenet} and CIFAR~\cite{2009cifar}.

In the experiments, we apply GGD to a wider range of uni-modal and multi-modal tasks, including visual classification, linguistic question answering, and visual question answering. Quantitative and qualitative evaluations on all the tasks show that our framework is feasible to general dataset bias on different tasks and gains improvement on both in-distribution (ID) and out-of-distribution (OOD) performance without extra annotations in training and extra computational cost in inference. 

The main contributions of this paper are summarized as:
\begin{itemize}
	\item We present a de-bias framework, General Greedy Debias Learning, which encourages unbiased based model learning by the robust ensemble of biased models.
	GGD is more generally applicable compared to task-related explicit de-bias learning methods while more flexible and effective compared to implicit de-bias methods.
	\item We propose Curriculum Regularization for GGD, which results in a new training scheme GGD$_{cr}$ that can better alleviate the ``bias over-estimation" phenomenon. 
	Compared with previous methods~\cite{xie2020n,kumar2021fine}, GGD$_{cr}$ comes to a better trade-off between in-distribution and out-of-distribution performance without either extra unbiased data in training or model ensemble in inference.
	\item Experiments on image classification, question answering, and visual question answering demonstrate the effectiveness of GGD on different types of biases.  
\end{itemize}

This paper provides a more general debias learning framework compared to our preliminary study~\cite{han2021greedy}. First, the previous work~\cite{han2021greedy} only aims at the VQA task, in which we pay attention to the bias analysis on VQA-CP~\cite{2018vqacp} and new evaluation metric for models' visual grounding ability, while this paper considers the general de-bias learning problem and extend our framework to various datasets and applications. Second, we provide discussions for the ``bias over-estimation" phenomenon in previous GGE~\cite{han2021greedy}. We propose a flexible GGD$_{cr}$ optimization scheme that effectively improves the in-distribution performance on different tasks.
Third, we provide more in-depth analysis for the greedy de-bias strategy, and the differences between GGD and previous GGE are demonstrated from both theories and experiments.
Fourth, we provide more experiments on GGD with known biases and unknown biases, and comparison with the latest de-bias methods with both explicit and implicit bias modelling are also provided. Finally, we apply GGD on three additional tasks, {\it i.e.}, image classification, adversarial question answering, and visual question answering, which are the representative tasks from CV, NLP, and Vision-Language, respectively, demonstrating the circumstances under one single bias, unknown bias, long-tailed bias and multiple biases. In addition to VQA-CP and VQA v2 in \cite{han2021greedy}, we add experiments on more datasets, {\it i.e.}, Biased MNIST~\cite{2020rebias}, SQuAD~\cite{2016squad}, Adversarial SQuAD~\cite{2017ad_squad}, GQA-OOD~\cite{2021gqa-ood}, CIFAR-10 and CIFAR-100~\cite{cui2019class}. Experiments on various tasks and datasets fully demonstrate the general applicability of GGD in debias learning.

%% file: related_work.tex
\section{Related Work}
\subsection{De-biasing from Data Sources}
When collecting real-world datasets, biases in the data are inevitable. Torralba and Efros~\cite{2011unbiased} show how biases affect some commonly used datasets. It draws consideration on the generalization performance and classification capability of the trained deep models. Recent dataset construction protocols have tried to avoid certain kinds of biases. For example, on both CoQA~\cite{2019coqa} and QuAC~\cite{2018quac} for QA task, annotators are prevented from using words that occur in the context passage. For VQA, Zhang {\it et al.}~\cite{2016yin} collect complementary abstract scenes with opposite answers for all binary questions.
Similarly, VQA v2~\cite{2017mfh} is introduced to weaken the language priors in the VQA v1 dataset~\cite{2015vqa} by adding similar images with different answers for each question. 

However, constructing large-scale datasets is costly. It is crucial to develop models that are robust to biases~\cite{2021towards}.
Towards this goal, new diagnosing datasets are established by amplifying some specific biases. For instance, Agrawal {\it et al.}~\cite{2018vqacp} constructed a diagnosing VQA dataset under Changing Prior (VQA-CP), with different answer distributions between the train and test splits. Adversarial SQuAD~\cite{2017ad_squad} is built by adding distracting sentences to the passages in SQuAD~\cite{2016squad}. He {\it et al.} collect NICO~\cite{2021NICO} dataset that consists of images with different backgrounds and gestures. 
All these new datasets can be used to test the models' generalization ability on out-of-distribution scenarios.
 
\subsection{Explicitly De-biasing with Known Bias}
To train a de-biased model, some works utilize an intentionally biased model to de-bias another model.
For VQA, Ramakrishnan {\it et al.}~\cite{2018overcoming} introduce an adversarial regularization to remove the discriminative features related to the answer categories from the questions. RUBi~\cite{2019rubi} and PoE~\cite{2020end} re-weight samples based on the question-only predictions. Kim {\it et al.}~\cite{2019lnl} propose a regularization term based on mutual information between the feature embedding and the bias, to remove the known bias for image classification. 
Similarly, Clark {\it et al.}~\cite{2019don} construct bias-only models for VQA, reading comprehension, and natural language inference (NLI), then reduce them with bias production and entropy maximization. 
Xiong {\it et al.}~\cite{2021ebd} further conduct uncertainty calibration on the bias-only models for a better de-biasing performance. 
It can detect sample outliers and feature noises simultaneously. 
Bahng {\it et al.}~\cite{2020rebias} find that Hilbert-Schmidt Independence Criterion (HSIC)~\cite{2005hsic} can encourage a set of features to be statistically independent. 
They capture local texture bias in image classification and static bias in the video action recognition task using small-capacity models and then train a de-biased representation that is independent of biased representations based on HSIC.

Teney {\it et al.}~\cite{2020counterfactualgs} generate counterfactual samples with specific prior knowledge for different tasks. The vector difference between pairs of counterfactual examples serves to supervise the gradient orientation of the network. 
Liang {\it et al.}~\cite{2021A-INLP} propose A-INLP that dynamically finds bias-sensitive tokens and mitigates social bias in text generation.
Tartaglione {\it et al.}~\cite{2021end} propose a new regularization named EnD, which aims to disentangle the features having the same “bias label”. Sagawa {\it et al.}~\cite{2019GDRO} avoid bias over-fitting by defining prior data sub-groups and controlling their generalization. HEX~\cite{2019hex} pushes the model to learn representations from which the texture representation is not predictable with the reverse gradient method. Gat {\it et al.}~\cite{2020mfe} introduce a regularization by maximizing functional entropies (MFE), which forces the model to use multiple information sources in multi-modal tasks. 
Zhu {\it et al.}~\cite{2021CSAD} explicitly extract target and bias features from the latent space. Then they learn to discover and remove their correlation with the mutual information estimation. 
Hong {\it et al.}~\cite{2021bcbb} leverage the knowledge of bias labels and propose Bias-Contrastive and Bias-Balanced losses based on the contrastive learning.

The above methods only focus on one specific bias but cannot work well on compositional biases. GGD can sequentially mitigate multiple bias variables as long as they can be characterized with prior knowledge, which is much more flexible than explicit de-biasing methods.

\subsection{Implicitly De-biasing without Known Bias}
In real-world scenario, bias presented in the dataset is often hard to characterize and disentangle. 
To address this issue, there have been several recent works to resolve dataset bias without explicit supervision on the biases. 
For linear models, to alleviate the co-linearity among variables, Shen {\it et al.}~\cite{2020stable} propose to learn a set of sample weights that can make the design matrix nearly orthogonal. Kuang {\it et al.}~\cite{kuang2020stable} further propose a re-weighting strategy so that the weighted distribution of treatment and confounder could satisfy the independent condition.

For deep models, most implicit methods assume that easy-to-learn biases can be captured by models with limited capacity and model parameters~\cite{2021lfm}, using a small subset of training instances in a few epochs~\cite{2020nlu}, and a classifier attached to intermediate layers~\cite{2020mce}. 
Apart from limited capacity biased models, Huang {\it et al.}~\cite{2020rsc} iteratively discard the dominant features activated on training data and force the network to activate the remaining features correlated with labels.
Nam {\it et al.}~\cite{2020lff} amplify the biases using generalized cross-entropy (GCE) loss and train a de-biased classifier with resampling based on the biased classifier. 
Still based on GCE, BiaSwap~\cite{2021biaswap} further generates bias-swapped images from bias-contrary images as bias-guided data augmentation.
Zhang {\it et al.}~\cite{2021stablenet} introduce a non-linear feature decorrelation approach based on Random Fourier Features, which can approximate the Hilbert-Schmidt norm in Euclidean space. Spectral Decoupling~\cite{2020GS} decouples the learning dynamics between features. It aims to overcome the issue of gradient starvation, which indicates the tendency to only rely on statistically dominant features. 
Moreover, the setting of implicit de-biasing is similar to the Domain Generalization (DG)~\cite{blanchard2021domain,christiansen2021causal} but has different challenges. In DG, the model is encouraged to generalize to a new domain that is not accessible during training while the de-bias has a small amount of training data that is bias-conflicted. Meanwhile, since there is no clear ``domain discrepancy'' in the biased sets, most existing DG methods do not work on the dataset bias problem.

Implicit methods are much more flexible. However, compared with explicit de-bias methods, totally ignoring prior knowledge limits their capability upper-bound for some tasks. If multiple types of biases are characterized, they cannot fully leverage all the valuable information. In contrast, GGD makes use of task-specific knowledge so that it can mitigate compositional biases. For tasks without prior knowledge of the biases, it can also learn a more robust model with a self-ensemble biased model like implicit de-biasing methods, gaining more flexibility in real world applications.

%% file: method.tex
\section{Proposed Method}
\subsection{Preliminaries}
In this section, we first introduce the notations used in the rest of this paper.
$(X,Y) \in \mathcal{X} \times \mathcal{Y}$ denotes the training set, where $\mathcal{X}$ is the feature space of observations, and $\mathcal{Y}$ is the label space.
Assume $\mathcal{B} = \{B_1, B_2, \dots, B_M\}$ to be a set of task-specific bias features that can be extracted in priority, such as texture features in Biased-MNIST and the language shortcut in VQA. Correspondingly, $h_m(B_m; \phi_m) : B_m\rightarrow \mathcal{Y}$ is a biased model that makes prediction with certain biased feature $B_m$, where $\phi_{m}$ is the parameter set of $h_m(.)$ that maps $B_m$ to the label space $\mathcal{Y}$. Similarly, $f(X;\theta): \mathcal{X} \rightarrow \mathcal{Y}$ denotes the base model, {\it i.e.}, our target model for inference. For supervised learning, the training objective is to minimize the distance between the predictions and the labels $Y$ as
\begin{equation}\label{base}
\min_\theta \mathcal{L}\left(f(X; \theta), Y  \right),
\end{equation}
where the loss function can be various types of supervision loss, such as cross-entropy (CE) loss for single-label classification, binary cross-entropy (BCE) loss for multi-label classification, triplet loss for retrieval, \emph{etc}. Similar to previous works~\cite{2020lff,2020GS,2020rebias,2021end}, considering that the classification (and its variants) is the most common task that seriously suffers from the dataset bias problem, we also take classification tasks as a demonstration in this paper.

\subsection{Greedy Gradient Ensemble}
Given Eq.~\ref{base}, $f(.)$ is chosen to be an over-parametrized DNN, so the model is easy to over-fit the biases in the datasets and suffers from poor generalization ability. We take advantage of the easy-to-overfit property of deep models, and joinly fit the ensemble of bias models $\sum_{m=1}^M h_m(B_m; \phi_m)$ and base model $f(X; \theta)$ to label $Y$
\begin{equation}\label{base1}
\min_{\phi,\theta} \mathcal{L}\left( f(X; \theta) + \sum_{m=1}^{M}h_m(B_m; \phi_m), Y  \right).
\end{equation}
Ideally, we hope the spurious correlations are \emph{only} over-fitted by the bias models, thus the base model $f(.)$ can be learned with a relatively unbiased data distribution. To achieve this goal, GGE adopts a greedy strategy that encourages biased models to have a higher priority to fit the dataset. In practice, $f(.)$ can be ResNet for image classification, UpDn~\cite{2018bottomup} for VQA, {\it etc.}, while $h(.)$ can be low capability model for the texture bias, question-answer classifier for the question shortcut bias, {\it etc.}. 

Viewing from a general ensemble model in the functional space~\cite{2000gradient}, suppose we have $\mathcal{H}_m = \sum_{{m'}=1}^{m}h_{m'}(B_{m'})$ and we wish to find $h_{m+1}(B_{m+1})$ added to $\mathcal{H}_m$ so that the loss $\mathcal{L}\left(\sigma(\mathcal{H}_m + h_{m+1}(B_{m+1})), Y  \right)$ decreases. Theoretically, the desired direction of $h_{m+1}$ should be the negative derivative of $\mathcal{L}$ at $\mathcal{H}_m$, {\it i.e.},
\begin{equation}\label{direction}
-\nabla \mathcal{L}(\mathcal{H}_{m,j}) := \frac{\partial \mathcal{L}\left(\mathcal{H}_m, Y  \right)}{\partial \mathcal{H}_{m,j}}, j \in {1, 2, ..., C}.
\end{equation}
where $\mathcal{H}_{m,j}$ denotes the prediction for the $j$-th class among the overall $C$ classes. For a classification task, we only care about the probability for class $j$: $\sigma(f_j(x)) \in (0,1) $. Therefore, we treat the negative gradients as pseudo labels for classification and optimize the new model $h_{m+1}(B_{m+1})$ with
\begin{equation}\label{gradient}
\mathcal{L} \left(h_{m+1}(B_{m+1};\phi_{m+1}), -\nabla \mathcal{L}(\mathcal{H}_m)\right).
\end{equation}

After integrating all biased models, the expected base model $f$ is optimized with
\begin{equation}\label{fx}
\mathcal{L} \left(f(X;\theta), -\nabla \mathcal{L}(\mathcal{H}_M) \right).
\end{equation}
In the test stage, we only use the base model for prediction. 
In order to make the above paradigm adaptive to mini-Batch Gradient Decent (MBGD), we implement an iterative optimization scheme~\cite{han2021greedy} as shown in Algorithm~\ref{alg:iter}.
Note that our framework learns the base model and biased models jointly, which is different from existing work~\cite{xie2020n,kumar2022calibrated} where the biased model is learned via another independent process or additional annotations.

\begin{algorithm}[t]
	\caption{GGD$_{gs}$}
	\label{alg:iter}
	\KwIn{ Observations $X$, Labels $Y$,\\
		Biased feature Observations $\mathcal{B} = \{B_m\}_{m=1}^{M} $, \\
		Base function $f(.|\theta): X \rightarrow \mathbb{R}^{|Y|}$,\\
		Bias functions $\{h_m(.|{\phi_m}): B_m \rightarrow \mathbb{R}^{|Y|} \}_{m=1}^{M} $ }
	{\bf Initialize: $\mathcal{H}_0 = 0 $ }  \;
	\For{Batch $t= 1 \dots T$}{
		\For{$m = 1 \dots M$}{
			$L_m(\phi_m) \leftarrow  \mathcal{L}' \left(h_{m}(B_{m};\phi_{m}),-\nabla \mathcal{L}(H_{m-1}, Y) \right) $\\
			Update $\phi_m \leftarrow \phi_m - \alpha \nabla_{\phi_m} L_m(\phi_m) $\\
		}
		$L_{M+1}(\theta) \leftarrow  \mathcal{L}' \left(f(X;\theta), -\nabla \mathcal{L}(H_M, Y) \right) $ \\
		Update $\theta \leftarrow \theta - \alpha \nabla_{\theta} L_{M+1}(\theta) $\\
	} 
	\Return $f(X; \theta) $  
\end{algorithm}

\begin{algorithm}[t]
	\caption{GGD$_{cr}$}
	\label{alg:cr}
	\KwIn{ Observations $X$, Labels $Y$,\\
		Biased feature Observations $\mathcal{B} = \{B_m\}_{m=1}^{M} $, \\
		Base function $f(.|\theta): X \rightarrow \mathbb{R}^{|Y|}$,\\
		Bias functions $\{h_m(.|{\phi_m}): B_m \rightarrow \mathbb{R}^{|Y|} \}_{m=1}^{M} $ }
	{\bf Initialize: $\mathcal{H}_0 = 0$ }  \;
	\For{Batch $t= 1 \dots T$}{
		$\lambda_t \leftarrow \sin (\frac{\pi t}{2T})$ \\
		\For{$m = 1 \dots M$}{
			$L_m(\phi_m) \leftarrow  \mathcal{L}' \left(h_{m}(B_{m};\phi_{m}),-\nabla \mathcal{L}(H_{m-1}, Y) \right) $\\
			Update $\phi_m \leftarrow \phi_m - \alpha \nabla_{\phi_m} L_m(\phi_m) $\\
		}
		$\hat{\sigma}(\mathcal{H}_M) \leftarrow  Y \odot \sigma(\mathcal{H}_M))$
		$L_{M+1}(\theta) \leftarrow  \mathcal{L}\left(f(X;\theta), Y  \right) - \lambda_t CE(f(X),  \hat{\sigma}(\mathcal{H}_M))$ \\
		Update $\theta \leftarrow \theta - \alpha \nabla_{\theta} L_{M+1}(\theta) $\\
	} 
	\Return $f(X; \theta) $  
\end{algorithm}

\begin{figure*}[t]
	\begin{center}
		\subfigure[GGD$_{gs}$]{
			\label{subfig:gge}
			\includegraphics[width=0.48\linewidth]{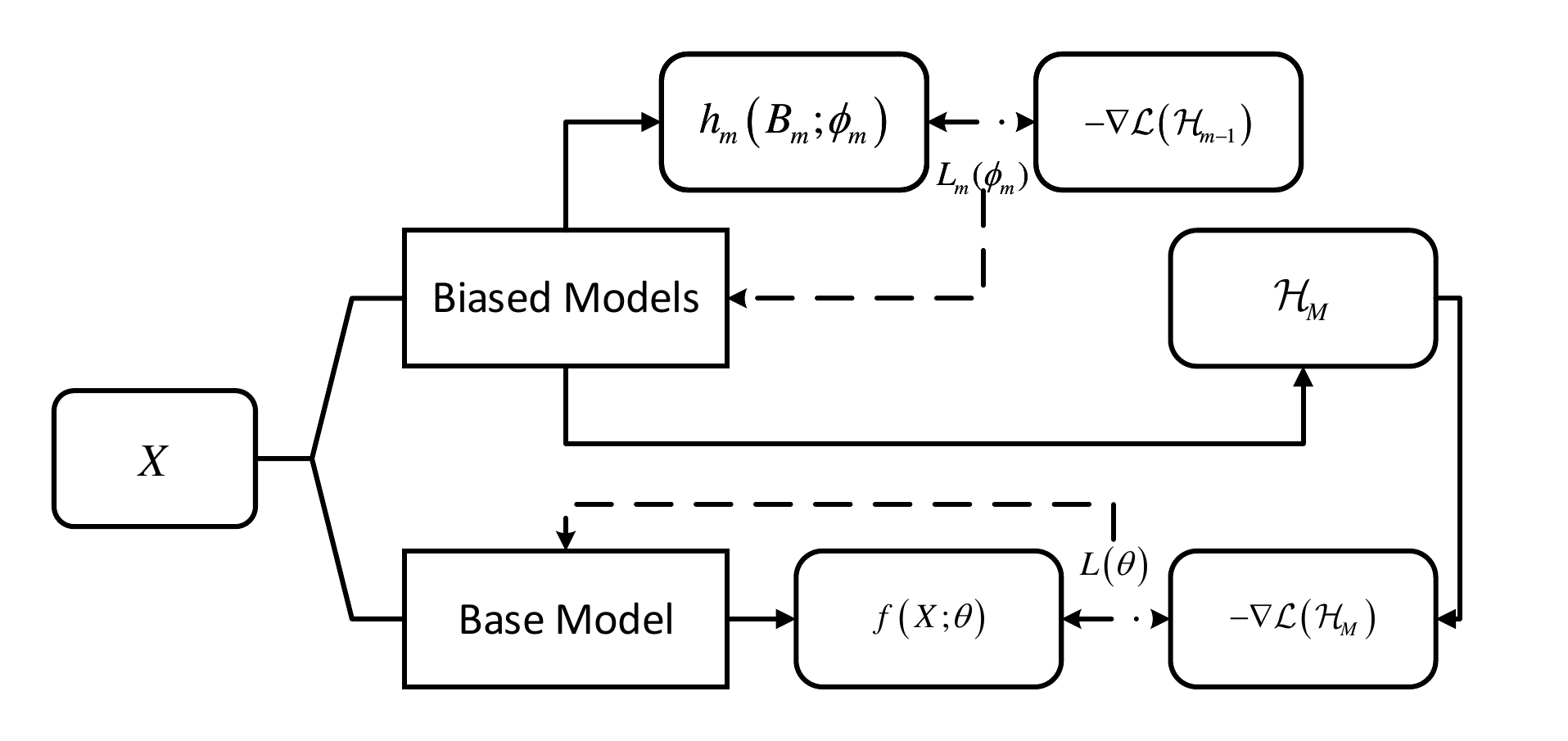}
		}
		\subfigure[GGD$_{cr}$]{
			\label{subfig:cr}
			\includegraphics[width=0.48\linewidth]{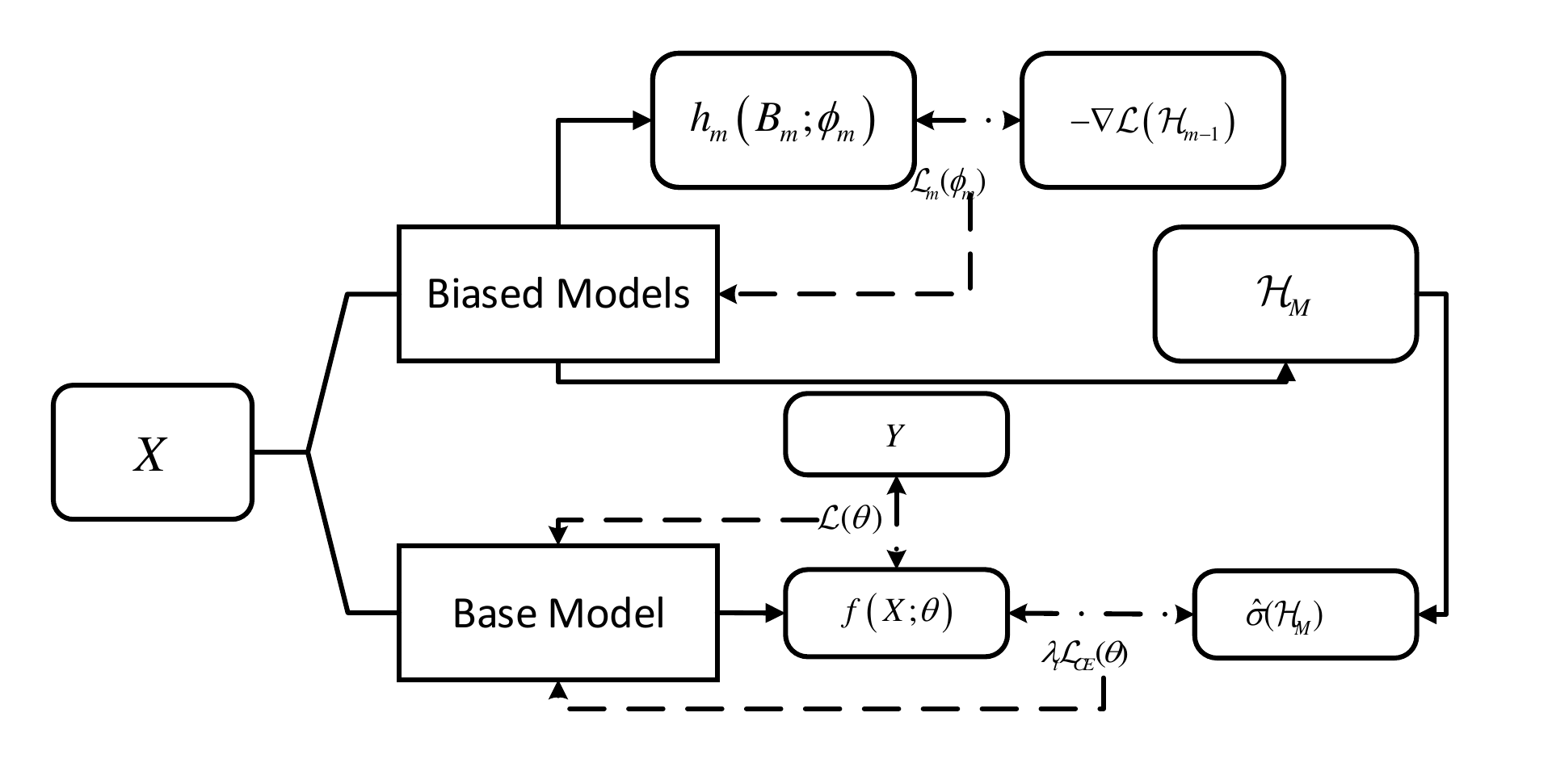}
		}
	\end{center}
\vspace{-1em}
	\caption{Comparison between GGD$_{gs}$ and GGD$_{cr}$. GGD$_{gs}$ (GGE) uses the gradient from the biased model as the pseudo label while GGD$_{cr}$ enlarges the prediction discrepancy between the base model and the biased model with curriculum learning. GGD$_{gs}$ is a special case of GGD when $\lambda_t=1$ under CE loss.}
	\label{fig:model}
\end{figure*}


\subsection{General Greedy De-bias Learning \label{cr}}
As shown in \cite{han2021greedy}, GGD$_{gs}$~(GGE) often over-estimates the biases of datasets. It achieves remarkable improvement on out-of-distribution data but may significantly degrades under the in-distribution setting. To overcome this critical issue, we first re-analyse the biased model in GGE under CE loss
\begin{equation}\label{ce}
\mathcal{L}_{CE}(Z, Y) = -\sum_{j=1}^{C} y_j \log(\sigma_j),
\end{equation}
with
\begin{equation}\label{softmax}
\sigma_j = \frac{e^{z_j}}{\sum_{k=1}^{C} e^{z_k}},
\end{equation}
where $Z = \{z_j\}_{j=1}^{C} $ is the predicted logits, and $y_j \in \{0,1\}$ is the ground-truth label for the $j$-th class. $\sigma_j$ indicates the confidence of the biased model on $j$-th class.
The negative gradient of the loss function is 
\begin{equation}\label{g_sfce}
-\nabla \mathcal{L}(z_j) = y_j - \sigma_j.
\end{equation}
To make the range of pseudo labels consistent with the classification label space [0,1], $-\nabla L(z_j)$ is clipped to
\begin{equation}\label{se}
-\nabla \mathcal{\hat{L}}(z_j) = 
\begin{cases}
y_j - \sigma_j & y_j > 0\\
0              & y_j = 0
\end{cases}.
\end{equation}
The negative gradients access whether a sample can be solved based on the spurious correlation captured by certain biased model. 

Now, casting aside the viewpoint of gradient descent in functional space, we can also decompose the CE loss with $-\nabla \mathcal{\hat{L}}$ as pseudo label
\begin{equation}\label{ce_decom}
\begin{split}
\mathcal{L}_{CE}(f(X),& -\nabla \mathcal{\hat{L}}) = -\sum_{j=1}^{C} (y_j - \hat{\sigma_j}) \log(p_j) \\
&= -\sum_{j=1}^{C} y_j \log(p_j) + \sum_{j=1}^{C}  \hat{\sigma_j} \log(p_j) \\
&= \mathcal{L}_{CE}(f(X), Y) - \mathcal{L}_{CE}(f(X),  \hat{\sigma}),
\end{split}
\end{equation} 
where the reference prediction $\hat{\sigma} = Y \odot \sigma$ and $\odot$ is the element-wise product that equals to the clipping in Eq.~\ref{se}.

Based on Eq.~\ref{ce_decom}, the gradient ensemble actually aims to provide predictions that agree with the ground-truth but disagree with the biased models. $- \mathcal{L}_{CE}(f(X), \hat{\sigma})$ controls the degree of spurious relation to be reduced.
To this end, we can treat $- \mathcal{L}_{CE}(f(X), \hat{\sigma})$ as a regularization:
\begin{equation}\label{cl}
L(\theta) =  \mathcal{L}\left(f(X; \theta), Y  \right) - \lambda_t \mathcal{L}_{CE}(f(X),  \hat{\sigma}(\mathcal{H}_M)),
\end{equation}
where $\lambda_t$ denotes the weight of the regularization term. This more general framework is noted as General Greedy De-bias (GGD), where we only keep greedy strategy but get free from the negative gradient supervision. GGE is a special case of GGD when $\lambda_t=1$. We will denote GGE as GGD$_{gs}$ (Gradient Supervision) in the following paper. 

Furthermore, inspired by Curriculum Learning~\cite{2009curriculum}, $- \mathcal{L}_{CE}(f(X), \hat{\sigma})$ can be regarded as a \emph{soft} difficulty measurement for curriculum sample selection function. 
In practice, we formulate a Curriculum Regularization training scheme (GGD$_{cr}$), which gradually increases $\lambda_t$ along with the training process.
In this way, samples with spurious correlations to the labels can participate in the early stage of training. In the consequent training stage, the model will focus on the hard samples that cannot be solved by biased models, resulting in more stable prediction on out-of-distribution data. The overall optimization procedure GGD$_{cr}$ is shown in Algorithm~\ref{alg:cr}. Comparison between GGD$_{gs}$ and general GGD$_{cr}$ is shown in Fig.~\ref{fig:model}. 

\subsection{Discussions}

\subsubsection{Intuitive Explanation of GGD}
Section 3.2 has presented theoretical evidence for GGD$_{gs}$ from the aspect of model learning in functional space. More intuitively, GGD$_{gs}$ can also be regarded as a re-sampling strategy~\cite{2021disentangling}. 
For a sample that is easy to fit by biased models, $-\nabla \mathcal{\hat{L}}(z_i)$ (\emph{i.e.}, the pseudo label produced by the base model) will become relatively small. This makes $f(X;\theta)$ pay more attention to samples that are hard to fit by previous ensemble biased classifiers. As a result, the base model is not likely to learn biased features. 
This hard example mining process is experimentally demonstrated in Section~\ref{ic} and Fig.~\ref{fig:hard}. 

However, according to Eq.~\ref{se}, samples that demonstrate high spurious correlations (\emph{i.e.}, $-\nabla \mathcal{\hat{L}}(z_i) = 0$) will be discarded. If large groups of data are absent because of zero supervision, the representation learning of the base model with the gradient supervision may be under-fitted. 
Moreover, when the label distribution is skewed ({\it e.g.}, distribution bias in VQA-CP~\cite{2018vqacp}), the base model may over-estimate the bias in labels. This results in ``inverse" training bias and significant degradation on in-distribution test data. 
Experiments on long-tailed classification also revealed similar findings~\cite{2020decoupling,2021disalign}, which indicate that re-sampling a part of the data encourages a more balanced classifier but harms the representation learning stage, while learning with unbalanced data results in a biased classifier but still provides a good representation.

To alleviate the ``bias over-estimation", GGD$_{cr}$ provides a good relaxation of GGD$_{gs}$ by replacing the gradient supervision with a ``softer" Curriculum Regularization.
By adjusting $\lambda_t$, all data can participate in the base model learning in the early stage, thus the bias over-estimation can be well alleviated. 
We will further experimentally demonstrate these findings in Section~\ref{ic} and Section~\ref{discuss}.

\begin{table*}[t]
		\centering
		\renewcommand{\arraystretch}{1.4}
		\setlength{\tabcolsep}{1.8mm}
		\caption{Comparison on Biased-MNIST. $\rho_{\text{train}}$ and $\rho_{\text{test}}$ denote the level of texture bias during training and testing, respectively. $1k$ uses SimpleNet-1k as the biased model, $bg$ adopts the backgrounds as biased feature, and $se$ stands for the self-ensemble version. `Original' is the original MNIST without texture bias.}
		\label{tab:cmnist}
		\begin{tabular}{lccclccclcccc}
			\hline
			\multirow{2}{*}{$\rho_{\text{train}}$/$\rho_{\text{test}}$} & \multicolumn{3}{c}{0.990} &  & \multicolumn{3}{c}{0.995} &  & \multicolumn{3}{c}{0.999} & \multirow{2}{*}{Original} \\ 
			\cline{2-4} \cline{6-8} \cline{10-12} 
			& 0     & 0.1    & 0.990    &  & 0     & 0.1    & 0.995    &  & 0     & 0.1    & 0.999  & 
			\\ \cline{1-4} \cline{6-8} 
			\hline 
			Baseline                   & 77.80\tinypm{1.30} & 80.11\tinypm{0.81} & 99.76\tinypm{0.06} &  & 53.78\tinypm{2.08} & 58.03\tinypm{2.73} & 99.82\tinypm{0.08} &  & 10.44\tinypm{2.36} & 17.39\tinypm{3.71} & 99.80\tinypm{0.09}  & 98.78\tinypm{0.15} \\
			\hline \hline 
			ReBias$^{1k}$~\cite{2020rebias}   & 85.21\tinypm{0.59} & 87.80\tinypm{0.62} & 99.77\tinypm{0.06} &  & 73.60\tinypm{1.18} & 75.95\tinypm{1.58} & 99.70\tinypm{0.22} &  & 32.84\tinypm{4.02} & 37.52\tinypm{2.37} & 99.87\tinypm{0.01} & 99.05\tinypm{0.08} \\
			RUBi$^{1k}$~\cite{2019rubi}    & 87.36\tinypm{3.59} & 91.30\tinypm{2.18} & 99.15\tinypm{0.29} &  & 77.84\tinypm{6.53} & 82.22\tinypm{4.14} & 99.21\tinypm{0.67} &  & 30.25\tinypm{9.64} & 37.20\tinypm{7.90} & 92.15\tinypm{7.71}  & 98.90\tinypm{0.04} \\
			GGD$_{gs}^{1k}$                & {\bf92.79\tinypm{0.76}} & {\bf 94.22\tinypm{0.78}} & 98.69\tinypm{0.56} &  & {\bf 91.27\tinypm{0.25}} & {\bf 91.80\tinypm{0.59}} & 98.16\tinypm{1.00} &  & 67.57\tinypm{4.31} & {\bf 70.77\tinypm{3.99}} & 86.84\tinypm{6.35} & 98.64\tinypm{0.04}  \\
			GGD$_{cr}^{1k}$                  & 91.78\tinypm{1.18} & 92.30\tinypm{1.17} & 99.64\tinypm{0.30} &  & 83.90\tinypm{2.30} & 84.91\tinypm{2.68} & 99.28\tinypm{0.15} &  & {\bf 68.36\tiny{$\pm$ 1.89}} & 70.70\tinypm{2.02} & 99.25\tinypm{0.35} & 99.14\tinypm{0.05} \\
			\hline
			ReBias$^{bg}$~\cite{2020rebias}   & 84.95\tinypm{1.63} & 86.88\tinypm{1.96} & 99.66\tinypm{0.25} &  & 74.27\tinypm{3.50} & 76.20\tinypm{1.78} & 99.74\tinypm{0.12} &  & 27.74\tinypm{8.07} & 34.20\tinypm{6.67} & 99.87\tinypm{0.01} & 99.87\tinypm{0.17} \\
			RUBi$^{bg}$~\cite{2019rubi}    & 88.65\tiny{$\pm$0.47} & 89.67\tiny{$\pm$0.64} & 99.59\tiny{$\pm$0.33} &  & 78.19\tinypm{5.06} & 80.50\tinypm{4.18} & 98.79\tinypm{1.07} &  & 21.07\tinypm{6.78} & 27.59\tinypm{6.39} & 90.16\tinypm{4.35} & 98.70\tinypm{0.06}  \\
			GGD$_{gs}^{bg}$                & \textbf{93.78\tinypm{1.34}} & \textbf{94.46\tinypm{1.09}} & 99.01\tinypm{0.42} &  & \textbf{90.34\tinypm{0.95}} & \textbf{91.26\tinypm{1.06}} & 99.38\tinypm{0.39} &  & 61.81\tinypm{4.29} & 66.00\tinypm{4.77} & 91.25\tinypm{1.98}  & 98.77\tinypm{0.07} \\
			GGD$_{cr}^{bg}$                  & 90.64\tinypm{0.84} & 91.95\tinypm{0.91} & 99.82\tinypm{0.05} &  & 86.20\tinypm{1.41} & 87.02\tinypm{1.04} & 99.68\tinypm{0.10} &  & \textbf{62.96\tinypm{5.74}} & \textbf{67.62\tinypm{4.51}} & 99.41\tinypm{0.41} & 99.07\tinypm{0.13}  \\
			\hline
			ReBias$^{se}$~\cite{2020rebias}   & \textbf{83.77\tinypm{0.81}} & \textbf{85.76\tinypm{0.71}} & 99.77\tinypm{0.08} &  & \textbf{75.06\tinypm{3.35}} & \textbf{77.25\tinypm{3.61}} & 99.84\tinypm{0.08} &  & 31.82\tinypm{3.49} & 38.41\tinypm{2.61} & 99.87\tinypm{0.02} &  99.03\tinypm{0.06} \\
			RUBi$^{se}$~\cite{2019rubi}    & 27.37\tinypm{8.04} & 33.47\tinypm{6.20} & 89.14\tinypm{8.02} &  & 16.23\tinypm{6.95} & 22.96\tinypm{5.84} & 95.77\tinypm{5.18} &  & 10.21\tinypm{5.28} & 16.67\tinypm{3.79} & 83.67\tinypm{11.51} & -  \\
			GGD$_{gs}^{se}$                & 79.35\tinypm{1.53} & 80.78\tinypm{2.11} & 94.65\tinypm{5.41} &  & 69.70\tinypm{3.22} & 72.49\tinypm{3.13} & 90.61\tinypm{1.48} &  & 38.72\tinypm{4.00} & 42.74\tinypm{3.25} & 76.24\tinypm{3.74} & 93.88\tinypm{8.87}  \\
			GGD$_{cr}^{se}$                  & {83.28\tinypm{0.65}} & {85.53\tinypm{1.32}} & 99.27\tinypm{0.27} &  & 72.91\tinypm{2.49} & 76.19\tinypm{1.61} & 99.34\tinypm{0.27} &  & \textbf{43.78\tinypm{2.82}} & \textbf{48.92\tinypm{1.64}} & 99.46\tinypm{0.21} & 98.94\tinypm{0.05}  \\
			\hline
		\end{tabular}
\end{table*}

\subsubsection{Probabilistic Justification} 

Following the assumptions in \cite{2019don}, for a given sample $x$, let $x^b$ be the biased features and $x^{-b}$ be the features except the biases. $x^{b}$ and $x^{-b}$ are conditionally independent given the label $y$. We have
\begin{equation}\label{prob}
	\log p(y|x^{-b}) = \log p(y|x) - \log p(y|x^b) + C,
\end{equation}
where $C$ is a constant term related to the given datasets. The detailed derivation is provided in the Appendix A. It is hard to distinguish the core features for the task of interest ($x^{-b}$) but it is easier to identify the dominant biases ($x^{b}$) based on the prior knowledge.
Eq.~\ref{prob} indicates that maximizing the likelihood $\log p(y|x^{-b})$ equals to maximizing $\log p(y|x)$ while minimizing $\log p(y|x^b)$. 

Assume the optimal biased model $h(x^b; \phi^*)$ has
\begin{equation}
	\phi^* = \arg \min_{\phi} \mathbb{E}_{<X,Y>} \mathcal{L}(h(x^b; \phi), y).
\end{equation}
Taking $q_{\phi^*}(y|x^b)$ as the distribution of optimal biased prediction $h(x^b; \phi^*)$, GGD alternatively minimizes $\log p(y|x^b)$ by enlarging the divergence between $p(y|x)$ and the biased reference $q_{\phi^*}(y|x^b)$. 
Maximizing Eq.~\ref{prob} is approximated as
\begin{equation}\label{app}
\arg \max_\theta \left(\log p_\theta(y|x) + D(p_\theta(y|x) || q_{\phi^*}(y|x^b))\right)
\end{equation}
where $\theta$ is the parameter of the base model that produce distribution $p(y|x)$ and $D(.||.)$ is the divergence between two distributions. In practice, we get diverse predictions by maximizing the cross-entropy between $p(y|x)$ and $q(y|x^b)$. Similar implementation also appears in \cite{pagliardini2022agree}. Eq.\ref{prob} provides a new justification of GGD from probabilistic formulation, which aims to maximum the log-likelihood of $\log p(y|x^{-b})$. Moreover, the precision of $q(y|x^b)$ is crucial. If the biased model captures the true correspondence too much, maximizing the divergence will harm the base model.

	\subsubsection{On the Trade-off between ID and OOD Performance}
	The key idea of greedy ensemble is similar as the Boosting strategy~\cite{1995boosting,schapire1990strength}. Boosting is to combine multiple weak classifiers with high bias but low variance to produce a strong classifier with low bias and low variance. Each base learner has to be weak enough, otherwise, the first few classifiers will easily over-fit to the training data~\cite{2006some}. Different from boosting that ensembles all weak learners, we make use of this over-fitting phenomenon but only use the last base model for prediction. This strategy removes specific spurious correlations with the biased models but also encounters the bias-overestimation with a single base model. To solve this problem, GGD introduce Curriculum Regularization, which trains the model with all ID data in the early training stage and then gradually focus on the hard samples.
	
	The trade-off between ID and OOD performance has already attracted much attention in the study of OOD generalization. Most of these methods assume that the OOD data is available during training~\cite{xie2020n,2021calibration,pagliardini2022agree,2022improved} or the model can be pre-trained on balanced data with few biases~\cite{hendrycks2019using,kumar2021fine}.
	Therefore, they can adaptively adjust the model with the given OOD data. However, for de-bias learning, the absence of OOD data makes the bias estimation more ill-posed and challenging. 
	The works in~\cite{2021introspective, kumar2022calibrated} share similar idea with our GGD$_{cr}$ in that they aim to make full use of the biased ID data to pursue a good trade-off between ID and OOD performance. However, both \cite{2021introspective} and \cite{kumar2022calibrated} have to train a biased and a de-biased model separately and then combine the two to achieve more robust predictions.
	In comparison, GGD learns the two models under the unified framework as in Algorithm 2. It does not require extra training cost of an original model, and can well adapt to any choice of base model, thus it gains more flexibility in real applications.

\subsection{General Applicability of GGD}
This section provides the detailed instantiation of GGD on specific tasks. In the following part, let $h(.)$ denote the biased model and $\hat{y}^. \in \mathcal{Y}$ denote the biased predictions, where the super-script represents the bias type.

\subsubsection{GGD with Single Explicit Bias}
In order to compare with existing explicit de-bias methods that focus on one single type of bias, we first test GGD on the texture bias in Biased-MNIST~\cite{2020rebias}. 

The dataset $\mathcal{D} = \{x_i, y_i, b_i \}_{i=1}^N$ consists of a synthetic image $x_i$, the annotated digit label $y_i$, and the background color $b_i$. We aim to predict the digit number $\hat{y_i}$ with the input image $x_i$
\begin{equation}\label{bmnist}
\hat{y_i} = f(x_i),
\end{equation} 
where the base model $f(.)$ is a neural network trained with CE loss.

The bias for Biased-MNIST comes from the spurious correlation between the digits and the background colors.
In practice, we define two different kinds of bias models. In the first case, the biased prediction $B^i_t$ of an image sample $x_i$ is extracted with a low capacity model
\begin{equation}\label{1k}
\hat{y}^t_i = h_{1k}(x_i).
\end{equation} 
$h_{1k}(.)$ is the SimpleNet-1k~\cite{2020rebias} with kernel size $1 \times 1$.
It will predict the target class of an image only through the local texture cues due to small receptive fields. 

In the second case, we provide the explicit background $b_i$ for bias extraction
\begin{equation}\label{bg}
\hat{y}^t_i = h_{bg}(b_i).
\end{equation} 
$h_{bg}(.)$ is a common neural network similar to the base model but the input is only a background image without digits. Therefore, the biased model will purely make predictions according to the texture bias. The experimental analysis is provided in Section~\ref{image_cls}.

\subsubsection{GGD with Self-Ensemble}
For tasks like Adversarial QA~\cite{2017ad_squad}, the task-specific biases are hard to distinguish. 
For de-bias learning at the lack of prior knowledge, we design a more flexible version of GGD with Self-Ensemble, named GGD$^{se}$. The biased predictions $B_{se}$ is captured with
\begin{equation}\label{self}
\hat{y}^{se}_i = h_{se} \left(x_i  \right),
\end{equation}
where $h_{se}(.)$ is another neural network that has the same architecture and optimization scheme as the baseline model. 
Since the baseline model usually tends to over-fit the dataset biases, $h_{se}(.)$ can implicitly capture the biases without task-specific prior knowledge. 

In the experiments, we will demonstrate the hard-example-mining mechanism of GGD$^{se}$ on Adversarial SQuAD~\cite{2017ad_squad} in Section~\ref{qa} and further verify its generalization ability on all the other three tasks.


\begin{figure*}[t]
	\centering
	\subfigure[{Baseline}]{
		\begin{minipage}{0.25\linewidth}
			\includegraphics[width=\linewidth]{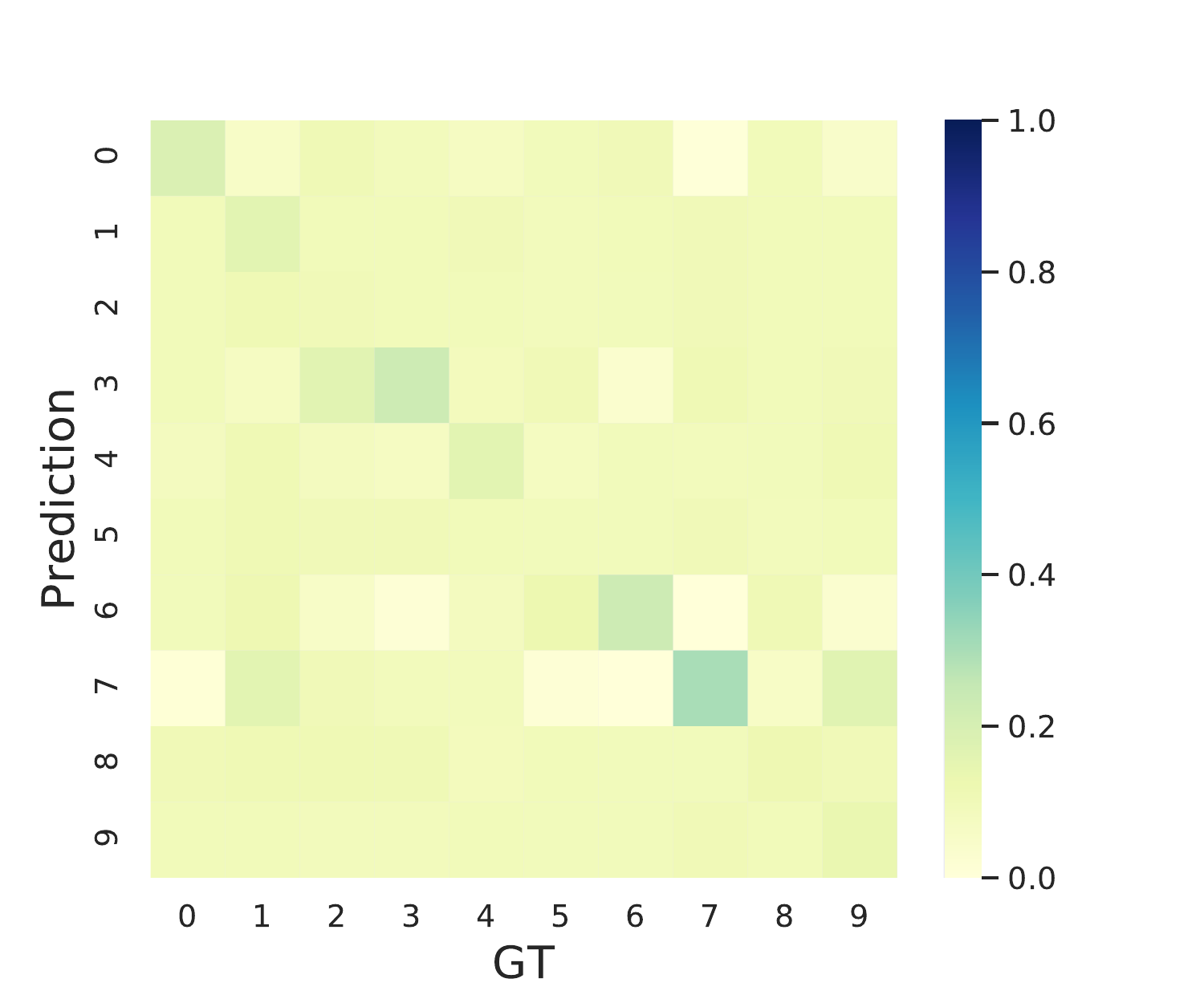}
			\includegraphics[width=\linewidth]{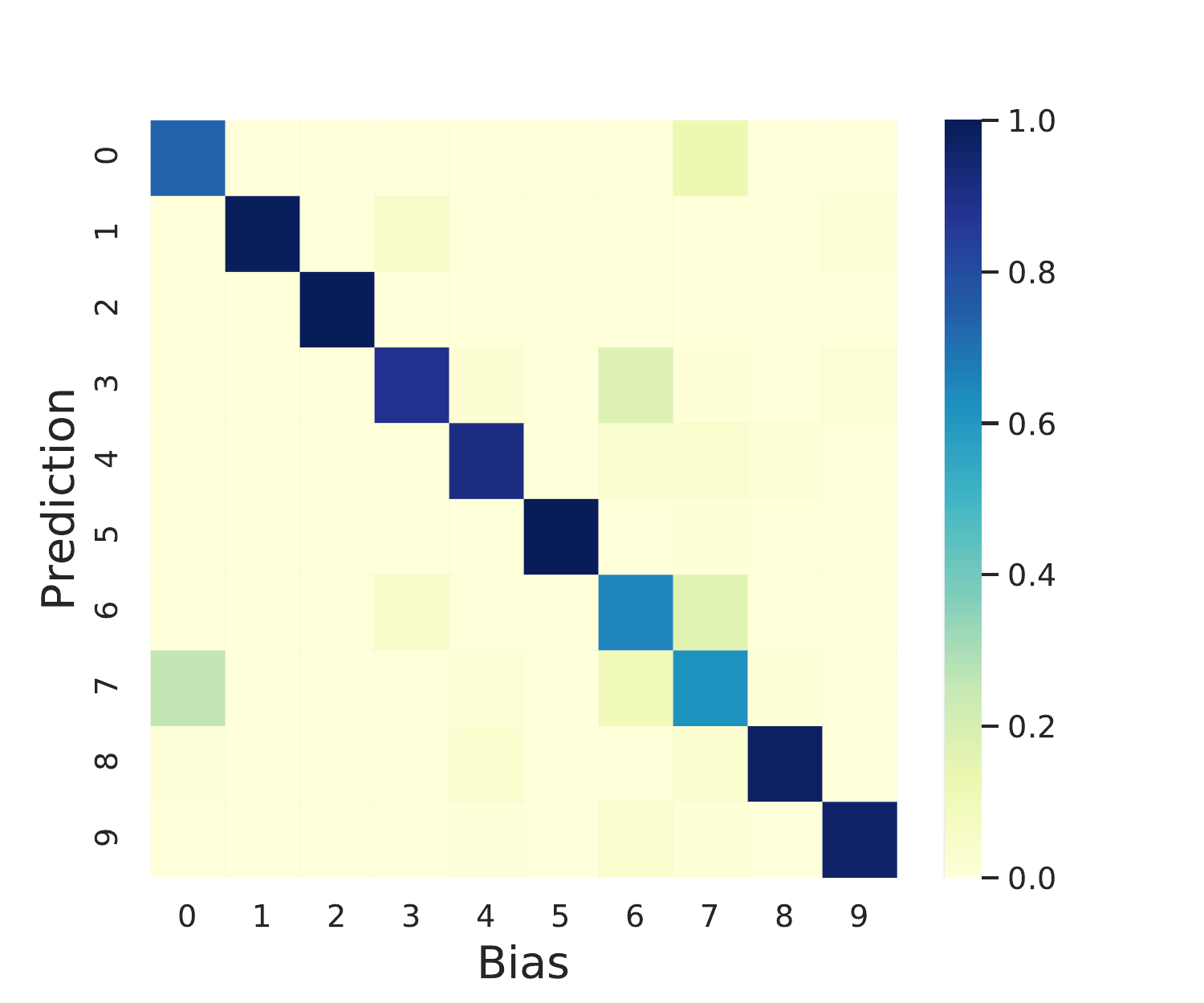}
		\end{minipage}	
	}
	\hspace{-1.5em}
	\subfigure[{ReBias}]{
		\begin{minipage}{0.25\linewidth}
			\includegraphics[width=\linewidth]{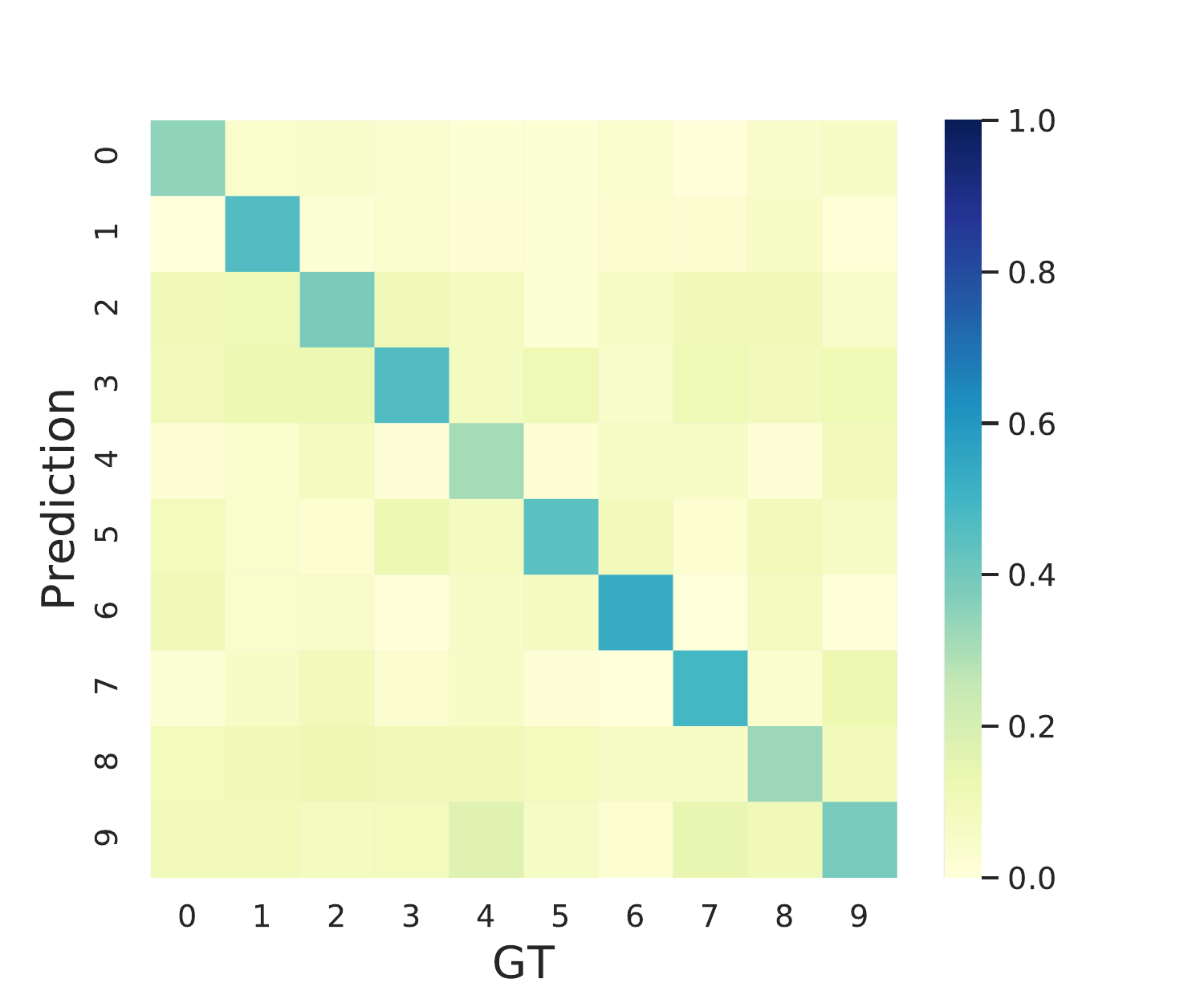}
			\includegraphics[width=\linewidth]{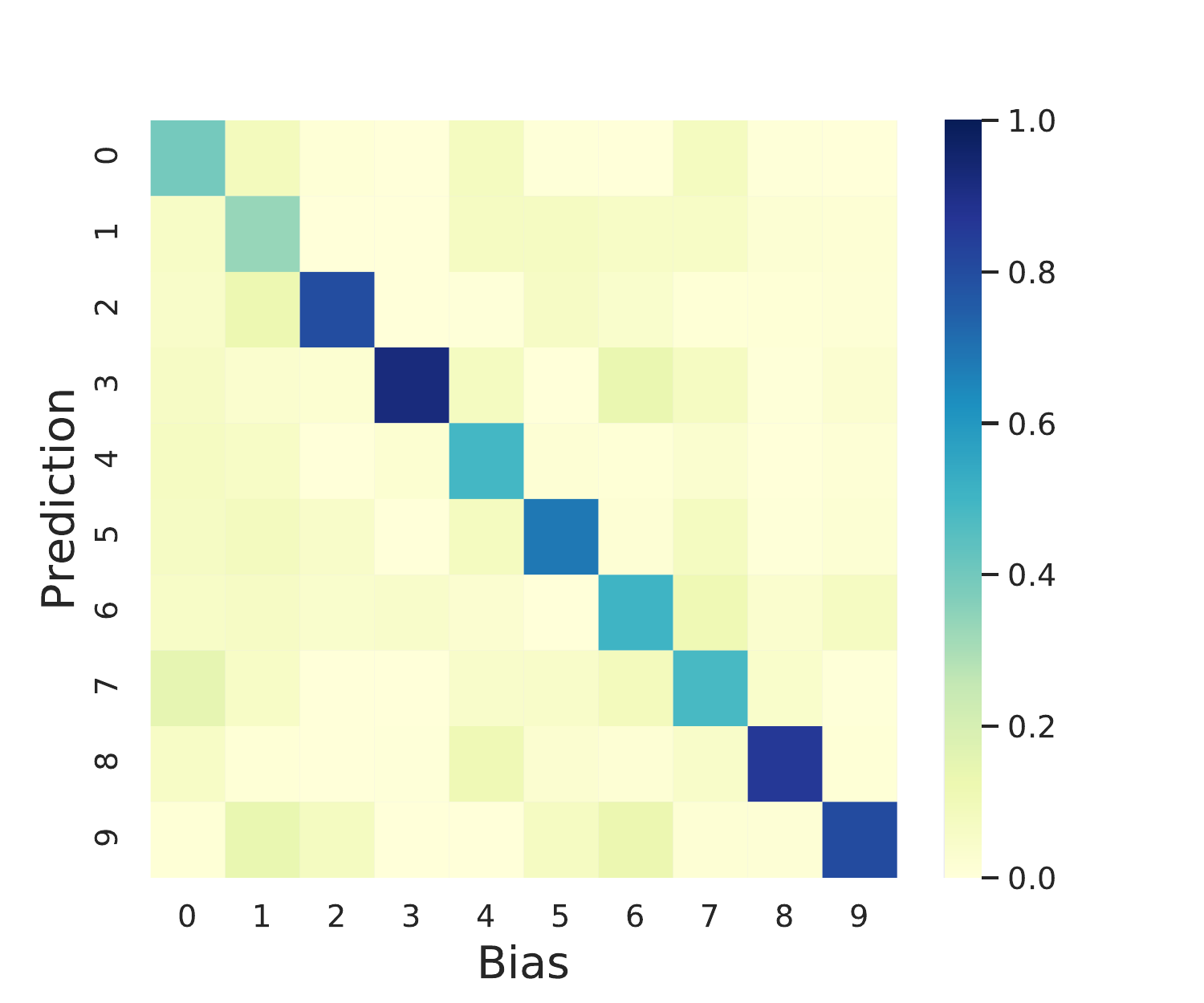}
		\end{minipage}	
	}
	\hspace{-1.5em}
	\subfigure[GGD$_{gs}^{1k}$]{
		\begin{minipage}{0.25\linewidth}
			\includegraphics[width=\linewidth]{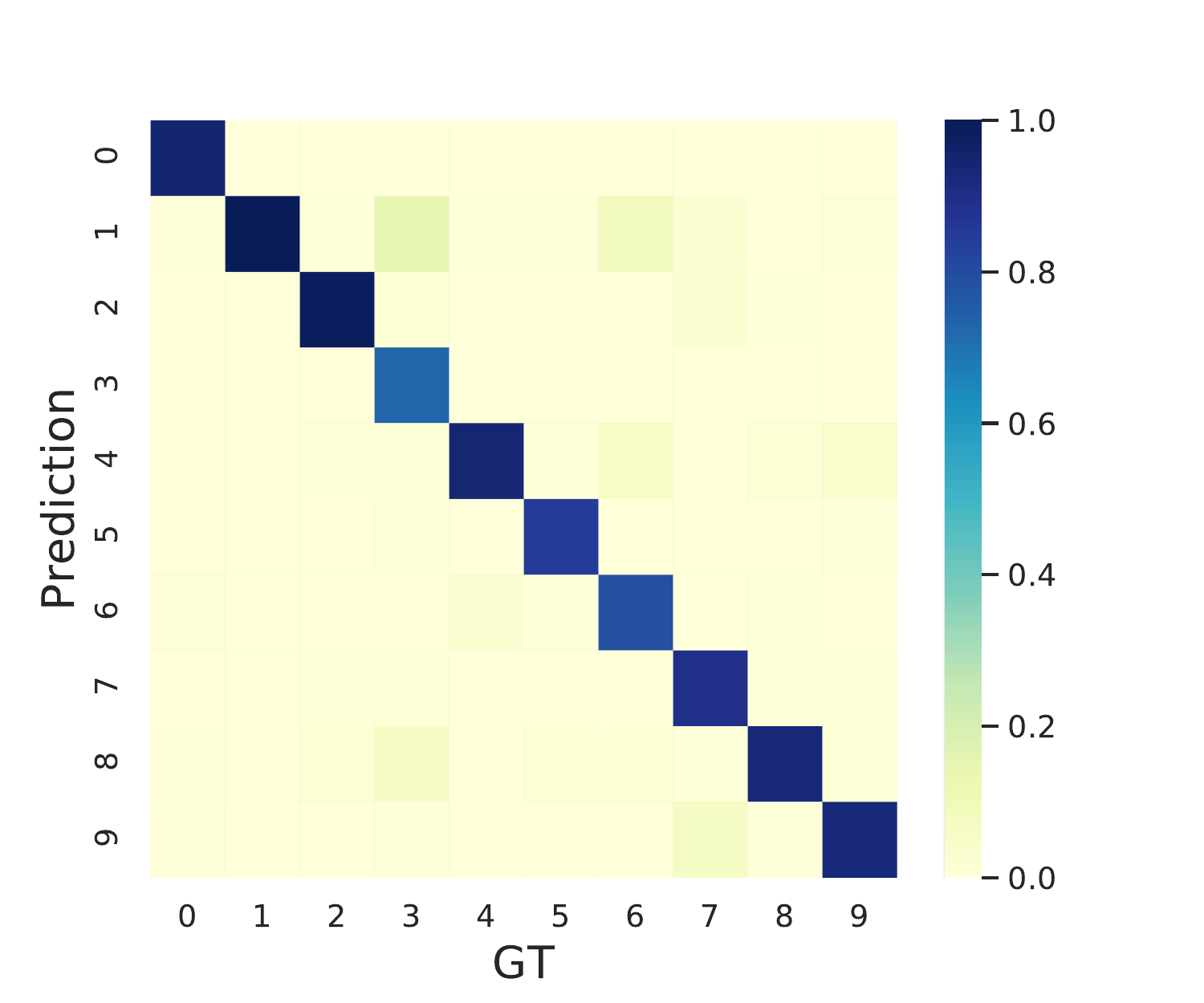}
			\includegraphics[width=\linewidth]{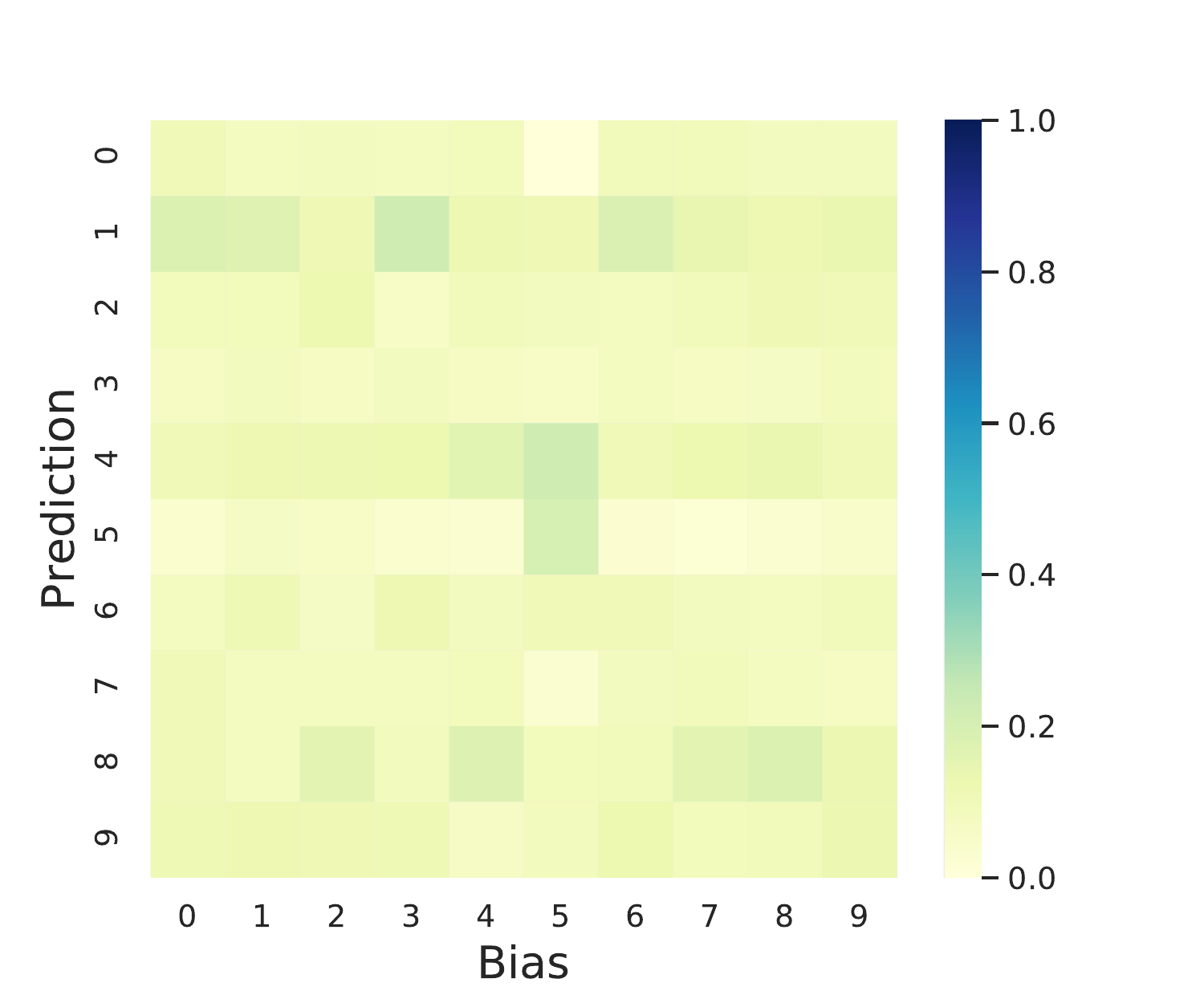}
		\end{minipage}	
	}
	\hspace{-1.5em}
	\subfigure[GGD$_{cr}^{1k}$]{
		\begin{minipage}{0.25\linewidth}
			\includegraphics[width=\linewidth]{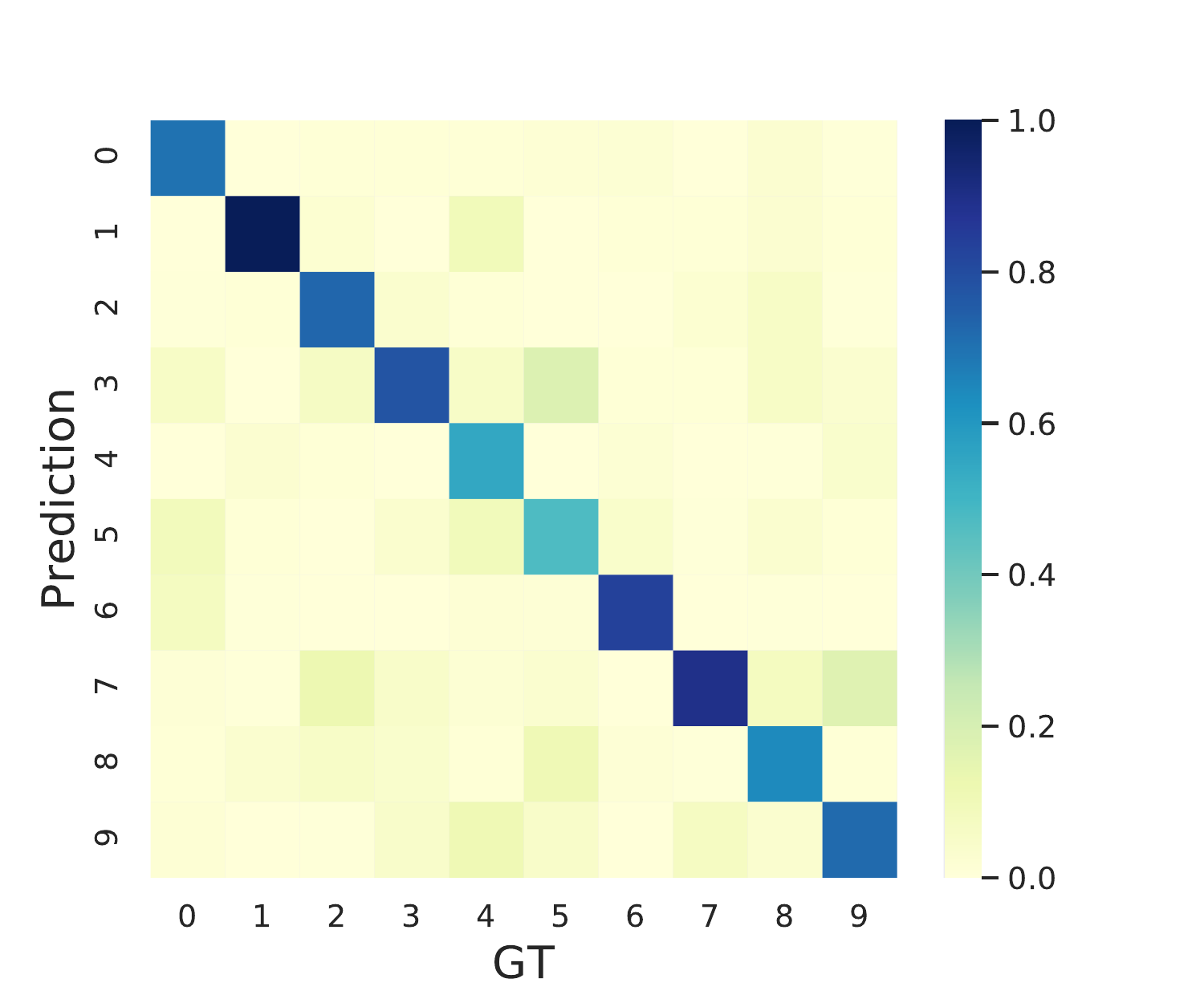}
			\includegraphics[width=\linewidth]{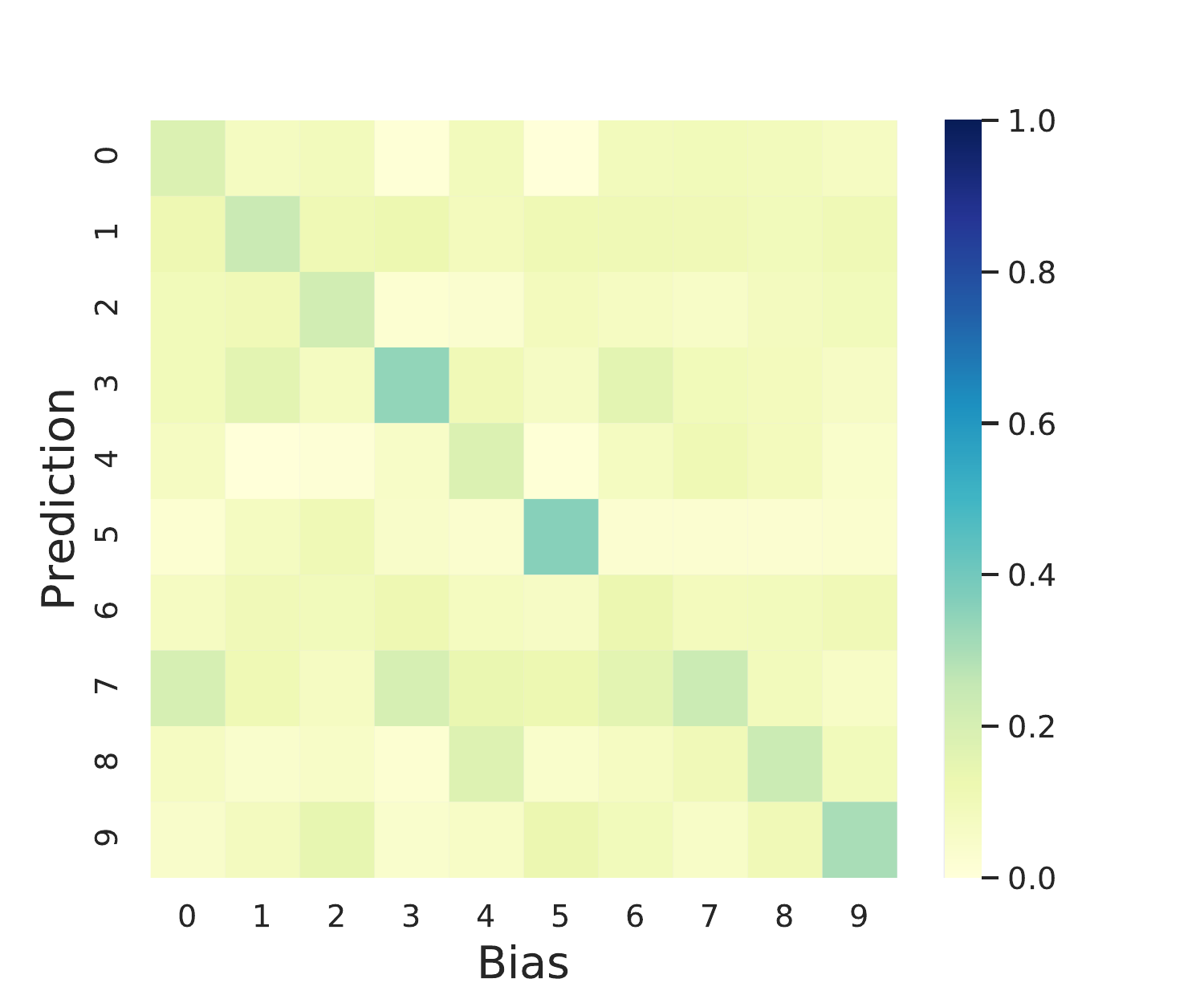}
		\end{minipage}	
	}
	
	\caption{Per-class Accuracies on Biased MNIST. All methods are trained with $\rho_{\text{train}}=0.999$ and tested on $\rho_{\text{test}}=0.1$. The upper row is the confusion matrix between the predictions and the ground-truth labels; the lower row shows the confusion matrix between the predicted labels and the background color labels.
	}
	\label{fig:matrix}
\end{figure*}

\subsubsection{GGD with Multiple Biases}
To verify whether GGD can handle multiple types of biases, we conduct experiment on the Language bias in VQA. As analysed in ~\cite{han2021greedy}, the language bias is mainly composed of two aspects, \emph{i.e.}, distribution bias and shortcut bias.

We consider the formulation of VQA task as a classification problem. Given a dataset $\mathcal{D} = \{v_i, q_i, a_i \}_{i=1}^N$ consisting of an image $v_i \in \mathcal{V}$, a question $q_i \in \mathcal{Q}$ and a labeled answer $a_i \in \mathcal{A}$, we need to optimize a mapping $f_{VQ}: V \times Q \rightarrow \mathbb{R}^C$ which produces a distribution over the $C$ answer candidates. 
The function is as follows
\begin{equation}\label{baseline}
\tilde{a}_i = f_\theta(v_i, q_i) = c\left(m\left(e_v(v_i), e_q(q_i) \right)\right),
\end{equation}
where $e_v: \mathcal{V} \rightarrow \mathbb{R}^{n_v \times d_v}$ is an image encoder, $e_q: \mathcal{Q} \rightarrow \mathbb{R}^{n_q \times d_q}$ is a question encoder, $m(.)$ stands for the multi-modal fusion module, and $c(.)$ is the multi-layer perception classifier. The output vector $\tilde{a} \in \mathbb{R}^C$ indicates the probability distribution on all the answer candidates. 

The distribution bias is the statistical answer distribution under certain question types
\begin{equation}\label{db}
\hat{y}^{d}_i = p(a_i | t_i),
\end{equation}
where $t_i$ denotes the type of question $q_i$, such as ``what color'', ``is this'', {\it etc.}, in VQA v2~\cite{2015vqa}.

The shortcut bias is the semantic correlation between specific QA pairs, which can be modeled as a question-only branch similar to \cite{2019rubi}
\begin{equation}\label{sb}
\hat{y}^{q}_i = c_q \left(e_q(q_i)  \right),
\end{equation}
where $c_q: Q \rightarrow \mathbb{R}^C$.

To verify whether GGD can handle compositional biases, we design different versions of GGD which ensemble distribution bias, shortcut bias and both biases. The experimental results are shown in Section~\ref{vqa}.

%% file: experiment.tex
\section{Experiments}
In this section, we present experiments for GGD on both ID and OOD settings. With respect to different types of biases, experiments on image classification~\cite{2020rebias}, QA~\cite{2017ad_squad} and VQA~\cite{2015vqa} are shown afterwards, corresponding to CV, NLP and Vision-Language tasks. Note that GGD is a general de-bias framework, and it will be an interesting issue in further study to apply our methods on other tasks that also suffer from dataset biases.

\subsection{Image Classification \label{image_cls}}
\subsubsection{Dataset}
{\bf Biased MNIST}. To better analyse the the properties of GGD, we first verify our model on Biased MNIST, where we can have full control over the amount of bias during training and evaluation. Biased MNIST~\cite{2020rebias} is modified from MNIST~\cite{1998mnist} which introduces the color bias that highly correlates with the label $Y$ during training. 

On Biased-MNIST, 10 different colors are selected for each digit $y \in \{0,\dots,9\}$. For each image of digits $y$, we assign a pre-defined color with probability $\rho$ and any other color with probability $1-\rho$. $\rho \in [0,1]$ controls the level of spurious correlation in train and test set. $\rho=0.99$ means 99\% images in the dataset are assigned with a background of the corresponding color. $\rho = 0.1$ is the unbiased condition with a uniform sampled background color.   


%

\subsubsection{Experimental Setups}
For image classification on biased MNIST, we use ResNet-18 as our baseline model. Model $1k$ in TABLE~\ref{tab:cmnist} denotes SimpleNet-1k with kernel size $1 \times 1$ proposed in \cite{2020rebias}. Biased model $bg$ uses the ResNet-18 with background color images as the input. All experiments use the same CE loss and baseline model. ``Original'' stands for MNIST dataset without biases. Considering the randomness in Biased-MNIST data generation, we report the mean and variance of 4 repeated experiments under different random seeds.

\subsubsection{Experimental Results \label{ic}} 
\noindent {\bf GGD Overcomes Bias}. 
As shown in TABLE~\ref{tab:cmnist}, GGD largely improves the OOD Accuracy on Biased MNIST. Under extremely biased training data ($\rho_{\text{train}} = 0.999$), the best performed method GGD$_{cr}^{1k}$ achieves 68\% accuracy on the unbiased test data ($\rho_{\text{test}} = 0$), which is over 6 times compared with the baseline model.
In Fig.~\ref{fig:matrix}, we provide per-class accuracy matrix for more detailed analysis. The diagnostic heat-map corresponds to unbiased and biased correlation respectively. We can observe that the vanilla ResNet-18 mainly captures the spurious correlation but confuses on the ground-truth digits. ReBias~\cite{2020rebias} can better capture the core correlations compared with the baseline but will be still fooled by the texture bias. GGD hardly relies on the background color as shown in Fig.~\ref{fig:matrix} (c) and (d), which demonstrate that our method can help a model to overcome certain kinds of bias via specially designed biased model.

\noindent {\bf Performance on Different Bias Level}. As shown in TABLE~\ref{tab:cmnist}, GGD achieves prominent performance gain on out-of-distribution tests across all bias levels while remaining stable on in-distribution data. 
Comparing with other methods that use the ensemble strategy, GGD surpasses RUBi~\cite{2019rubi} and ReBias~\cite{2020rebias} by a large margin under the same base model and biased models, especially when the training and testing data are extremely different ($\rho_{\text{train}} = 0.999$).
When training and testing under the unbiased situation, both GGD$_{cr}$ and GGD$_{gs}$ are stable if the bias type is known ahead. RUBi$^{se}$ fails on unbiased training set, which even continuously decreases under the original MNIST (`-' in Table~\ref{tab:cmnist}).

\begin{table}[t]
	\centering
	\renewcommand{\arraystretch}{1.4}
	\setlength{\tabcolsep}{5mm}
	\caption{Ablations for $\lambda_t$. ``Anneal" indicates the Curriculum Regularization that changes $\lambda_t$ from 0 to 1 along with training process.}
	\label{tab:iter&cr}
	\begin{tabular}{cccc}
		\hline
		\multirow{2}{*}{$\lambda_t$} & \multicolumn{3}{c}{Train 0.999} \\ \cline{2-4} 
		& 0     & 0.1    & 0.999    \\ \hline
		0 (Baseline)          & 8.95  & 16.24  & 99.87    \\  
		1 (GGD$_{gs}$)          & 68.31 & 71.34  & 91.42    \\
		0.95                   & 58.79 & 63.41  & 99.96    \\
		Anneal (GGD$_{cr}$)        & 67.01 & 70.17 & 99.58    \\
		\hline
	\end{tabular}
	
\end{table}

\noindent {\bf Ablations on Biased Models}. Besides biased model with small receptive field, we test another version of the biased model with a ground-truth background image as the biased features, shown as $bg$ in TABLE~\ref{tab:cmnist}. 
GGD works well with different biased models comparing with other methods. 

For implicit de-biasing, a biased model with the same structure as the baseline ResNet-18 is trained in the Self-Ensemble version GGD$^{se}$. As shown in TABLE~\ref{tab:cmnist}, GGD$_{cr}^{se}$ achieves the best performance when the bias information is not available. RUBi$^{se}$ corrupts under the self-ensemble setting. 
This demonstrates that GGD$^{se}$ can also implicitly remove biases even without the task-specific biased models, which is much more flexible compared with existing explicit de-bias methods.

\noindent {\bf GGD$_{gs}$ vs. GGD$_{cr}$}. As shown in TABLE~\ref{tab:cmnist}, although achieving high Accuracy on the out-of-distribution test data, GGD$_{gs}$ is not as robust as GGD$_{cr}$ with the increase of texture bias. Especially on the in-distribution test data, the accuracy is significantly lower than the baseline ResNet-18. 
Moreover, GGD$_{gs}^{se}$ is also very unstable in the later training stage, resulting in large variance according to different training data. On the other hand, GGD$_{cr}^{se}$ achieves comparable in-distribution performance against the baseline even on the original MNIST dataset.  

For better analysis of the GGD$_{gs}$ and GGD$_{cr}$, we design another ablation study on $\lambda_t$ in Eq.~\ref{cl} under $\rho_{\text{train}} = 0.999$.  As shown in TABLE~\ref{tab:iter&cr}, by slightly relaxing the regularization (changing $\lambda_t$ from 1.0 to 0.95), the in-distribution accuracy will be increased to the level of vanilla ResNet-18. This verifies our assumption that such degradation mainly comes from the \emph{completely} absence of samples with the spurious correlation (Section~\ref{cr}). Starting from this insight, we define $\lambda_t=\sin (\frac{\pi t}{2T})$ in GGD$_{cr}$, where $t$ is the current training epoch and $T$ is the number of total epochs. With this Curriculum Regularization, GGD$_{cr}$ achieves a good trade-off between in-distribution and out-of-distribution tests, remaining comparable in-distribution test accuracy against the baseline and OOD test against GGD$_{gs}$.

\begin{table}[t]
		\renewcommand{\arraystretch}{1.4}
		\setlength{\tabcolsep}{2.5mm}
		\centering
		\caption{Experimental Results on Adversarial QA. We provide the F1 score on Adversarial SQuAD AddSent split and SQuAD v1 dev split. ``Original" is trained with SQuAD train split, and ``Extra" is trained with extra Adversarial SQuAD AddSentOne split.  }
		\begin{tabular}{lcclcc}
			\hline
			\multirow{2}{*}{Method} & \multicolumn{2}{c}{Original} &  & \multicolumn{2}{c}{Extra} \\ \cline{2-3} \cline{5-6} 
			& AddSent         & Dev        &  & AddSent       & Dev       \\ \hline
			Baseline                & 46.61\tinypm{0.30}   & 87.61\tinypm{0.16} &  & 50.11\tinypm{0.35}  & 87.72\tinypm{0.21}     \\
			GGD$_{gs}^{se}$             & 48.05\tinypm{0.11}    & 87.98\tinypm{0.38}      &  & 52.44\tinypm{0.49}         & 87.01\tinypm{0.61}     \\
			GGD$_{cr}^{se}$               & 48.42\tinypm{0.20}   & 87.89\tinypm{0.20}      &  & 53.94\tinypm{0.09}         & 88.38\tinypm{0.27}     \\
			\hline
		\end{tabular}
		\label{tab:qa}
\end{table}

\subsection{Adversarial Question Answering \label{qa}}
\subsubsection{Dataset}
\begin{figure}[h]
	\begin{center}	
		\includegraphics[width=1.\linewidth]{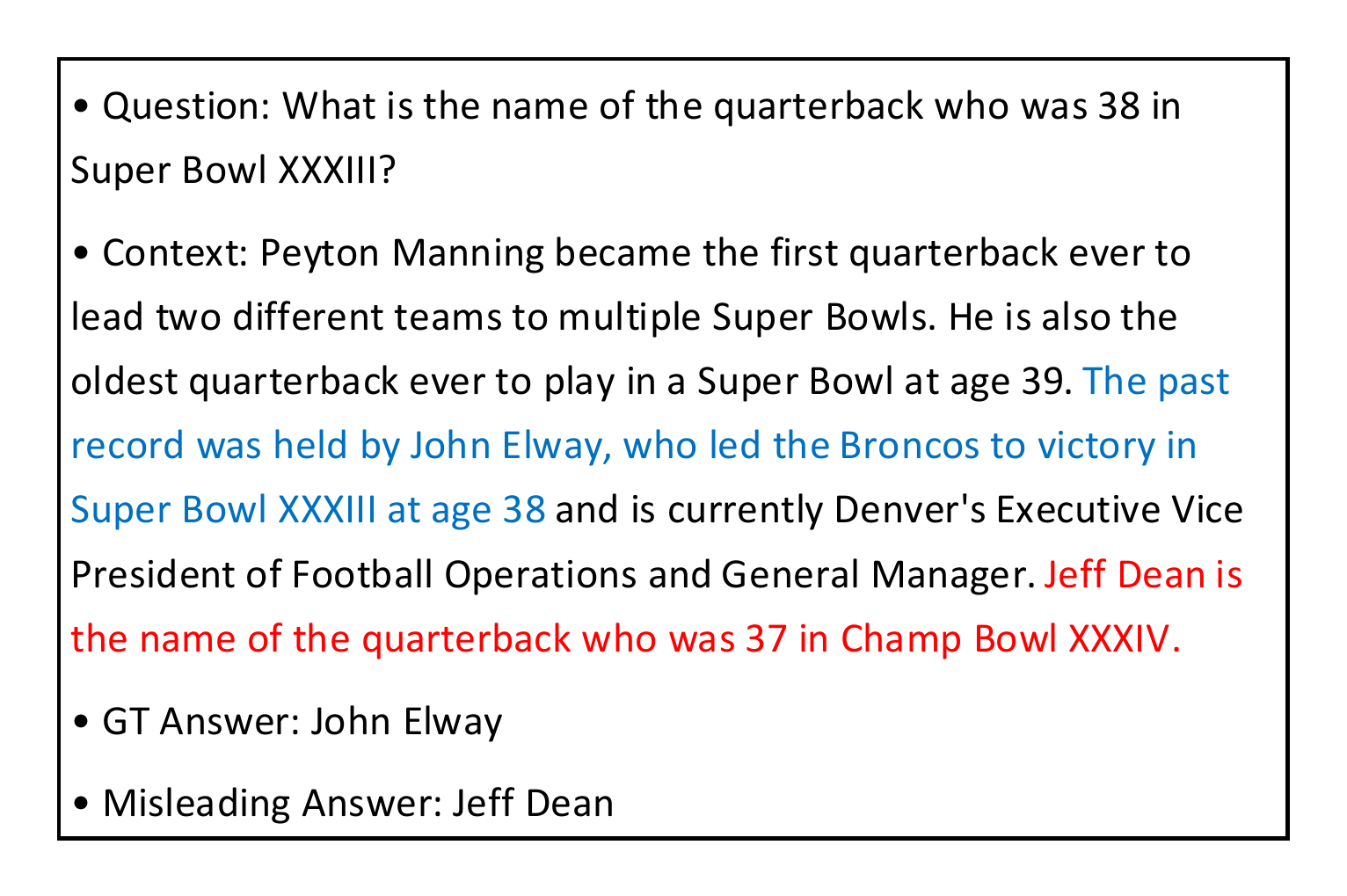}
	\end{center}
	\vspace{-2em}
	\caption{An example from Adversarial SQuAD. Blue sentence is the expected evidence, while red sentence is the distracting sentence to fool the models.}
	\label{fig:qa}
\end{figure}

For the NLP tasks, we choose the adversarial question answering (AdQA) to demonstrate the effectiveness of GGD.
We evaluate on the Adversarial SQuAD~\cite{2017ad_squad} dataset, which was built by adding distractive sentences to the passages in SQuAD~\cite{2016squad}. These sentences are designed to be very similar to the corresponding questions but with a few key semantic changes to ensure that they do not indicate the correct answer. Models that only focus on the similarity between question and context will tend to be misled by the new sentence. A sample from Adversarial SQuAD is shown in Fig.~\ref{fig:qa}.

\begin{table*}[t]
		\small
		\centering
		\renewcommand{\arraystretch}{1.4}
		\setlength{\tabcolsep}{1.4mm}
		\caption{Experimental results on VQA-CP v2 test set and VQA v2 val set of state-of-the-art methods. \textbf{\underline{Best}} and \textbf{second} performance are highlighted in each column. 
			Methods with * use extra annotations ({\it e.g.}, human attention (HAT), explanations (VQA-X), or object label information). Methods with $^\dagger$ use extra datasets at their pre-training stage. Methods with CGD are our reimplementation using released codes. Other results are reported in the original papers. The results are from 4 repeated experiments under different random seeds.}
		\label{tab:sota}
		\begin{tabular}{lccccccccccc}
			\hline
			\multirow{2}{*}{Method} & \multirow{2}{*}{Base} & \multicolumn{5}{c}{VQA-CP test} &  & \multicolumn{4}{c}{VQA v2 val} \\ \cline{3-7} \cline{9-12} 
			&                       & All & Y/N & Num. & Others & CGD &  & All  & Y/N  & Num. & Others \\ \hline
			GVQA~\cite{2018vqacp}   &  -     & 31.30 & 57.99 & 13.68 & 22.14 & -     & & 48.24 & 72.03 & 31.17 & 34.65 \\
			UpDn~\cite{2018bottomup}&  -     & 39.81\tinypm{0.05} & 43.09\tinypm{0.11} & 12.01\tinypm{0.09} & 45.81\tinypm{0.01} & 2.57\tinypm{1.23}  & & 63.57\tinypm{0.22} & 80.90\tinypm{0.10} & 42.58\tinypm{0.27} & \textbf{55.75\tinypm{0.07}} \\
			S-MRL~\cite{2019rubi}   &  -     & 38.46 & 42.85 & 12.81 & 43.20 & -     & & 63.10 & -     & -     & -     \\ \hline
			HINT*~\cite{2019hint}   & UpDn   & 47.50 & 67.21 & 10.67 & 46.80 & - & & 63.38 & 81.18 & 42.14 & 55.66 \\
			SCR*~\cite{2019SCR}     & UpDn   & 49.45 & 72.36 & 10.93 & 48.02 & -     & & 62.2  & 78.8  & 41.6  & 54.4  \\
			AdvReg.~\cite{2018overcoming}& UpDn   & 41.17 & 65.49 & 15.48 & 35.48 & -     & & 62.75 & 79.84 & 42.35 & 55.16     \\
			RUBi~\cite{2019rubi}    & UpDn   & 46.68\tinypm{0.89} & 68.43\tinypm{3.86} & 11.64\tinypm{0.26} & 44.53\tinypm{0.17} & 7.88\tinypm{1.41}  & & 58.74\tinypm{0.54} & 67.17\tinypm{6.42} & 39.85\tinypm{1.40} & 54.12\tinypm{0.84}     \\
			LM~\cite{2019don}       & UpDn   & 49.13\tinypm{1.07} & 72.38\tinypm{3.12} & 14.49\tinypm{0.71} & 46.42\tinypm{0.04} & 9.17\tinypm{2.10}     & & 63.46\tinypm{0.17} & 81.15\tinypm{0.04} & 42.27\tinypm{0.21} & 55.61\tinypm{0.34} \\
			LMH~\cite{2019don}      & UpDn   & 53.30\tinypm{0.70} & 73.47\tinypm{0.38} & 31.90\tinypm{4.37} & 47.92\tinypm{0.10} & 10.54\tinypm{0.68} & & 58.06\tinypm{2.18} & 72.60\tinypm{6.73} & 37.33\tinypm{1.75} & 52.39\tinypm{2.59} \\
			DLP~\cite{2020DLP}      & UpDn   & 48.87 & 70.99 & 18.72 & 45.57 & - & & 57.96 & 76.82 & 39.33 & 48.54  \\
			GVQE*~\cite{2020reducing}& UpDn   & 48.75 & -     & -     & -     & -     & & \textbf{\underline{64.04}} & -     & -     & -    \\
			CSS*~\cite{2020counterfactual}& UpDn   & 40.32\tinypm{0.59} & 42.15\tinypm{1.28} & 12.44\tinypm{0.27} & 46.76\tinypm{0.33} & 8.58\tinypm{2.34}  & & 62.34\tinypm{1.85} & 79.50\tinypm{2.20} & 42.11\tinypm{1.06} & 55.50\tinypm{1.32}  \\
			CF-VQA(Sum)~\cite{2020cf-vqa}& UpDn   & 53.69 & \textbf{\underline{91.25}} & 12.80 & 45.23 & -     & & 63.65 & \textbf{\underline{82.63}} & \textbf{44.01} & 54.38 \\ \hline
			RUBi~\cite{2019rubi}    & S-MRL  & 47.11 & 68.65 & 20.28 & 43.18 & -     & & 61.16 & -     & -     & -     \\	
			GVQE*~\cite{2020reducing}& S-MRL  & 50.11 & 66.35 & 27.08 & 46.77 & -     & & 63.18 & -   & -  &  -   \\
			CF-VQA(Sum)~\cite{2020cf-vqa}& S-MRL  & 54.95 & \textbf{90.56} & 21.88 & 45.36 & -     & & 60.76 & 81.11 & 43.48 & 49.58 \\ \hline
			MFE~\cite{2020mfe}& LMH    & 54.55 & 74.03 &  \textbf{49.16} & 45.82 & -  & & - & - & - & -   \\
			CSS*~\cite{2020counterfactual}& LMH    & 58.27\tinypm{0.05} & 81.76\tinypm{1.34} & 48.99\tinypm{5.85} & 47.99\tinypm{0.12} & 6.34\tinypm{1.75}  & & 53.42\tinypm{0.26} & 58.32\tinypm{2.49} & 37.99\tinypm{1.15} & 54.53\tinypm{1.01}   \\
			SAR$^\dagger$~\cite{2021SAR}  & LMH    & \textbf{\underline{62.51}} & 76.40 & \textbf{\underline{59.40}} & \textbf{\underline{56.09}} & -  & & \textbf{\underline{65.79}} & 77.26 & \textbf{\underline{52.71}} & \textbf{\underline{60.52}}   \\
			\hline \hline
			GGD$_{gs}^{dq}$          & UpDn   & 56.95\tinypm{0.34} & 87.02\tinypm{0.30} & 25.97\tinypm{1.35} & 49.40\tinypm{0.28} & \textbf{\underline{15.24}\tinypm{0.93}} & & 59.51\tinypm{1.34} & 74.77\tinypm{3.22} & 39.46\tinypm{1.26} & 53.50\tinypm{0.69}  \\
			GGD$_{cr}^{dq}$          & UpDn   & \textbf{59.37\tinypm{0.26}} & 88.23\tinypm{0.29} & 38.11\tinypm{1.05} & \textbf{49.82\tinypm{0.40}} & \textbf{13.31\tinypm{1.69}} & & 62.15\tinypm{0.93} & \textbf{79.25\tinypm{2.19}} & 42.43\tinypm{0.21} & 54.66\tinypm{0.32} \\
			\hline
		\end{tabular}
\end{table*}

\subsubsection{Experimental Setups}
We use BiDAF~\cite{2017bidaf} as the base model. It introduces a multi-stage hierarchical process that represents the context at different levels of granularity and uses a bidirectional attention flow mechanism to obtain a query-aware context representation without early summarization. 
Since the word-level and semantic-level similarity is hard to distinguish by modelling, we only test GGD$^{se}$ in the following experiments.

The models are trained and validated on the original SQuAD train and val set, and test on the AddSent split of Adversarial SQuAD. In order to further verify the hard example mining mechanism behind GGD, we also design another ``Extra" setting in which we add the AddSentOne split of Adversarial SQuAD to the training set as additional hard samples. The performances are measured with F1 score, which are the weighted average of the precision and recall rate at the character level.

\subsubsection{Experimental Results}
The F1 scores for the OOD test on Adversarial SQuAD and ID test on SQuAD v1 are shown in TABLE~\ref{tab:qa}. The reported results are from four repeated experiments with different random seeds.
We find that both GGD$_{gs}$ and GGD$_{cr}$, trained with the SQuAD train set, only improve the performance by $\sim 2\%$. However, after adding a few hard examples from AddSentOne, both methods achieve much more improvement on Adversarial SQuAD compared with the baseline. GGD$_{cr}$ gets nearly 5 points gain over the BiDAF baseline that is already strong enough. This well demonstrates the power of the greedy learning in focusing on the hard/valuable samples from a biased dataset. 

However, the limited improvement in the original setting is likely caused by Self-Ensemble, where the baseline model BiDAF itself can hardly capture useful biases from the dataset. If we can define a better biased model that can access the word-level similarity, we may achieve better performance without extra training data.

\subsection{Visual Question Answering \label{vqa}}
\subsubsection{Dataset}
Data bias problems in multi-modal tasks are more challenging, where multiple data sources from different modalities should be jointly considered. In this section, we choose Visual Question Answering (VQA) as the representative multi-modal task for demonstration.
Neural networks~\cite{gao2019multi,2017n2nmn,2019lcgn,2019nscl,han2020interpretable} that model the correlations between vision and language have shown remarkable results on large-scale benchmark datasets~\cite{2015vqa,2017mfh,2017clevr,2019gqa}, but most VQA methods tend to rely on existing idiosyncratic biases in the datasets~\cite{2017mfh,2017analysis} and show poor generalization ability to out-of-domain data. In this section, we demonstrate the effectiveness of GGD on the challenging datasets VQA-CP v2~\cite{2018vqacp} and GQA-OOD~\cite{2021gqa-ood}.

{\bf VQA v2}~\cite{2015vqa} is a commonly used VQA dataset composed of real-world images from MSCOCO with the same train/validation/test splits. 
For each image, an average of three questions are generated, and 10 answers are collected for each image-question pair from human annotators. 
Following previous works, we take the answers that appeared more than 9 times in the training set as candidate answers, which produces 3129 answer candidates.

{\bf VQA-CP v2}~\cite{2018vqacp} dataset is derived from the VQA 2.0~\cite{2015vqa} but contains different answer distribution per question type between training and validation splits. Since it has different distribution on the train and test sets, the performance on this dataset better reflects models' generalization ability. 
VQA-CP v2 consists of 438,183 samples in the train set and 219,928 samples in the test set.

{\bf GQA-OOD}~\cite{2021gqa-ood} divides the test set of GQA~\cite{2019gqa} into majority (head) and minority (tail) groups based on the answer frequency within each ‘local group’, which is a unique combination of answer type ({\it e.g.}, colors) and the main concept ({\it e.g.}, ‘bag’, ‘chair’, {\it etc.}). The models are trained on the original GQA-balanced but tested on different fine-grained local groups.

\subsubsection{Experimental Setups}
In the following experiments, we use UpDn~\cite{2018bottomup} as our base model and the images are represented as object features pre-extracted with Faster R-CNN~\cite{2017frcnn}. The implementation details and the experiments on other base models are provided in the Appendix.
All methods are measured with Accuracy and Correct Grounding Difference (CGD) proposed in \cite{han2021greedy}. CGD evaluates whether the visual information is well taken in answer decision.

For ablation studies, we present five different versions of GGD. 
{\bf GGD$^{d}$} only removes the distribution bias.
{\bf GGD$^{q}$} only models the shortcut bias.
{\bf GGD$^{dq}$} makes use of both the distribution bias and the shortcut bias. 
{\bf GGD$^{se}$} is the self-ensemble version GGD, which takes the baseline model itself as the biased model.
{\bf GGD$^{dse}$} removes the distribution bias before Self-Ensemble.
The implementation details of above five ablations are provided in the Appendix.

\begin{table}[t]
	\renewcommand{\arraystretch}{1.4}
	\caption{Ablation study for different versions GGD on VQA-CP v2 test set and VQA v2 val set. ID indicates the overall Accuracy on VQA v2 val. \textbf{Best} results are highlighted in the columns. SE denotes the self-ensemble.}
	\label{tab:ablation}
	\begin{tabular}{l|lcccc|c}
		\hline
		Method      & All & Y/N & Others & Num. & CGD & ID\\ \hline
		Baseline & 39.89 & 43.01 & 45.80 & 11.88 & 3.91 & 63.79 \\
		SUM-DQ      & 35.46 & 42.66 & 38.01 & 12.38 & 3.10 & 56.85  \\
		LMH+RUBi    & 51.54 & 74.55 & 47.41 & 22.65 & 6.12 & 60.68  \\ \hline
		GGD$_{gs}^{d}$       & 48.27 & 70.75 & 47.53 & 13.42 & 14.31 & 62.79   \\
		GGD$_{gs}^{q}$  & 43.72 & 48.17 & 48.78 & 14.24 & 6.70 &  61.23 \\
		GGD$_{gs}^{dq}$ & 57.12 & 87.35 & 49.77 & 26.16 & \textbf{16.44} & 59.30   \\
		\hline
		GGD$_{cr}^{d}$ & 50.93 & 78.50 & 47.30 & 12.92 & 10.28 & 62.17   \\
		GGD$_{cr}^{q}$ & 55.81 & \textbf{88.59} & 48.74 & 20.96 & 13.46 & 62.48   \\
		GGD$_{cr}^{dq}$ & \textbf{59.57} & 88.44 & \textbf{50.23} & \textbf{36.95} & 13.92 & \textbf{63.11}  \\
		\hline\hline 
		GGD$_{gs}^{se}$   & 44.53 & 50.98 & 48.90 & 18.24 & 6.08 & 59.30    \\
		GGD$_{gs}^{dse}$  & 56.33 & 86.43 & 49.32 & 24.37 & 14.47 & 61.03 \\
		\hline
		GGD$_{cr}^{se}$ & 54.42 & 80.26 & 48.64 & 29.42 & 6.70 & 61.09  \\
		GGD$_{cr}^{dse}$ & 57.08 & 85.75 & 48.88 & 36.54 & 11.62 & 62.27 \\
		\hline
	\end{tabular}
\end{table}

\subsubsection{Experimental Results}
\noindent {\bf GGD can handle multiple biases}. In the first group of ablation study, we compare with the other two ensemble strategies to verify the effectiveness of the greedy learning.
{\bf SUM-DQ} directly sums up the outputs of biased models and the base model. 
{\bf LMH+RUBi} combines LMH~\cite{2019don} and RUBi~\cite{2019rubi}. It reduces distribution bias with LMH and shortcut bias with RUBi.
The implementation details for these two ablations are provided in the Appendix.

As shown in TABLE~\ref{tab:ablation}, SUM-DQ performs even worse than vanilla UpDn. LMH+RUBi does not make use of both kinds of biases, whose Accuracy is just similar to that of LMH.
On the other hand, both GGD$_{gs}$ and GGD$_{cr}$ surpass these two ablations by a large margin. This shows that the greedy strategy in GGD can really force the biased data to be learned with biased models in priority. As a result, the base model has to pay more attention to hard examples that are hard to solve based on the estimation of either distribution bias or shortcut bias. It needs to consider more visual information for the final decision. 

\begin{figure}[t]
	\begin{center}	
		\includegraphics[width=0.95\linewidth]{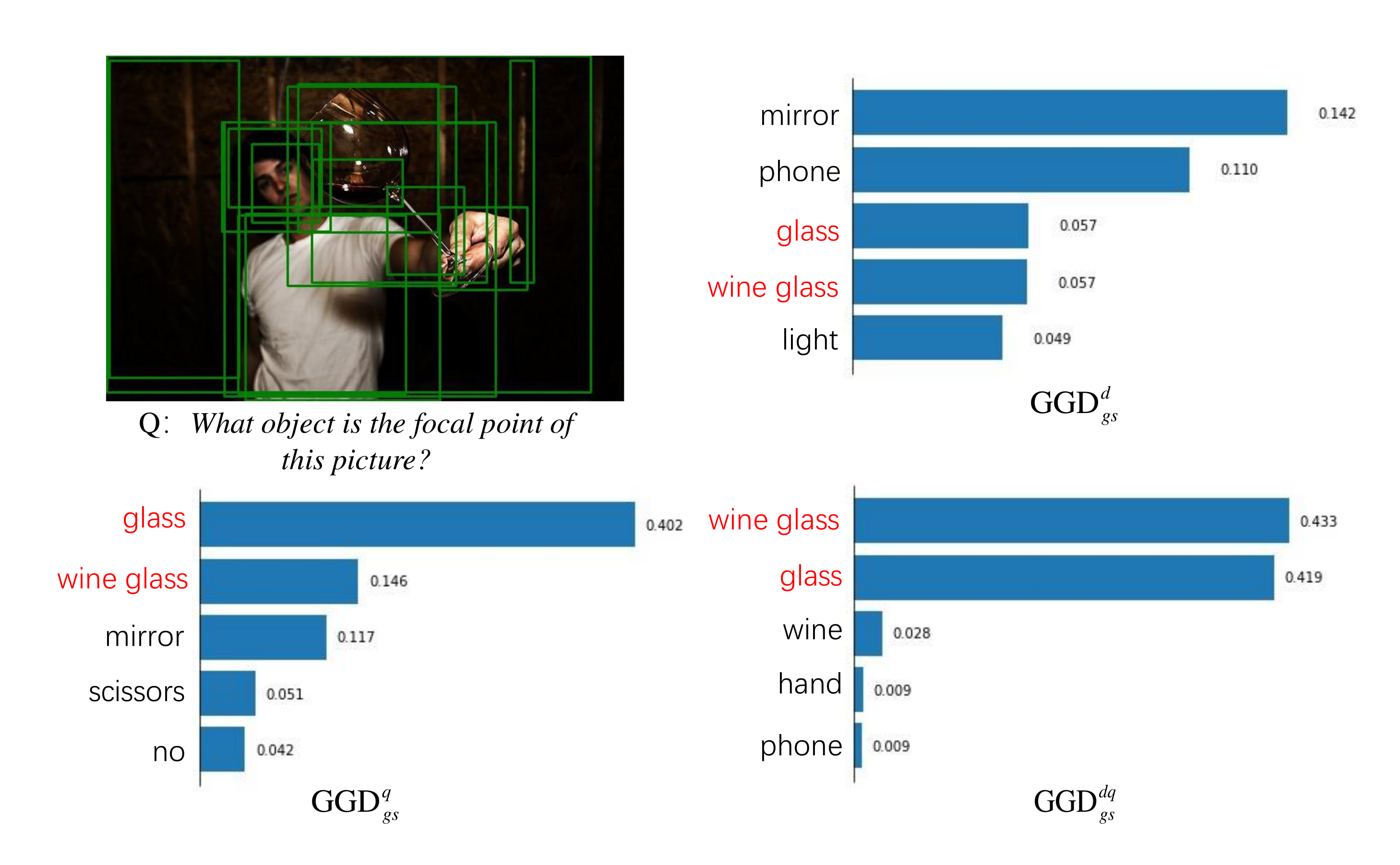}
	\end{center}
	\vspace{-2em}
	\caption{Predicted distribution for three variants of GGD$_{gs}$.}
	\label{fig:ab}
\end{figure}

In the second group of experiments, we directly compare GGD$^{d}$, GGD$^{q}$, and GGD$^{dq}$. 
As shown in TABLE~\ref{tab:ablation}, GGD$_{gs}^{dq}$ surpasses single-bias versions GGD$^{d}$ and GGD$^{q}$ by $\sim$10\%. This well verifies that GGD can reduce multiple biases with the greedy learning procedure.  
The case analysis in Figure~\ref{fig:ab} provides a more qualitative evaluation. It shows that GGD$^{d}$ uniforms predictions, which mainly improves Y/N as shown in TABLE~\ref{tab:ablation}. $B_q$ works like ``hard example mining" but will also introduce some noise ({\it e.g.}, ``mirror" and ``no" in this example) due to the unbalanced data distribution. GGD$^{dq}$ can make use of both biases. Reducing $B_d$ at first can further help the discovery of the hard examples with $B_q$ and encourage the base model to capture essential visual information.

\noindent {\bf Implicitly De-bias with Self-Ensemble}. 
In order to further discuss the generalizability of GGD, we also test a more flexible Self-Ensemble fashion GGD$^{se}$ on VQA-CP. 
As shown in TABLE~\ref{tab:ablation}, GGD$^{se}$ still surpasses UpDn without predefined biased features. Moreover, if we first remove distribution bias before Self-Ensemble, the performance of GGD$^{dse}$ is comparable to existing state-of-the-art methods as well. 

\noindent {\bf GGD$_{gs}$ vs. GGD$_{cr}$}. As shown in TABLE.~\ref{tab:ablation}, GGD$_{cr}$ largely alleviates the degradation on in-distribution test data VQA v2 val, which is even comparable to the original UpDn baseline. Moreover, It also gets better performance on VQA-CP under all GGD$^{d}$, GGD$^{q}$, GGD$^{dq}$, and GGD$^{se}$. The major improvement comes from the ``Num." and ``Other" question types which contain fewer samples in the training set. GGD$_{gs}$ harms the performance on these question types because of the greedily discarding of samples that is easy to answer.

\begin{figure*}[t]
	\begin{center}
		\includegraphics[width=0.99\linewidth]{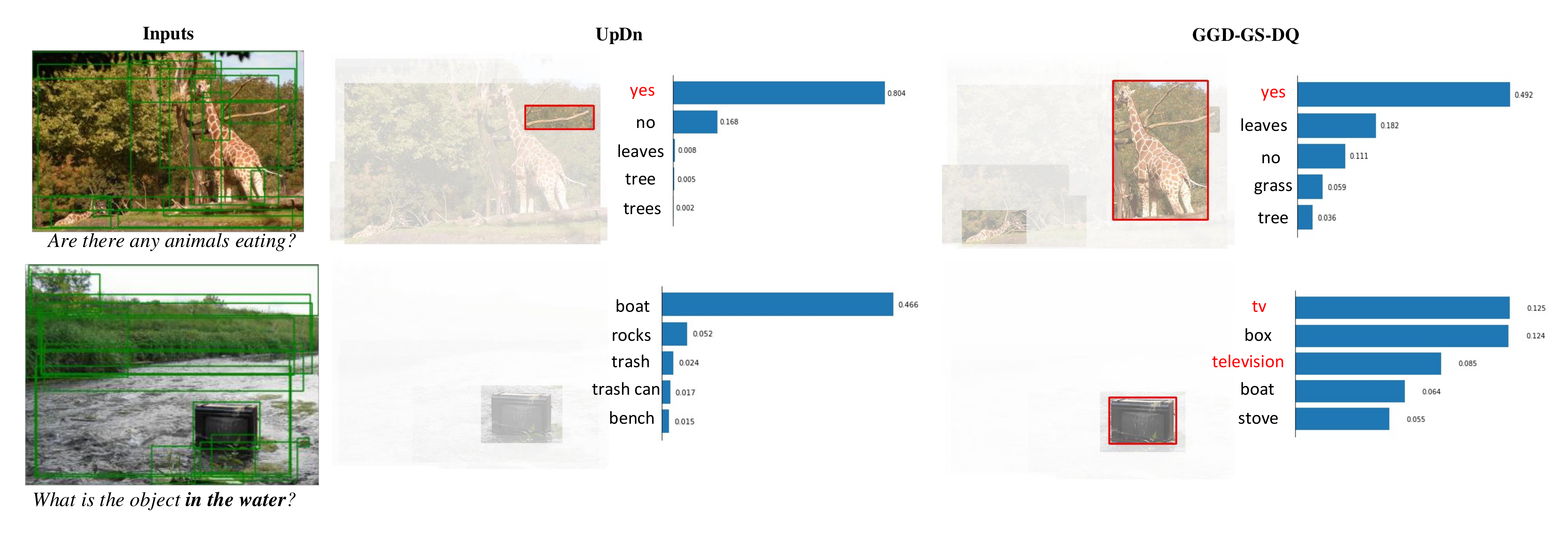}
	\end{center}
	\vspace{-2em}
	\caption{{\bf Qualitative Evaluation for GGD$_{gs}^{dq}$}. We provide a comparison between UpDn and GGD$_{gs}^{dq}$ on the visualization of the most sensitive regions and confidence of the top-5 answers. Red answers denote the ground-truth.}
	\label{fig:vis}
	\vspace{-1em}
\end{figure*}
\begin{figure*}[t]
	\centering
	\begin{center}
		\includegraphics[width=0.99\linewidth]{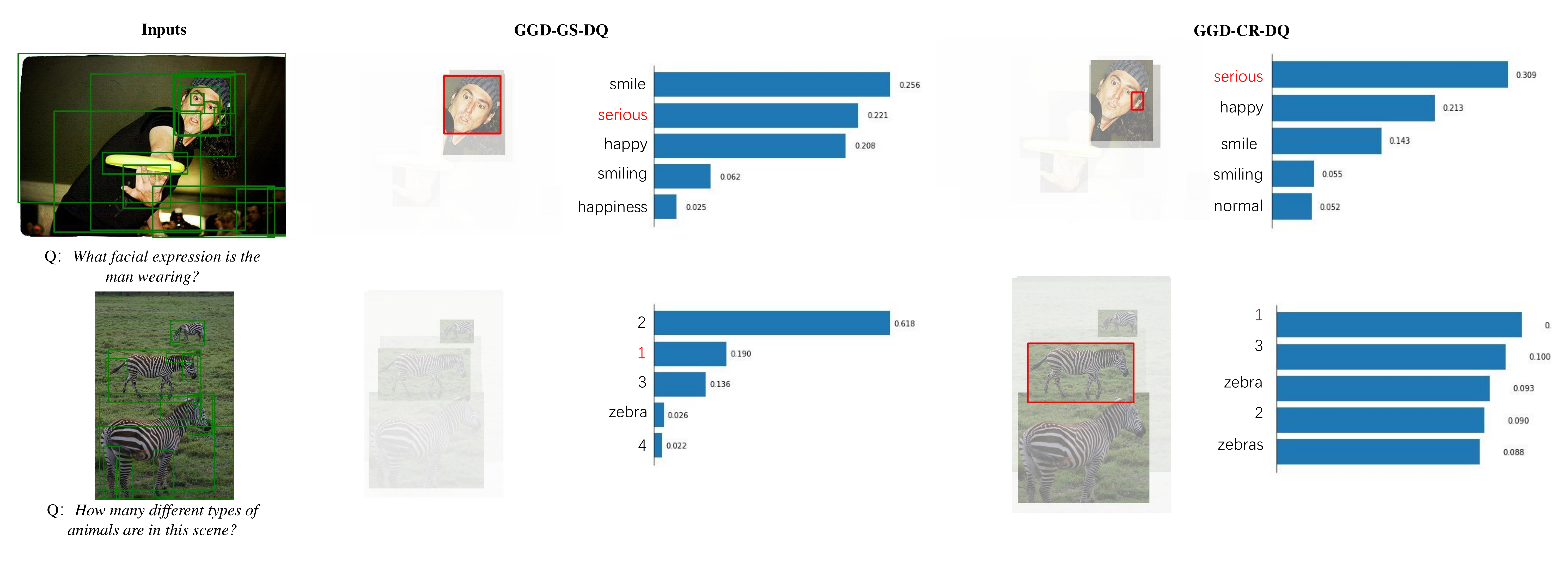}
	\end{center}
	\vspace{-2.4em}
	\caption{{\bf Comparison between GGD$_{gs}^{dq}$ and GGD$_{cr}^{dq}$}. The major improvements are reflected on counting problems and questions that rarely appear in the training data.}
	\label{fig:iter_cr}
	\vspace{-1em}
\end{figure*}
\begin{figure*}[t]
	\centering
	\begin{center}
		\includegraphics[width=0.99\linewidth]{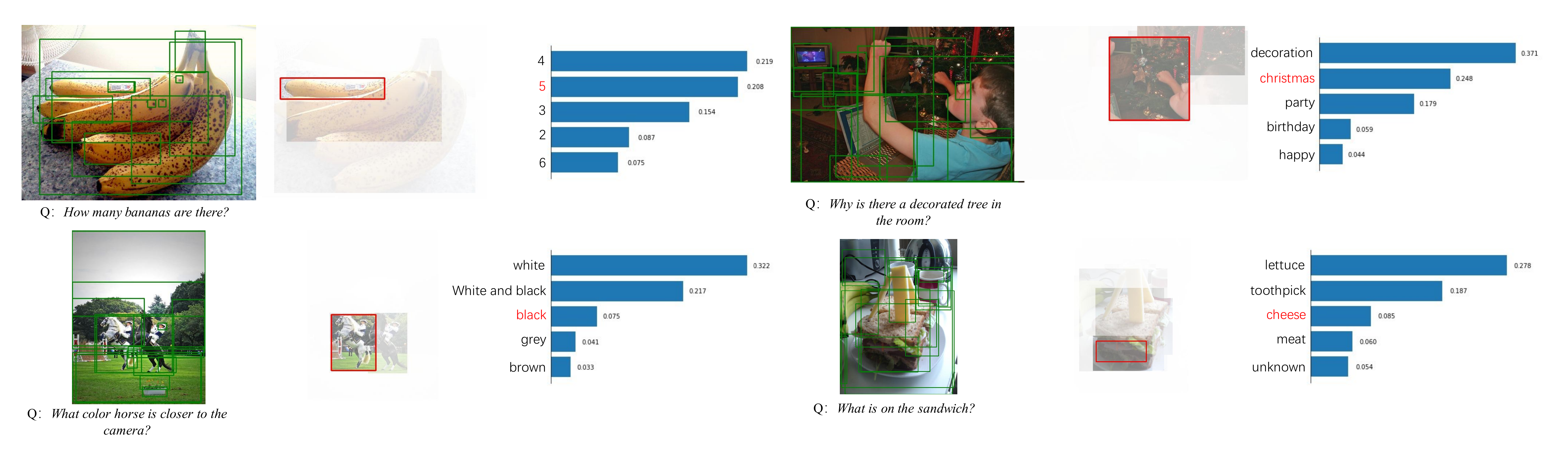}
	\end{center}
	\vspace{-2.4em}
	\caption{{\bf Failure cases for GGD$_{cr}^{dq}$}. From top to bottom, the failure cases are 1) counting problems;  2) Synonym answers; 3) incorrect visual evidences.}
	\label{fig:fail}
	\vspace{-1em}
\end{figure*}

\noindent {\bf Comparison with State-of-the-art Methods}.
We compare our best performed model GGD$_{cr}^{dq}$ with existing state-of-the-art bias reduction techniques, including visual-grounding-based methods HINT~\cite{2019hint}, SCR~\cite{2019SCR}, ensemble-based methods AdvReg.~\cite{2018overcoming}, RUBi~\cite{2019rubi}, LM (LMH)~\cite{2019don}, MFE~\cite{2020mfe}, new-question-encoding-based methods GVQE~\cite{2020reducing}, DLP~\cite{2020cf-vqa}, counterfactual-based methods CF-VQA~\cite{2020cf-vqa}, CSS~\cite{2020counterfactual}
, recent proposed regularization method MFE~\cite{2020mfe}, and SAR~\cite{2021SAR} that models VQA as Visual Entailment with pre-trained model. 

As shown in TABLE~\ref{tab:sota}, GGD$_{cr}^{dq}$ achieves state-of-the-art performance without extra bias annotation. It outperforms the baseline model UpDn by 20\% higher in terms of Accuracy and 10\% higher in terms of CGD, which verifies the effectiveness of GGD on both answer classification and visual-grounding ability.

For the comparison of question-type-wise results, incorporating GGD improves the performance for all the question types, especially the more challenging ``other" question type~\cite{teney2020value}. 
CF-VQA~\cite{2020cf-vqa} performs the best in Y/N, but worse than our methods in all the other question types and metrics. LMH~\cite{2019don}, LMH-MFE~\cite{2020mfe}, and LMH-CSS~\cite{2020counterfactual} work well on Num. questions. Comparing with LM and LMH, it is obvious that the performance gains in Num. are mainly due to the additional regularization for entropy. However, methods with entropy regularization drop nearly 10\% on VQA v2. This indicates that these models may over-correct the bias. SAR~\cite{2021SAR} achieves the best performance on both VQA-CP v2 and VQA v2, using extra datasets for model pre-training.
On the other hand, GGD$_{cr}$ improves both Num. Accuracy and in-distribution performance only with the Curriculum Regularization and work well without any extra data sources.

\noindent {\bf Qualitative Evaluation}.
Examples in Fig.~\ref{fig:vis} illustrate how GGD$_{gs}^{dq}$ makes difference compared with the baseline UpDn~\cite{2018bottomup}.
The first example is about using visual information for inference. Despite offering the right answer ``yes", the prediction from UpDn is not based on the right visual grounding result. In comparison, GGD$_{gs}$ correctly grounds the giraffe that is eating leaves. 
The second example is a case of reducing language prior apart from Yes/No questions. UpDn answers ``boat" just based on the language context ``in the water", while GGD$_{gs}^{dq}$ provides correct answers ``tv" and ``television" with more salient visual grounding. These examples qualitatively verify our improvement on both Accuracy and visual explanation for the predictions.

Fig.~\ref{fig:iter_cr} provides qualitative illustration for improvements achieved by GGD$_{cr}$. Compared with GGD$_{gs}$, GGD$_{cr}$ majorly improves on the ``Num." and ``Other" question types. Questions about facial expression and counting do not frequently appear in the dataset. If these questions can be correctly predicted by fitting the biases, the base model in GGD$_{gs}$ may not have enough data to learn a good representation. 

Fig.~\ref{fig:fail} shows the examples of failure cases from GGD$_{cr}^{dq}$. The model appears to be weak on the complicated counting problem. Some failure cases are due to missing annotation in the dataset (``decoration" can also be regarded as the right answers). 
Although making wrong predictions, answers for failure cases in the last row are still consistent with visual explanations rather than language bias, which further indicates that GGD$^{dq}$ truly makes use of the visual information.


\begin{table}[]
		\centering
		\renewcommand{\arraystretch}{1.4}
		\setlength{\tabcolsep}{2.6mm}
		\caption{Experimental results on GQA-OOD test-dev. $g$ denotes the method addressing global group distribution bias, $l$ is method addressing the local group distribution bias, and $q$ is the method addressing the question shortcut bias. ``Avg" is the mean accuracy of head and tail groups. Methods with only Avg are reported from \cite{2021investigation}.} 
		\begin{tabular}{l|cccc}
			\hline
			Method & All & Head & Tail & Avg \\ \hline
			UpDn~\cite{2018bottomup}   & 46.93\tinypm{0.29} & 49.41\tinypm{0.26} & 42.73\tinypm{0.57} & 46.07\tinypm{0.42}     \\
			LfF~\cite{2020lff}         & 47.06\tinypm{0.23} & 49.53\tinypm{0.63} & 42.99\tinypm{0.47} & 46.26\tinypm{0.09}     \\
			SD~\cite{2020GS}           & 47.59\tinypm{0.33} & 50.05\tinypm{0.32} & 44.49\tinypm{1.15} & 47.27\tinypm{0.42}     \\
			\hline
			RUBi$^q$~\cite{2019rubi}     & 45.51\tinypm{0.18} & 47.87\tinypm{0.02} & 41.67\tinypm{0.47} & 44.71\tinypm{0.23}     \\
			RUBi$^g$~\cite{2019rubi}     & 7.15\tinypm{0.53}  & 6.86\tinypm{0.91}  & 6.60\tinypm{1.22}  & 6.573\tinypm{1.02}      \\
			RUBi$^l$~\cite{2019rubi}     & 20.48\tinypm{3.84} & 22.30\tinypm{4.04} & 17.51\tinypm{3.67} & 19.91\tinypm{3.79}      \\
			Up Wt$^g$~\cite{2020overpara}& - & - & - & 26.4      \\
			Up Wt$^l$~\cite{2020overpara}& - & - & - & 26.2      \\
			LNL$^g$~\cite{2019lnl}       & - & - & - & 32.4      \\
			LNL$^l$~\cite{2019lnl}       & - & - & - & 10.7      \\
			\hline \hline
			GGD$_{gs}^{q}$                 & 47.41\tinypm{0.45} & 50.07\tinypm{1.19} & 43.09\tinypm{1.06} & 46.58\tinypm{0.31}      \\
			GGD$_{gs}^{g}$                 & 48.96\tinypm{0.08} & \textbf{52.07\tinypm{0.12}} & 44.00\tinypm{0.27} & 48.03\tinypm{0.13}       \\
			GGD$_{gs}^{l}$                 & 47.84\tinypm{0.49} & 50.36\tinypm{0.60} & 43.74\tinypm{0.33} & 47.05\tinypm{0.45}      \\
			GGD$_{gs}^{gq}$                & 47.15\tinypm{0.29} & 49.62\tinypm{0.43} & 43.27\tinypm{0.89} & 47.01\tinypm{0.47}      \\
			GGD$_{gs}^{lq}$                & 48.25\tinypm{0.53} & 50.74\tinypm{0.99} & 44.18\tinypm{0.24} & 47.46\tinypm{0.39}      \\
			\hline
			GGD$_{cr}^{q}$                 & 47.87\tinypm{0.59} & 50.19\tinypm{0.63} & 44.25\tinypm{0.31} & 47.22\tinypm{0.47}      \\
			GGD$_{cr}^{g}$                 & 48.09\tinypm{0.35} & 51.27\tinypm{0.63} & 43.25\tinypm{0.35} & 47.26\tinypm{0.44}      \\
			GGD$_{cr}^{l}$                 & 48.01\tinypm{0.60} & 51.23\tinypm{0.87} & 42.74\tinypm{0.17} & 46.99\tinypm{0.51}      \\
			GGD$_{cr}^{gq}$                & \textbf{49.21\tinypm{0.08}} & 52.01\tinypm{0.30} & \textbf{44.67\tinypm{0.68}} & \textbf{48.34\tinypm{0.19}}      \\
			GGD$_{cr}^{lq}$                & 47.03\tinypm{0.52} & 49.37\tinypm{0.71} & 43.05\tinypm{0.62} & 46.21\tinypm{0.42}      \\
			\hline
			GGD$_{gs}^{se}$              & 47.00\tinypm{0.10} & 49.16\tinypm{1.44} & 42.61\tinypm{1.14} & 45.89\tinypm{0.37}      \\
			GGD$_{cr}^{se}$              & 47.50\tinypm{0.35} & 51.10\tinypm{0.39} & 42.06\tinypm{0.75} & 46.58\tinypm{0.41}      \\
			\hline
		\end{tabular}
		\label{tab:gqa}
\end{table}

\subsubsection{Experimental Results on GQA-OOD}\label{GQA}
{\bf Biases}. GQA-OOD is a more challenging dataset for Visual Question Answering. It has biases from multiple sources including imbalanced answer distribution, visual concept co-occurrences, question word correlations, and question type/answer distribution. Since the training set for GQA-OOD is the manually balanced GQA-balanced-train split, it is hard to specify the explicit biases to ensure that the models can generalize to even the rarest local groups. Following \cite{2021investigation}, apart from question shortcut bias similar to that in VQA-CP~\cite{2018vqacp}, we define two kinds of distribution bias according to global group labels (115 groups) and local group labels (133328 groups), and the corresponding models are denoted as GGD$^{g}$ and GGD$^{l}$.

\noindent{\bf Comparison with State-of-the-art Methods}. We compare GGD$_{gs}$ with implicit de-bias methods LfF~\cite{2020lff} and SD~\cite{2020GS}; explicit de-bias methods RUBi~\cite{2019rubi}, Up Wt~\cite{2020overpara}, and LNL~\cite{2019lnl}. 
As shown in TABLE~\ref{tab:gqa}, all three previous explicit methods fail on both global and local group distribution bias, performing even worse than the original baseline UpDn. RUBi~\cite{2019rubi} with the question-only branch also degrades on GQA-OOD compared with the baseline. Implicit methods LfF~\cite{2020lff} and SD~\cite{2020GS} are more stable on handling complicated biases in GQA. SD~\cite{2020GS} achieves the highest accuracy on the Tailed group. This indicates that both distribution bias and shortcut bias in GQA-OOD are not as obvious as those in VQA-CP, since the data from GQA is synthetic and the train split has been manually balanced. 

On the other hand, GGD works well with hard-example mining. It surpasses the baseline under all bias settings and is comparable to existing implicit methods. If the biased models can not make a prediction with high confidence, the pseudo labels for the base model remain almost unchanged for most of the data. Although the biased models cannot well model the biases in the dataset, GGD will not harm the performance like previous explicit de-bias methods.

\noindent{\bf Ablation Study}. According to the ablation studies, GGD$^g$ addressing distribution bias on global groups works better compared with GGD$^l$ on the local groups. Sequentially reducing the distribution bias and the shortcut bias will improve the Tail group Accuracy but slightly degrades the Head group accuracy.
GGD$_{cr}^{gq}$ achieves the highest overall accuracy because it alleviates the over-estimated bias in GGD$_{gs}^{gq}$. However, since the biased models do not capture biases with high confidence, GGD$_{cr}$ does not show much difference compared with GGD$_{gs}$ under most of the settings.

\begin{figure}[t]
		\centering
		\includegraphics[width=0.9\linewidth]{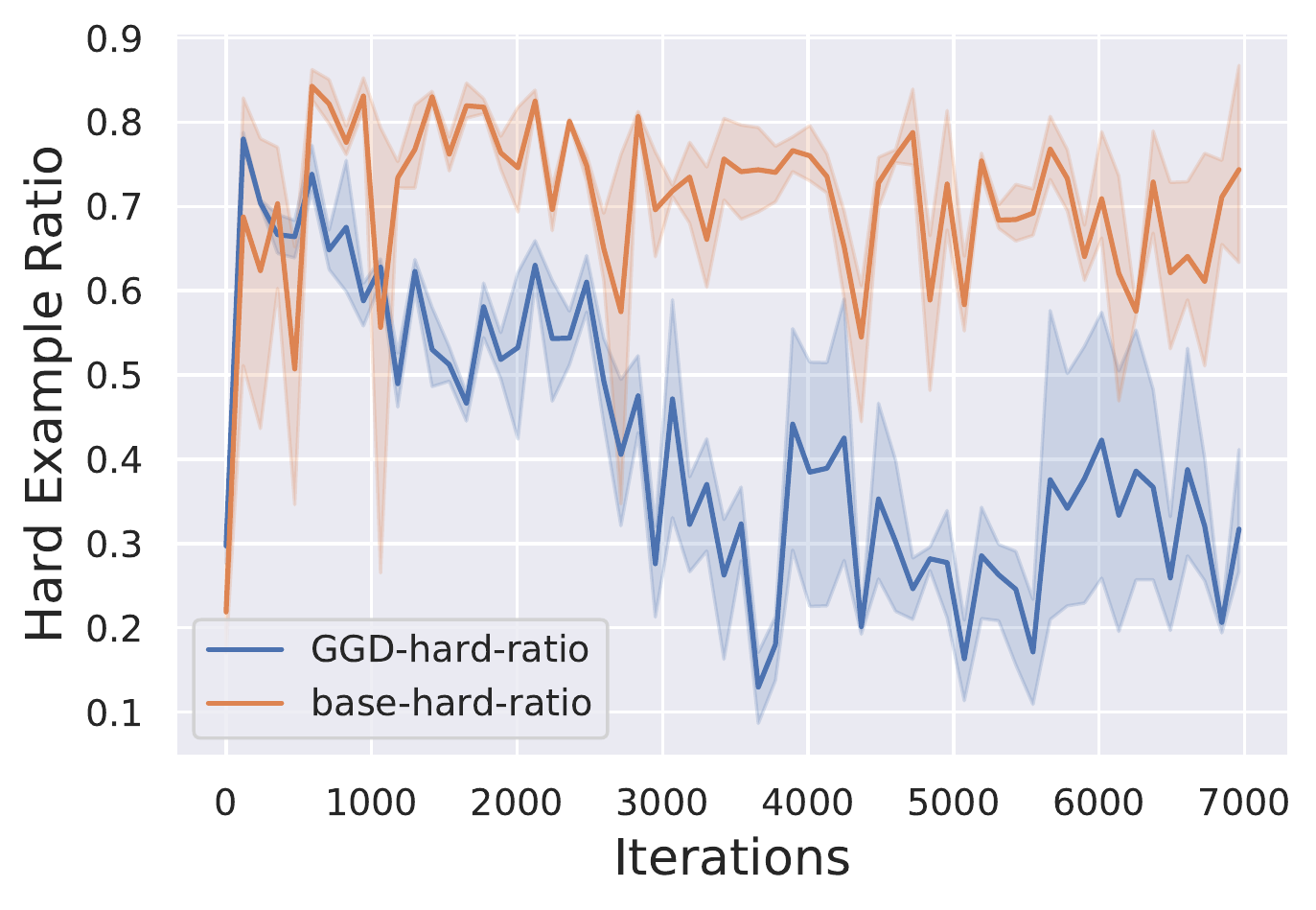}
		\vspace{-1.5em}
		\caption{The loss ratio of the hard examples from baseline model and GGD$_{cr}$ base model.}
		\label{fig:hard}
	\vspace{-1em}
\end{figure}

\begin{figure}[t]
	\begin{center}	
		\includegraphics[width=0.95\linewidth]{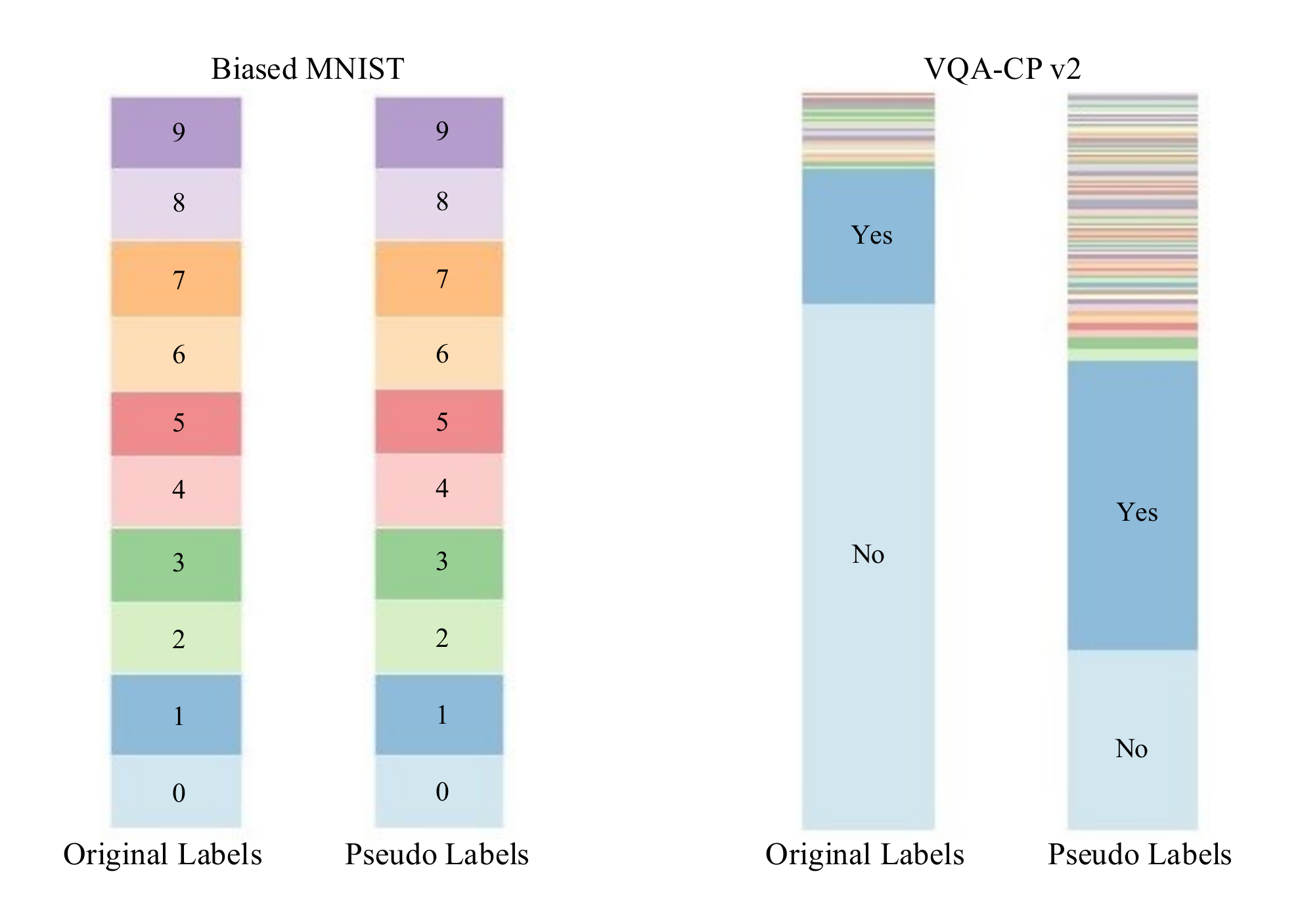}
	\end{center}
	\vspace{-2em}
	\caption{The original Labels and Pseudo Labels from GGD$_{gs}$. The left figure is the label changes in the Biased MNIST training set, while the right figure is that of ``is this" question type in VQA-CP v2.}
	\label{fig:dist_ref}
\end{figure}

\subsection{Discussion \label{discuss}}

{\bf Hard Example Mining Mechanism}. To demonstrate the hard example mining mechanism of GGD, we provide analysis on the training process. 
We first evaluate whether the base model focuses on examples. We define the hard ratio $R_h$ as
\begin{equation}\label{hard_ratio}
R_h = \frac{L_{hard}}{L_{all}},
\end{equation}
where $L_{hard}$ is the loss of hard examples ({\it i.e.}, samples that do not choose corresponding background color), $L_{all}$ is the loss for all samples. We calculate the accumulated loss for every 118 iterations of 4 repeated experiments with different random seeds. For more obvious comparison, experiments are done on GGD$_{gs}^{1k}$ ($\lambda_t=1$) on Biased MNIST with $\rho_{\text{train}} = 0.998$. As shown in Fig.~\ref{fig:hard}, although the proportion of hard examples is no more than 1\%, $R_h$ from the baseline is always over 0.6 and hardly decreases. The loss is trapped in local minimum due to the low \emph{average} training error dominated by the majority groups of data. On the other hand, $R_h$ from the base model of GGD is always lower than the baseline in any observed iteration and continuously decreases along with the training process. This reflects that GGD can well handle the hard examples compared with vanilla ResNet-18. 

{\bf Bias Over-estimation}. An interesting phenomenon is that GGD$_{gs}$ degrades on VQA v2~\cite{2015vqa}, but is relatively stable on the in-distribution test of Biased MNIST~\cite{2020rebias}. We find that this is due to the distribution bias on VQA v2. As shown in Fig.~\ref{fig:dist_ref}, taking ``is this" question type as an example, GGD$_{gs}$ will amplify the distribution bias and result in an ``inverse biased distribution". On the other hand, the pseudo labels for Biased MNIST are still balanced because the texture bias is independent of the label distribution. This can also partially explain the improvement of GGD$_{cr}$ on VQA, where the Curriculum Regularization also reduces such ``bias over-estimation", apart from better low-level representation learning ability. In practice, one can get a more balanced classifier by selecting better $\lambda_t$ according to the bias level of the dataset. It can also be a valuable research to adaptively estimate the bias level of a dataset in the future.

{\bf Limitations}. 
Although the bias over-estimation problem in our previous GGE model has been alleviated with the Curriculum Regularization, there is still two major shortcomings of GGD. 
First, the hard-example mining mechanism in GGD is an instance-level sample re-weighting. 
If all samples are following a certain spurious correlations, GGD will fail to discover it as a spurious correlation ({\it e.g.}, $\rho_{\text{train}} = 1.0$ in Biased-MNIST). The gradients from the biased models will decline to 0. Even though the spurious feature is identified with the greedily learned biased models, the base model cannot learn a de-biased feature accordingly. Eq.\ref{prob} indicates that we can also directly optimize $\log p(y|x^b)$ with the biased feature $x^b$, which can be obtained from the optimal biased model $h(x^b; \phi^*)$. We will investigate how to select network activations according to the bias models towards feature-level ensemble in the future.

Second, if the biased model can well capture the biases in the dataset, GGD will largely improve both the in-distribution and the out-of-distribution performance. However, when the biased model can not perfectly disentangle the spurious correlations, the improvement from GGD is limited (see experiments on Adversarial SQuAD and GQA-OOD). Although the Self-Ensemble fashion GGD$^{se}$ can implicitly model the biases, it largely relies on the bias level of the dataset and the existence of hard examples in datasets. 
It can be a future work to design a more robust strategy that can capture spurious correlations needless of prior knowledge.

\section{Conclusion}
In this paper, we propose General Greedy De-bias Learning, a general de-bias framework with flexible regularization and wide applicability.
Accompanied with Curriculum Regularization, the relaxed GGD$_{cr}$ comes to a good trade-off between in-distribution and out-of-distribution performance. Experiments on image classification, Adversarial QA, and VQA demonstrate the effectiveness of GGD under both task-specific biased models and self-ensemble fashion without prior knowledge on both ID and OOD scenarios.

In theory, the core of our method is the greedy strategy, which sequentially learns biased models in priority. One can also replace the regularization with better metrics that are able to measure the distance between the predictions and the labels. It may further improve the performance on specific tasks. In the future, we will try de-bias learning at the feature level and design a better strategy to capture spurious correlations needless of dataset-specific knowledge.

%% file: appendix.tex
\section{Probabilistic Justification}
We consider the distribution $p(c|x)$:
\begin{equation}\label{prob2}
\begin{split}
p(y|x) & = p(y| x^b, x^{-b}) \\
& \propto p(x^{-b} | y, x^b) p(y | x^b)  \ \ \triangleright \text{Bayes Rule} \\
& = p(x^{-b} | y) p(y|x^b) \ \ \triangleright \text{Conditionally Independent} \\
& = p(y | x^b) \frac{p(y|x^{-b}) p(x^{-b})}{p(y)}  \ \ \triangleright \text{Bayes Rule}\\
& \propto \frac{p(y|x^{b})}{p(y)} p(y|x^{-b})
\end{split}
\end{equation}
Rearranging the log-likelihood of Eq.\ref{prob2} will lead to the probabilistic justification in Section 3.4.2. $p(y|x)$ is more likely to align with $p(y|x^{-b})$ in bias-conflicting samples. Moreover, it shows that the distribution $p(y)$ also has effects on the optimization of $p(y|x^{-b})$. This can partially explain the influence from the distribution bias.

\begin{figure*}[t]
	\begin{center}
		\subfigure[Baseline]{
			\label{subfig:baseline}
			\includegraphics[width=0.38\linewidth]{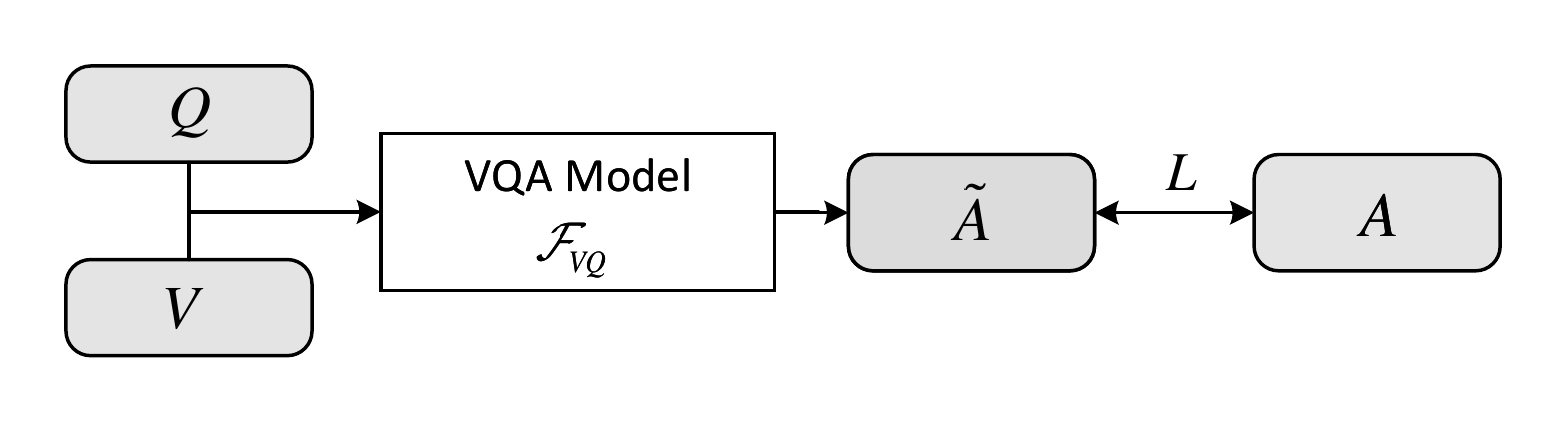}
		}
		\hspace{2em}
		\subfigure[GGD$^{d}$]{
			\label{subfig:dist_bias}
			\includegraphics[width=0.4\linewidth]{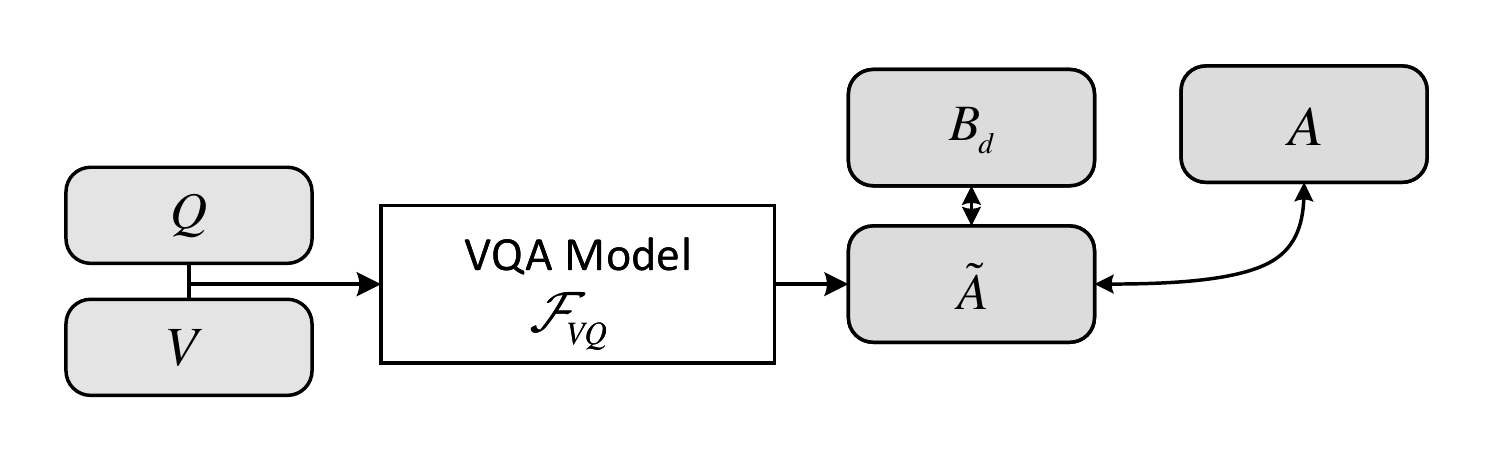}
		}\\
		\subfigure[GGD$^{q}$]{
			\label{subfig:q_bias}
			\includegraphics[width=0.4\linewidth]{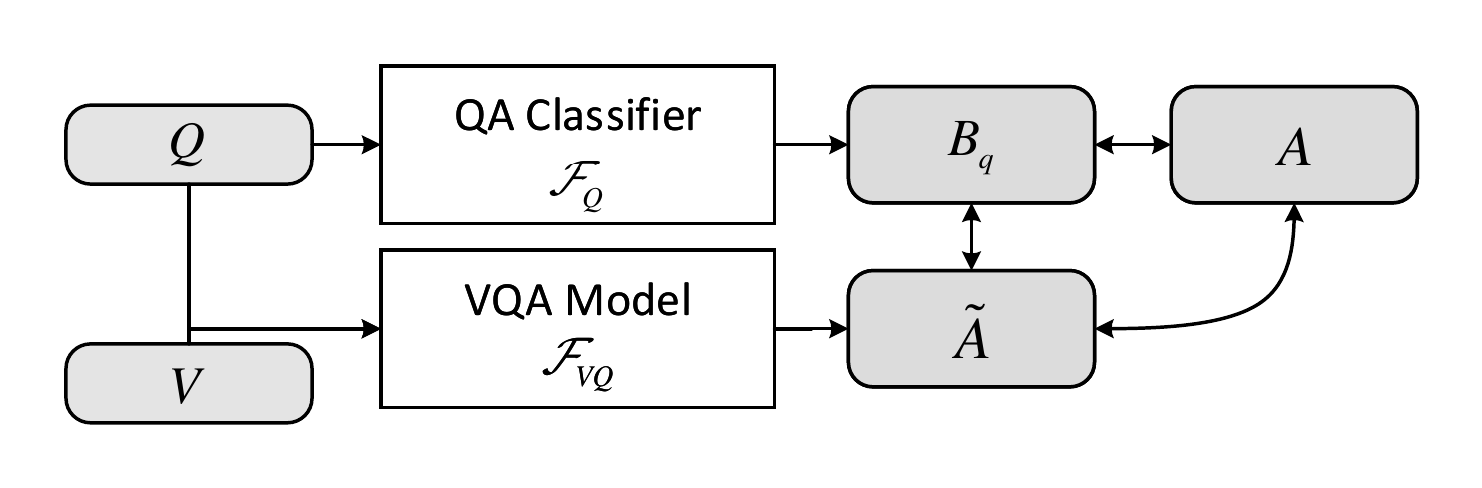}
		}
		\hspace{2em}
		\subfigure[GGD$^{dq}$]{
			\label{subfig:dist_q_bias}
			\includegraphics[width=0.4\linewidth]{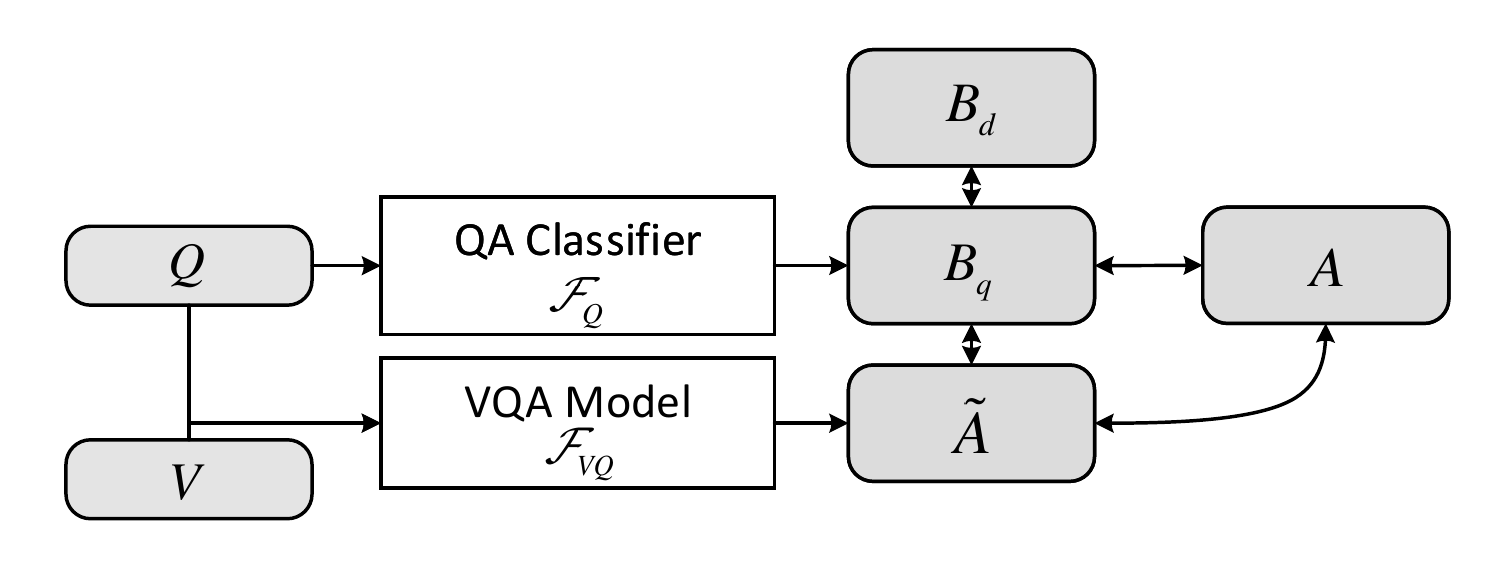}
		}\\
		\subfigure[GGD$^{se}$]{
			\label{subfig:se}
			\includegraphics[width=0.4\linewidth]{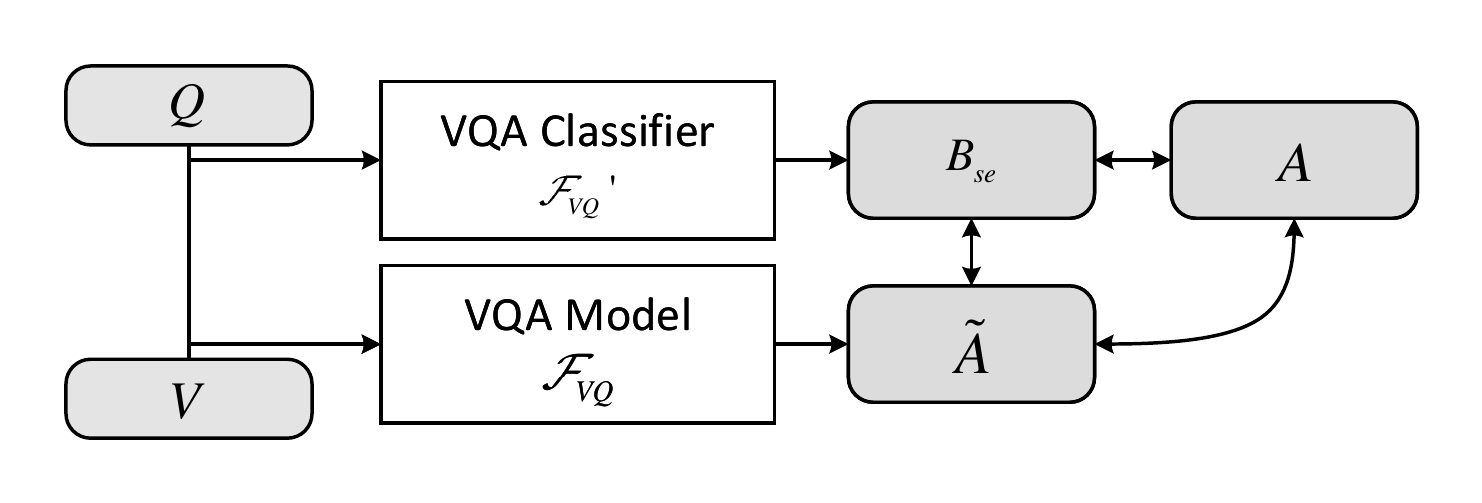}
		}
		\hspace{2em}
		\subfigure[GGD$^{dse}$]{
			\label{subfig:dse}
			\includegraphics[width=0.4\linewidth]{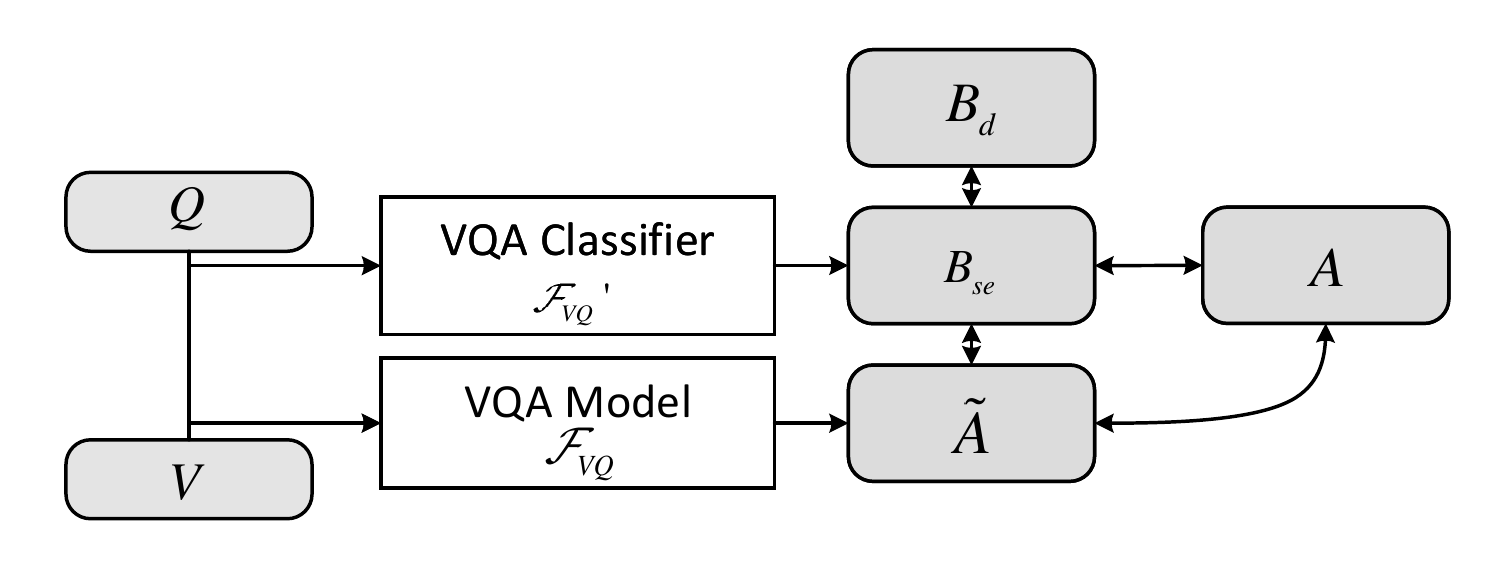}
		}
	\end{center}
	\vspace{-1em}
	\caption{{\bf Different versions of GGD for VQA}. $V,Q$ and $\tilde{A}$ denote image, question, and answer prediction respectively. $A$ is the human-annotated labels. $B_d$ and $B_q$ indicate the prediction from distribution bias and question shortcut bias respectively.} 
	\label{fig:model_vqa}
\end{figure*}

\section{Implementation Details}

\subsection{SimpleNet-1k \label{simnet}}
SimpleNet-1k is a fully convolutional CNN proposed in \cite{2020rebias}. It contains four convolutional layers with $1\times 1$ kernels and output channel \{16, 32, 64 ,128\}. Each convolutional layer is followed by batch normalisation~\cite{2015bn} and ReLU. The classification layer is consist of a Global Average Pooling (GAP) layer following a ($128 \times 10$) linear projection. 

All models for Biased MNIST is trained with batch size of 256 and Adam optimizer. The initial learning rate is set to be 1e-3.

\begin{table*}[t]
	\centering
	\renewcommand{\arraystretch}{1.4}
	\setlength{\tabcolsep}{3.5mm}
	\caption{Ablations for base model SimpleNet-7k on Biased MNIST.}
	\label{tab:7k}
	\begin{tabular}{lccclccclccc}
		\hline
		\multirow{2}{*}{$\rho_{\text{train}}$/$\rho_{\text{test}}$} & \multicolumn{3}{c}{0.990} &  & \multicolumn{3}{c}{0.995} &  & \multicolumn{3}{c}{0.999} \\ \cline{2-4} \cline{6-8} \cline{10-12} 
		& 0     & 0.1    & 0.990    &  & 0     & 0.1    & 0.995    &  & 0     & 0.1    & 0.999    \\ \cline{1-4} \cline{6-8} 
		\hline 
		SimpleNet-7k                   & 83.52 & 85.33 & 99.71 &  & 61.03 & 54.31 & 99.51 &  & 1.01 & 9.960 & 99.86   \\
		ReBias~\cite{2020rebias}   & 86.39 & 88.15 & 99.81 &  & 78.17 & 81.32 & \textbf{99.86} &  & 25.17 & 33.58 & \textbf{99.88}   \\
		RUBi~\cite{2019rubi}    & 88.91 & 90.13 & 99.79 &  & 74.67 & 76.52 & 97.72 &  & 19.78 & 26.53 & 93.05   \\
		GGD$_{gs}$                & \textbf{94.24} & \textbf{94.88} & 98.84 &  & \textbf{87.08} & \textbf{88.32} & 97.48 &  & \textbf{57.45} & \textbf{60.79} & 93.66   \\
		GGD$_{cr}$                  & 93.28 & 94.26 & \textbf{99.79} &  & 79.95 & 81.09 & 99.03 &  & 42.92 & 48.73 & 99.81   \\
		\hline
	\end{tabular}
	
\end{table*}

\begin{table*}[t]
	\centering
	\renewcommand{\arraystretch}{1.4}
	\setlength{\tabcolsep}{4mm}
	\setlength{\tabcolsep}{4mm}
	\caption{{Ablations of base model BAN and S-MRL for VQA-CP v2}.}
	\label{tab:ban}
	\begin{tabular}{lccccllc}
		\hline
		\multirow{2}{*}{Method} & \multicolumn{7}{c}{VQA-CP test}             \\ \cline{2-8} 
		& All & Y/N & Num. & Others & $\uparrow$CGR & $\downarrow$CGW & $\uparrow$CGD \\ \hline
		S-MRL~\cite{2019rubi}   & 37.90 & 43.68 & 12.04 & 41.97 & 41.94 & 27.32 & 14.62  \\
		+GGD$_{gs}^{dq}$  & 54.03 & 79.66 & 20.77 & 46.72 & 38.10 & 22.42 & 15.68  \\ 
		+GGD$_{cr}^{dq}$    & 54.46 & 86.43 & 14.16 & 47.16 & 39.24 & 25.69 & 14.55  \\ 
		\hline
		BAN~\cite{2018BAN}   & 35.94 & 40.39 & 12.24 & 40.51 & 5.33 & 5.19 & 0.14  \\
		+GGD$_{gs}^{dq}$   & 50.75 & 74.56 & 20.59 & 46.54 & 20.87 & 16.85 & 4.98   \\
		+GGD$_{cr}^{dq}$   & 51.72 & 77.58 & 23.70 & 46.11 & 33.93 & 22.92 & 11.01   \\
		\hline
	\end{tabular}
\end{table*}

\begin{figure*}[t]
	\centering
	\subfigure[{Input}]{
		\begin{minipage}{0.16\linewidth}
			\includegraphics[width=0.9\linewidth]{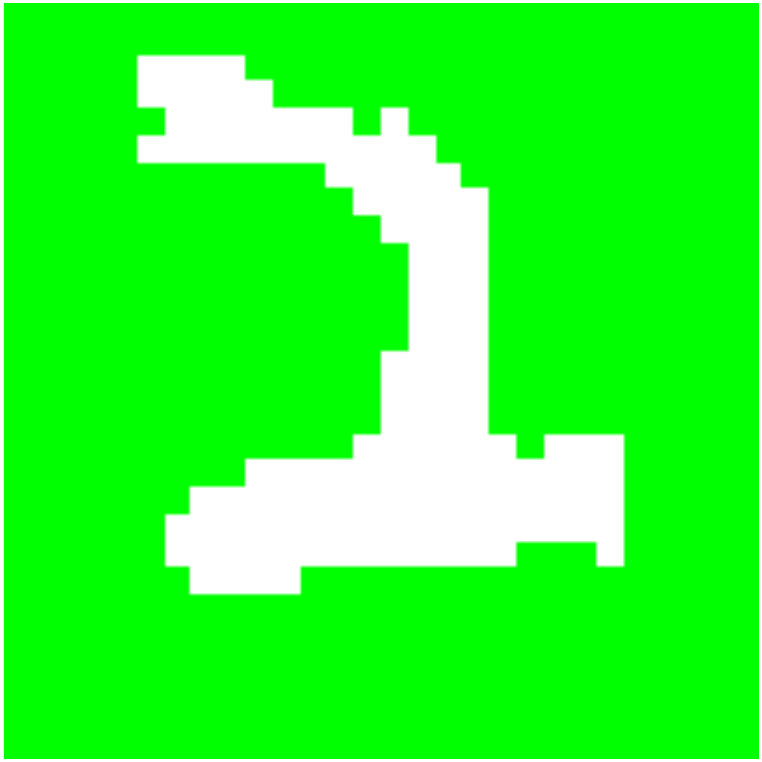}
			\includegraphics[width=0.9\linewidth]{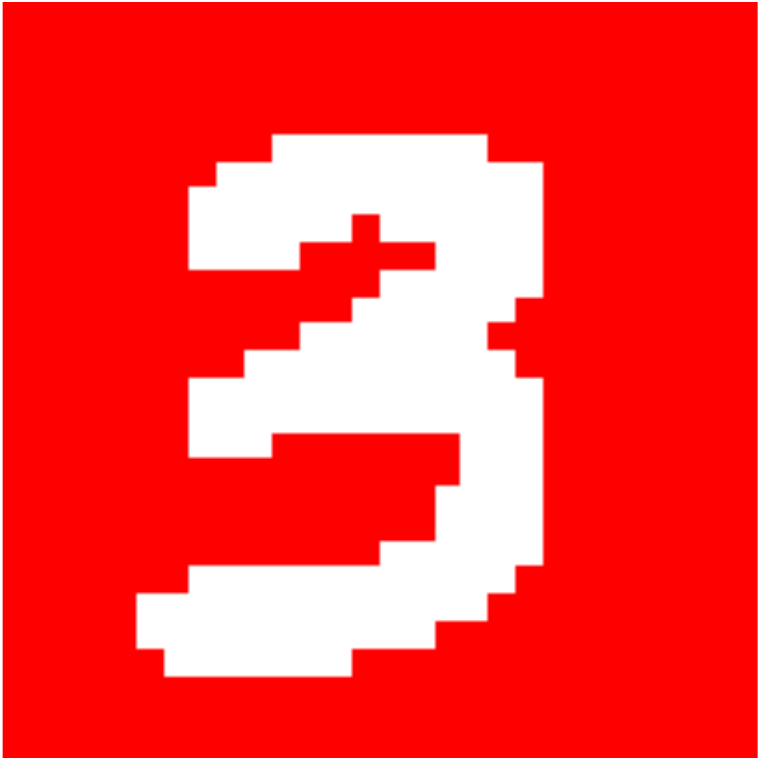}
			\includegraphics[width=0.9\linewidth]{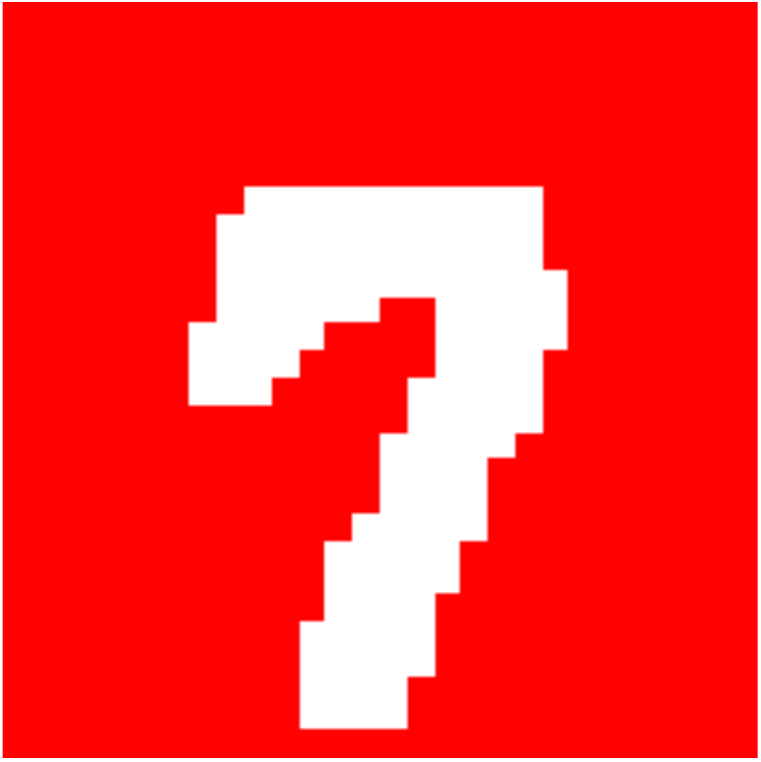}\vspace{1em}
		\end{minipage}	
	}
	\subfigure[{Vanilla ResNet-18}]{
		\begin{minipage}{0.16\linewidth}
			\includegraphics[width=0.9\linewidth]{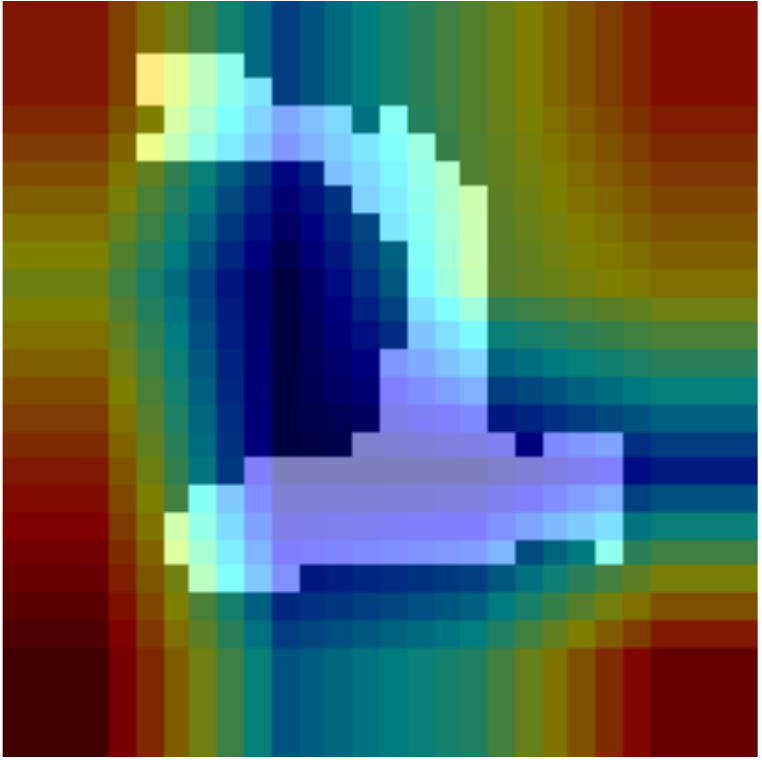}
			\includegraphics[width=0.9\linewidth]{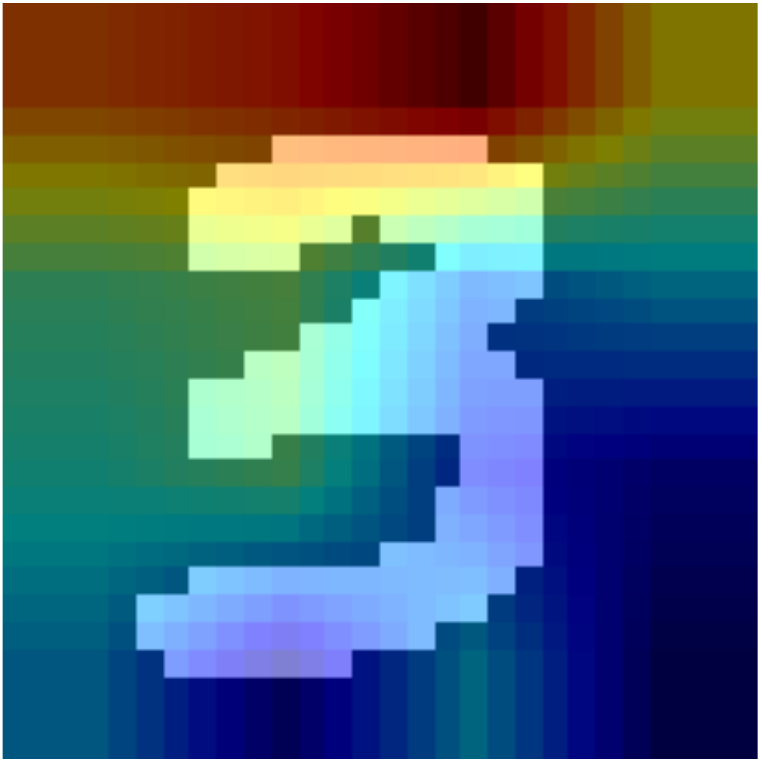}
			\includegraphics[width=0.9\linewidth]{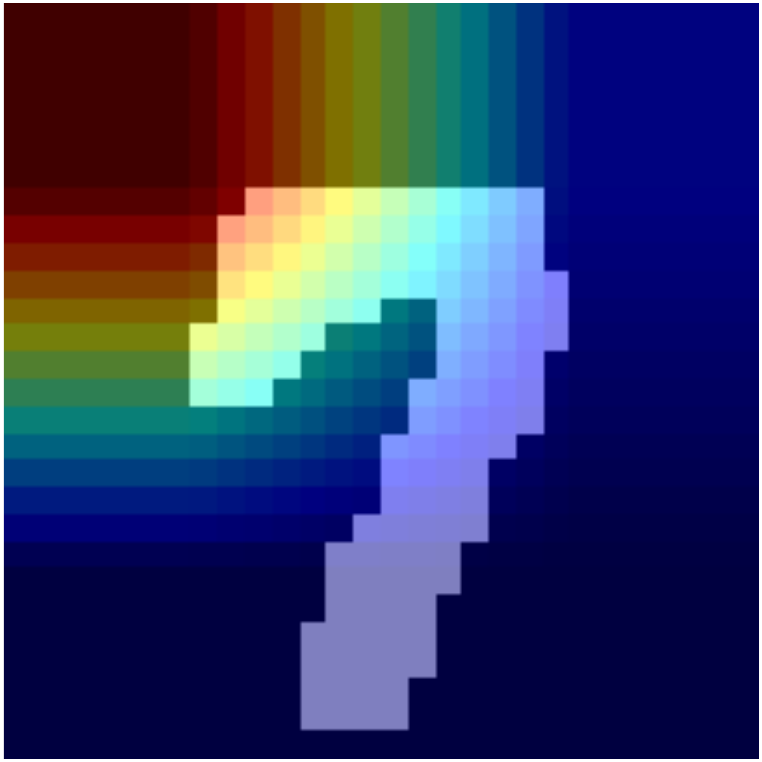}
			\vspace{1em}
		\end{minipage}	
	}
	\subfigure[SimpleNet-1k]{
		\begin{minipage}{0.16\linewidth}
			\includegraphics[width=0.9\linewidth]{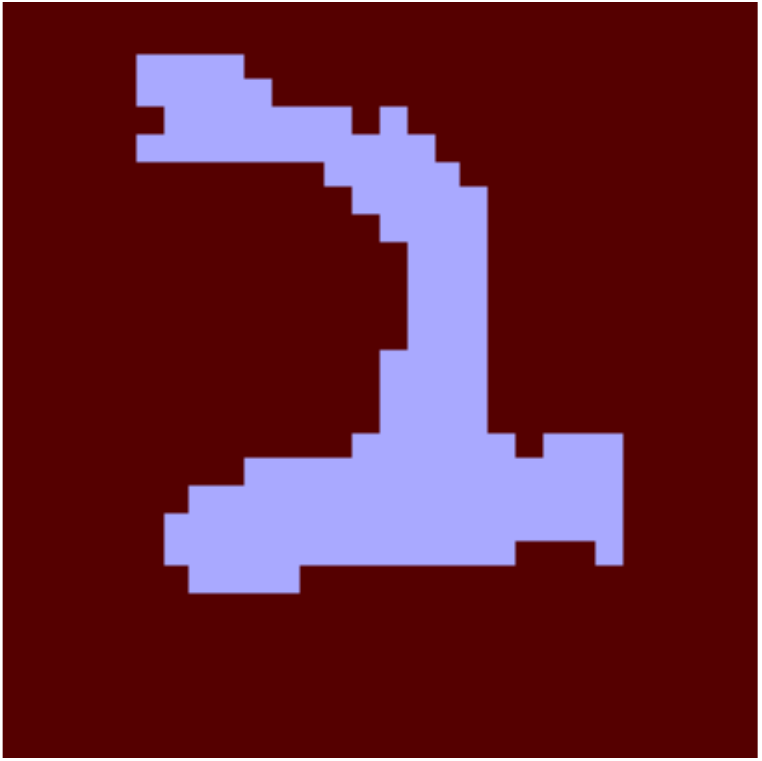}
			\includegraphics[width=0.9\linewidth]{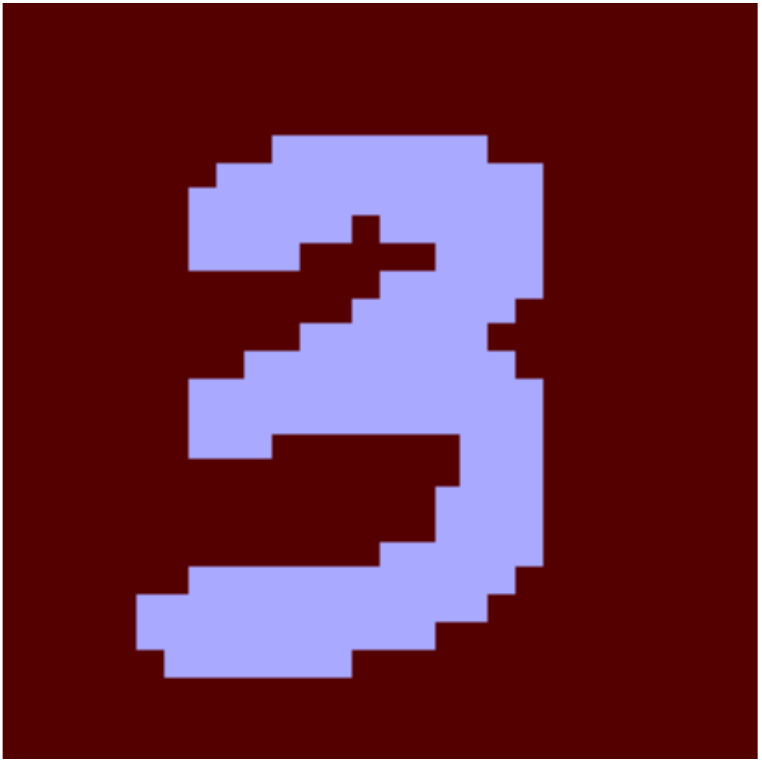}
			\includegraphics[width=0.9\linewidth]{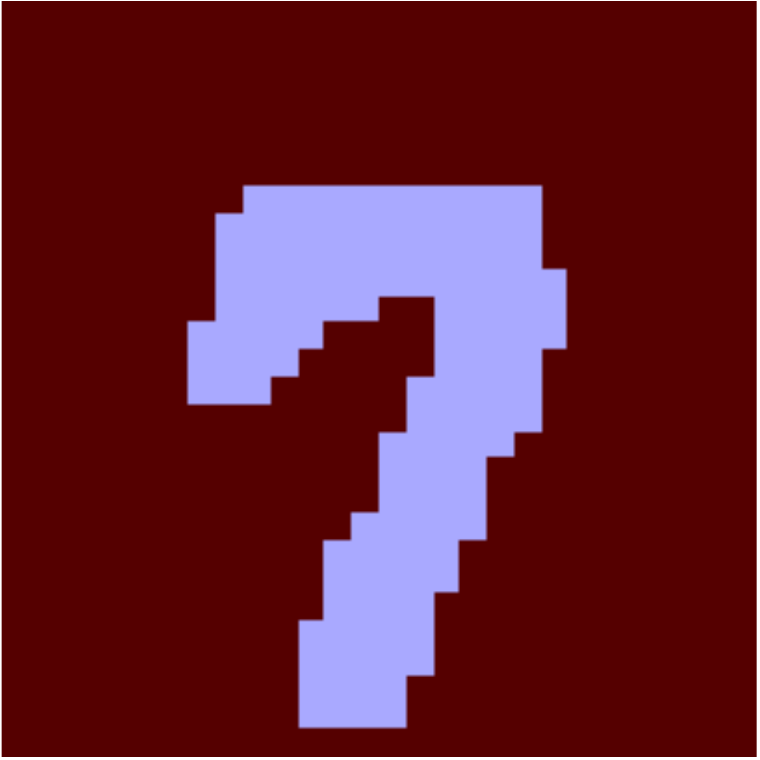}\vspace{1em}
		\end{minipage}	
	}
	\subfigure[GGD$_{gs}$ base model]{
		\begin{minipage}{0.16\linewidth}
			\includegraphics[width=0.9\linewidth]{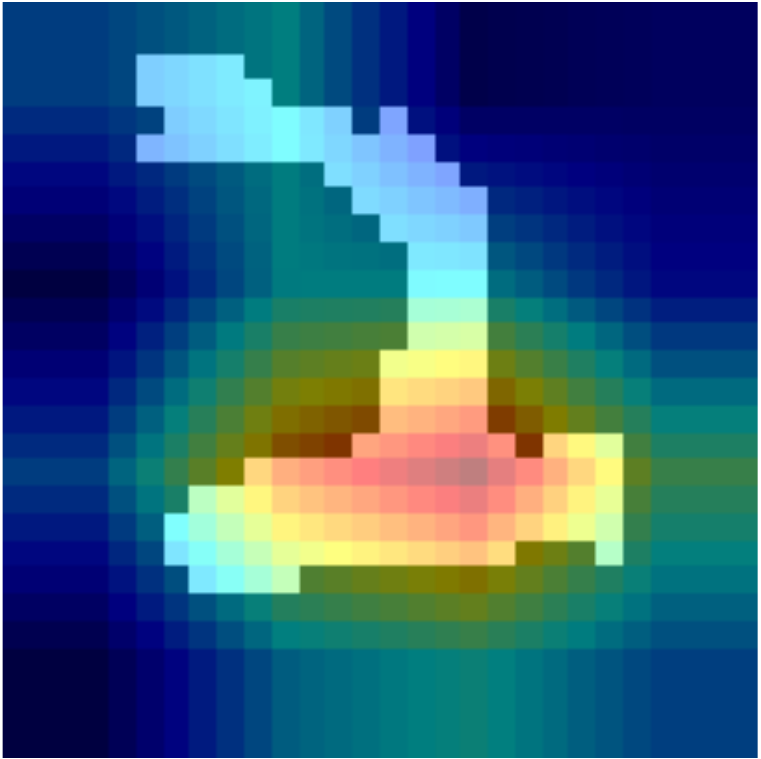}
			\includegraphics[width=0.9\linewidth]{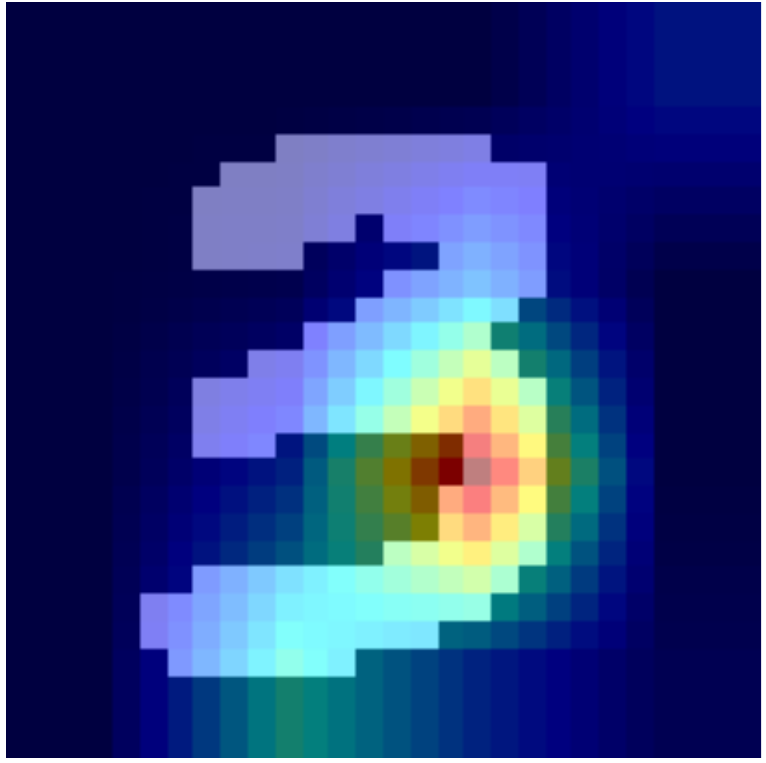}
			\includegraphics[width=0.9\linewidth]{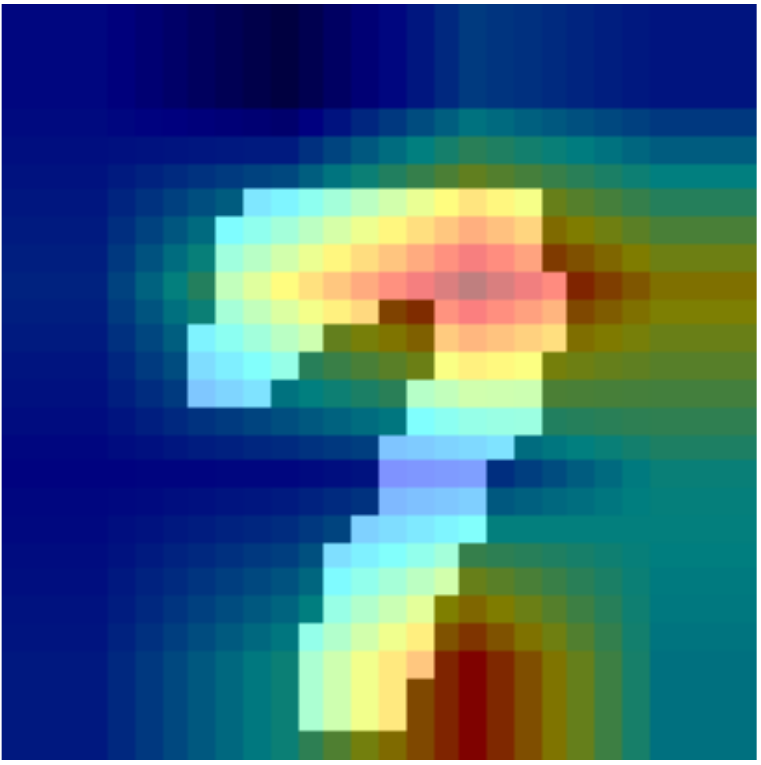}\vspace{1em}
		\end{minipage}	
	}
	\subfigure[GGD$_{cr}$ base model]{
		\begin{minipage}{0.16\linewidth}
			\includegraphics[width=0.9\linewidth]{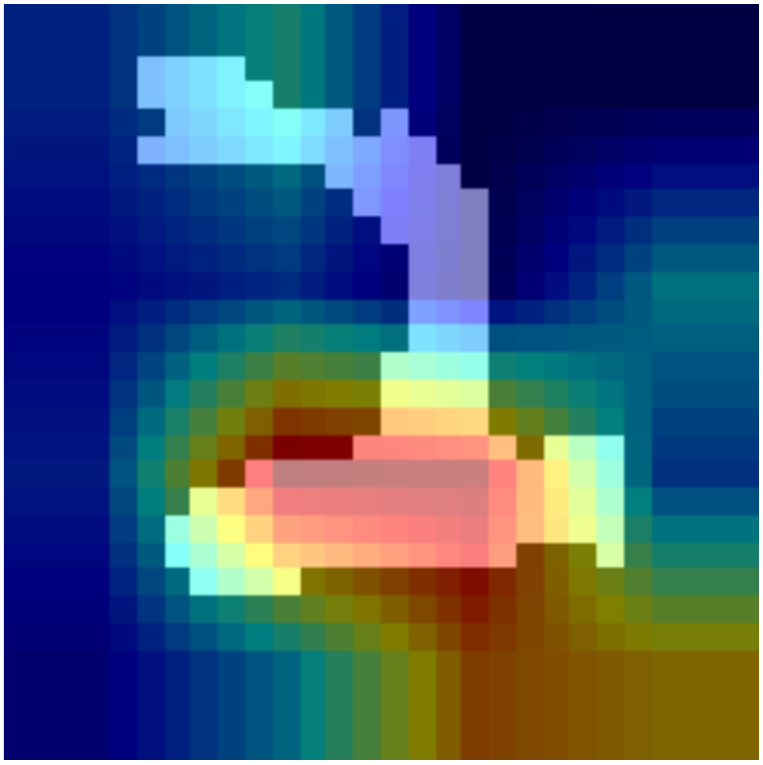}
			\includegraphics[width=0.9\linewidth]{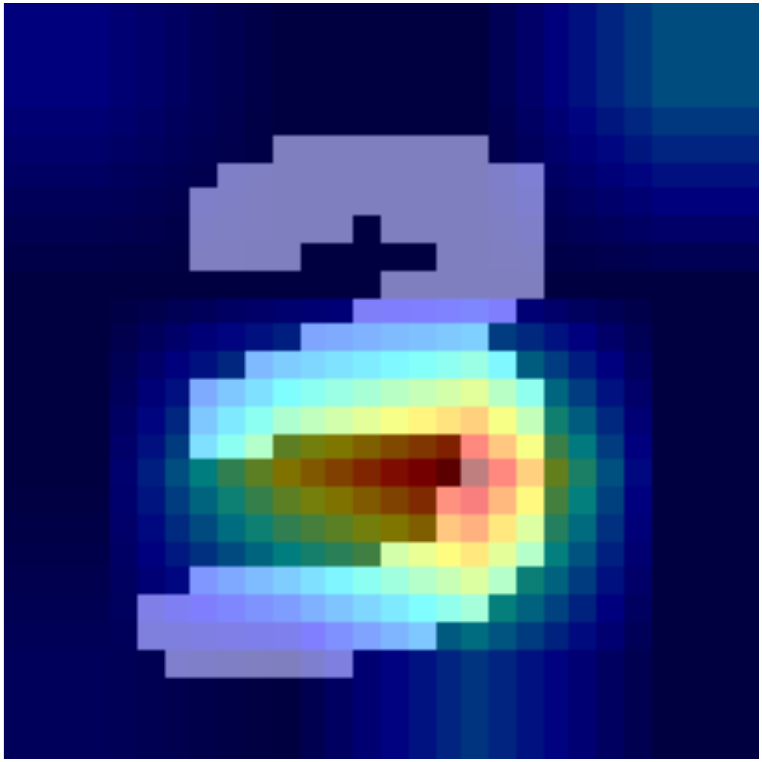}
			\includegraphics[width=0.9\linewidth]{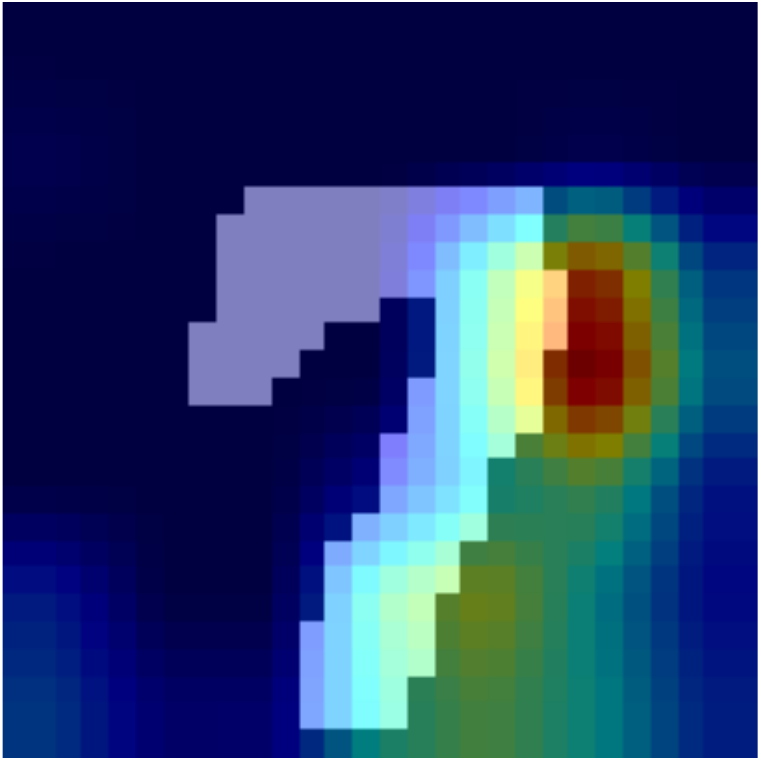}\vspace{1em}
		\end{minipage}
	}
	
	\caption{Saliency maps of the ResNet-18, bias model SimpleNet-1k, the base model from GGD$_{gs}$, and the base model from GGD$_{cr}$. The redder the pixel is, the more contributions it makes to prediction.
	}
	\label{fig:cam}
\end{figure*}

\subsection{Different versions of GGD for VQA}
The optimization paradigm for GGD$^{d}$, GGD$^{q}$, GGD$^{dq}$, GGD$^{se}$ and GGD$^{dse}$ are shown in Fig.~\ref{fig:model_vqa}.
$V,Q$ and $\tilde{A}$ denote images, questions, and answer predictions respectively. $A$ is the human-annotated labels. $B_d: \{\hat{y}^{d}_i \}_{i=1}^N$, $B_q: \{\hat{y}^{q}_i \}_{i=1}^N$ and $B_{se}: \{\hat{y}^{se}_i \}_{i=1}^N$ indicate the prediction from distribution bias, question shortcut bias and self-ensemble bias respectively.

{\bf GGD$^{d}$} only models distribution bias for ensemble. We define the distribution bias as answer distribution in the train set conditioned on question types:
\begin{equation}\label{disb}
\hat{y}^{d}_i = p(a_i | t_i),
\end{equation}
where $t_i$ denotes the type of question $q_i$. The reason for counting samples conditioned on question types is to maintain type information when reducing distribution bias. Question type information can only be obtained from the questions rather than the images, which does not belong to the language bias to be reduced. 

The regularization for the base model is 
\begin{equation}\label{gge_b}
L = \mathcal{L}\left(\tilde{A}, A  \right) - \lambda_t \mathcal{L}_{CE}\left(\tilde{A},  B_d \right),
\end{equation}
where $\tilde{A}$ is the predictions, and $A$ is the labelled answers.

{\bf GGD$^{q}$} only uses a question-only branch for shortcut bias. The shortcut bias is the semantic correlation between specific QA pairs. Similar to \cite{2019rubi}, we compose the question shortcut bias as a question-only branch
\begin{equation}\label{shortb}
\hat{y}^{q}_i = c_q \left(e_q(q_i)  \right),
\end{equation}
where $c_q: Q \rightarrow \mathbb{R}^C$.

We first optimize the question-only branch with labelled answers
\begin{equation}\label{gge_q}
L_1 = \mathcal{L}(B_q, A).
\end{equation}
The loss for base model is 
\begin{equation}\label{gge_q2}
L_2 = \mathcal{L}\left(\tilde{A}, A  \right) - \lambda_t \mathcal{L}_{CE}\left(\tilde{A}, B_q \right).
\end{equation}

{\bf GGD$^{dq}$} uses both distribution bias and question shortcut bias. The loss for $B_q$ is 
\begin{equation}\label{l_q}
L_1 = \mathcal{L}\left(B_q, A  \right) - \lambda_t \mathcal{L}_{CE}\left(B_q,  B_d \right).
\end{equation}
The loss for base model is
\begin{equation}\label{l_dq}
L_2 = \mathcal{L}\left(\tilde{A}, A  \right) - \lambda_t \mathcal{L}_{CE}\left(\tilde{A},  B_q + B_d \right).
\end{equation}
$L_1$ and $L_2$ are optimized iteratively.

{\bf GGD$^{se}$} takes the joint representation $r_i = m\left(e_v(v_i), e_q(q_i) \right)$ itself as the biased feature instead of predefined question-only branch in GGD$^{q}$, the biased prediction is 
\begin{equation}\label{sf}
\hat{y}^{se}_i = c_s \left(r_i  \right),
\end{equation}
where $c_s: r \rightarrow \mathbb{R}^C$ is the classifier of the biased model. 

We first optimize a baseline model with labelled answers
\begin{equation}\label{se_1}
L_1 = \mathcal{L}(B_{se}, A),
\end{equation}

The loss for base model is 
\begin{equation}\label{se_2}
L_2 = \mathcal{L}\left(\tilde{A}, A  \right) - \lambda_t \mathcal{L}_{CE}\left(\tilde{A}, B_{se} \right).
\end{equation}

{\bf GGD$^{dse}$} removes the distribution bias before Self-Ensemble, which is similar to GGD$^{dq}$
\begin{equation}\label{dse1}
L_1 = \mathcal{L}\left(\sigma(B_{se}), A  \right) - \lambda_t \mathcal{L}_{CE}\left(\sigma(B_{se}),  B_d \right).
\end{equation}
The loss for base model is
\begin{equation}\label{dse2}
L_2 = \mathcal{L}\left(\sigma(\tilde{A}), A  \right) - \lambda_t \mathcal{L}_{CE}\left(\sigma(\tilde{A}),  \sigma(B_{se}) + B_d \right).
\end{equation}
$L_1$ and $L_2$ are optimized iteratively.

\subsection{SUMB-DQ and LMH+RUBi}
{\bf SUM-DQ} directly sums up the outputs of biased models and the base model without greedy learning. The loss for the whole model is
\begin{equation}\label{sum_dq}
L = \mathcal{L}(B_d + B_q + \tilde{A}, A),
\end{equation}
where $B_d$ is the predicted distribution bias, $B_q$ is the predicted shortcut bias, $\tilde{A}$ is the predictions and $A$ is the labelled answers.

{\bf LMH+RUBi} combines LMH~\cite{2019don} and RUBi~\cite{2019rubi}. It reduces distribution bias with LMH and shortcut bias with RUBi. The loss for RUBi is written as
\begin{equation}\label{rubi}
L_{rubi}(\tilde{A}, A) = \mathcal{L}(\tilde{A} \odot \sigma(G_q), A) + \mathcal{L}(c_q(G_q), A ),
\end{equation}
where $G_q = g(e_q(q_i))$, $g(.): Q \rightarrow \mathbb{R}^C$. Combining with LMH, the prediction is composed as
\begin{equation}\label{lhm}
F(\tilde{A},B,M) = \log \tilde{A} + h(M)\log B,
\end{equation}
where $M$ and $B$ are the fused feature and the bias in LMH, $h(.): M \rightarrow \mathbb{R}^C$. The combined loss function is
\begin{equation}\label{lhm_rubi}
L = L_{rubi}(F(A,B,M), A) + wH(h(M)\log B),
\end{equation} 
where $H(.)$ is the entropy and $w$ is a hyper-parameter.

\begin{table*}[t]
	\centering
	\renewcommand{\arraystretch}{1.3}
	\setlength{\tabcolsep}{4mm}
	\setlength{\tabcolsep}{4mm}
	\caption{{\bf Top-1 Accuracy of ResNet-32 on CIFAR-10-LT under different imbalance settings}. }
	\label{tab:cifar10}
	\begin{tabular}{@{}lcccccccc@{}}
		\toprule
		Imbalance Factor  & 1     & 0.2   & 0.1   & 0.02  & 0.01  & Mean  & $\uparrow\overline{\Delta} $ & $\textbf{}\overline{\Delta}$\% \\ \midrule
		ResNet-32  & 92.03 & 86.28 & 83.30 & 79.83 & 66.61 & 81.61 & -           & -          \\
		\midrule
		GGD$_{gs}^{d}$  & 92.48 & 88.78 & 87.13 & 83.26 & 71.36 & 84.60 & 2.99        & 3.67\%     \\
		GGD$_{cr}^{d}$  & 92.66 & 88.77 & 87.39 & 81.98 & 70.77 & 84.31 & 2.70        & 3.31\%     \\ \midrule
		GGD$_{gs}^{se}$ & 76.61 & 59.85 & 60.26 & 54.53 & 44.68 & 59.19 & -22.42      & -27.48\%   \\
		GGD$_{cr}^{se}$ & 92.00 & 87.19 & 84.81 & 79.51 & 68.37 & 82.38 & 0.77        & 0.94\%     \\ \bottomrule
	\end{tabular}
\end{table*}

\begin{table*}[t]
	\centering
	\renewcommand{\arraystretch}{1.3}
	\setlength{\tabcolsep}{4mm}
	\caption{{\bf Top-1 Accuracy of ResNet-32 on CIFAR-100-LT under different imbalance settings}.  }
	\label{tab:cifar100}
	\begin{tabular}{@{}lcccccccc@{}}
		\toprule
		Imbalance Factor  & 1     & 0.2   & 0.1   & 0.02  & 0.01  & Mean  & $\uparrow\overline{\Delta} $ & $\uparrow\overline{\Delta}$\% \\ \midrule
		ResNet-32  & 66.91 & 51.71 & 42.88 & 31.74 & 27.85 & 44.22 &   -    &    -      \\
		\midrule
		GGD$_{gs}^{d}$  & 67.55 & 59.44 & 49.03 & 35.88 & 31.09 & 48.60 & 4.38   & 9.91\%   \\
		GGD$_{cr}^{d}$  & 67.13 & 54.77 & 45.52 & 34.85 & 31.03 & 46.66 & 2.44   & 5.52\%   \\
		\midrule
		GGD$_{gs}^{se}$ & 48.62 & 34.50 & 32.70 & 27.69 & 21.47 & 33.00 & -11.22 & -25.38\% \\
		GGD$_{cr}^{se}$ & 66.04 & 52.16 & 44.48 & 32.35 & 31.81 & 45.37 & 1.15   & 2.60\%   \\ \bottomrule
	\end{tabular}
\end{table*}

\subsection{UpDn}
We use the publicly available reimplementation of UpDn\footnote{https://github.com/hengyuan-hu/bottom-up-attention-vqa}~\cite{2018bottomup} for our baseline architecture, data preprocess and optimization in the VQA task.

{\bf Image Encoder}. Following the popular bottom-up attention mechanism~\cite{2018bottomup}, we use a Faster R-CNN~\cite{2017frcnn} based framework to extract visual features. We select the top-36 region proposals for each image $\mathbf{v} \in \mathbb{R}^{36\times 2048}$.

{\bf Question Encoder}. Each word is first initialized by 300-dim GloVe word embeddings~\cite{2014glove}, then fed into a GRU with 1024-d hidden vector. The question representation is the last state of GRU $h_T \in \mathbb{R}^{1024}$.

{\bf Multi-modal Fusion}. We use traditional linear attention between $h_T$ and $\mathbf{v}$ for visual representation. and the final representation for classification is the Hadamard product of vision and question representation.

{\bf Question-only Classifier}. The question-only classifier is implemented as two fully-connected layers with ReLU activations. The input question representation is shared with that in VQA base model.

{\bf Question types}. We use 65 question types annotated in VQA v2 and VQA-CP, according to the first few words of the question (e.g., ``What color is"). To save the training time, we simply use statistic answer distribution conditioned by question type in the train set as the prediction of distribution bias.

{\bf Optimization}. Following UpDn~\cite{2018bottomup}, all the experiments are conducted with the Adamax optimizer for 20 epochs with learning rate initialized as 0.001. We train all models on a single RTX 3090 GUP with PyTorch 1.7~\cite{2019pytorch} and batch size 512.

{\bf Data Preprocessing}. Following previous works, we filter the answers that appear less than 9 times in the train set. For each instance with 10 annotated answers, we set the scores for labels that appear 1/2/3 times as 0.3/0.6/0.9, more than 3 times as 1.0.

\section{Ablations of Base model}
GGD is agnostic for choice of the base model. In this section we provide extra experiments on Biased MNIST~\cite{2020rebias} and VQA-CP v2~\cite{2018vqacp}.

For Biased MNIST, we do experiments on SimpleNet-7k following \cite{2020rebias}. SimpleNet-7k has the same architecture with SimpleNet-1k introduced in Section~\ref{simnet} but with convolution kernel size $7\times 7$. SimpleNet-1k is chosen for the biased model for all experiments in TABLE~\ref{tab:7k}. 

For VQA-CP, We do experiments on other base models BAN~\cite{2018BAN} and S-MRL~\cite{2019rubi}. The models are re-implemented based on officially released codes. For BAN, we set the number of Bilinear Attention blocks as 3. We choose the last bi-linear attention map of BAN and sum up along the question axis, which is referred to as the object attention for CGR and CGW. Although Accuracy of our reproduced S-MRL is a litter lower than that in \cite{2019rubi}, GGD$^{dq}$ can improve the Accuracy over 10\% and surpass most of the existing methods. As shown in the TABLE~\ref{tab:ban}, GGD is a model-agnostic de-bias method, which can improve all three base models UpDn~\cite{2018bottomup}, S-MRL\cite{2019rubi} and BAN~\cite{2018BAN} by a large margin.

\section{Visualization on Biased-MNIST}
In this section we provide the saliency map visualizations of vanilla ResNet-18, SimpleNet-1k, and the GGD base model with grad-CAM~\cite{2017gradcam}. All experiments are demonstrate on Biased-MNIST with $\rho_{\text{train}}=0.999$. 

As shown in Fig.~\ref{fig:cam}, the vanilla ResNet-18 mainly focus on the background parts, since the background colour is highly correlated with the labels in the train stage.
SimpleNet-1k provides an uniform saliency map due to the small perceptive field.
In contrast, the visualization of GGD focus on the middle of the image, which means it capture more information about the digit number.

\begin{figure}[t]		
		\begin{center}
			\includegraphics[width=0.9\linewidth]{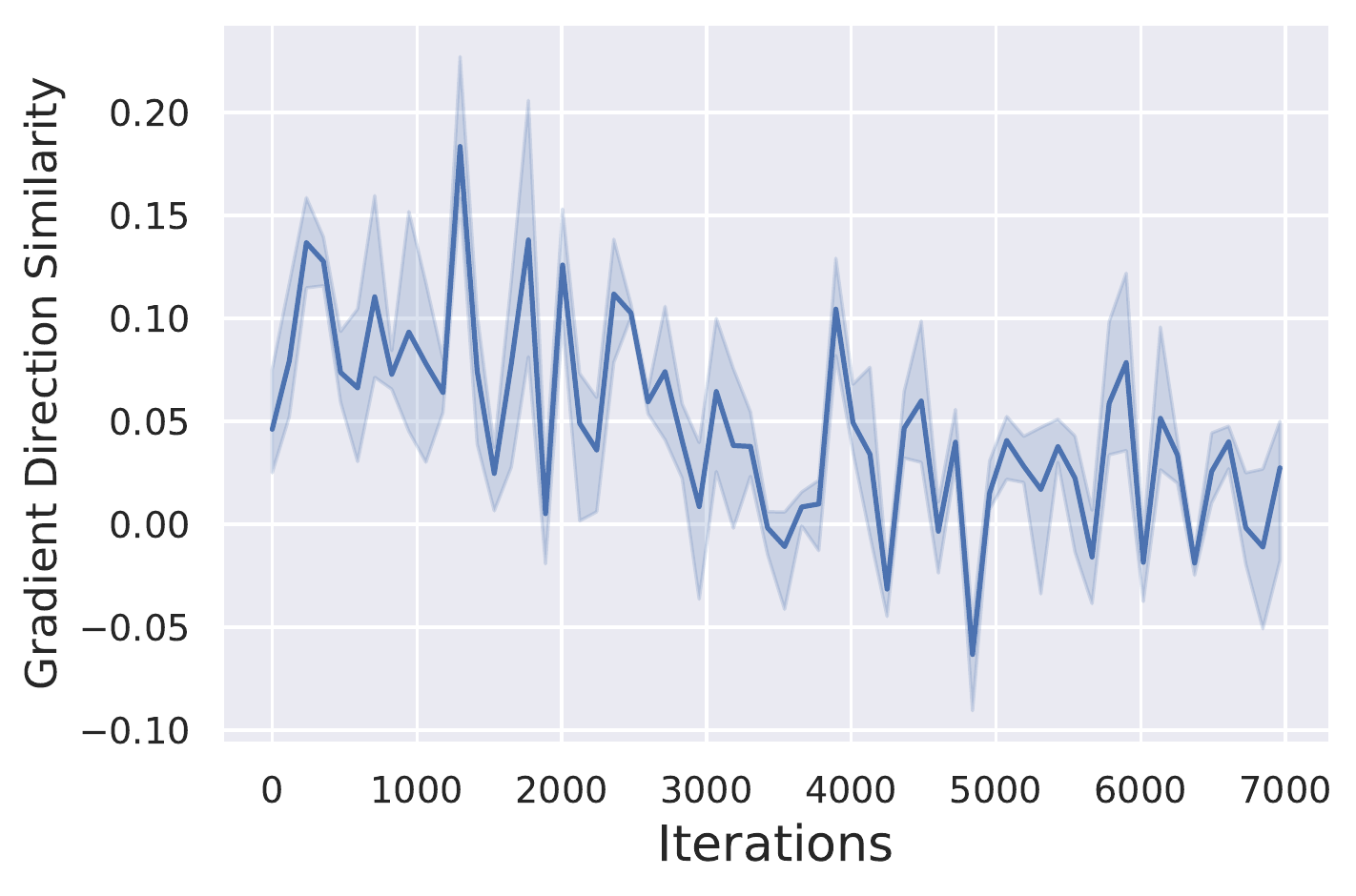}
		\end{center}
	\vspace{-2em}
		\caption{The gradient similarity between baseline model and GGD$_{cr}$ base model.}
		\label{fig:gradient}
		\vspace{-1em}
\end{figure}

\section{Gradient Comparison}

To further compare the training process between GGD and the baseline, we provide the cosine similarity of the direction of the gradient for the feature before the classifier:
\begin{equation}\label{cos_sim}
s = \frac{\mathbf{g}_{base} \mathbf{g}_{ggd}} { ||\mathbf{g}_{base}|| ||\mathbf{g}_{ggd}||},
\end{equation}
where $||.||$ indicates $L_2$ norm of a vector.
The gradients are also accumulated every 118 iterations and averaged with four repeated experiments with different random seeds. To guarantee that the baseline and the GGD base model are given the same initialization and trained with the same mini-batch of data at each iteration, this experiment is conducted on the Self-Ensemble GGD$_{gs}^{se}$. We plot the cosine similarity versus training iteration in Fig.\ref{fig:gradient}. It shows that the optimization direction with the same training data is extremely different between GGD and baseline, $s$ is no more than 0.15 during training. This indicates that GGD learns a different feature compared with the baseline, which can better classify the hard examples in spite of the spurious correlations.

\section{Experiments on Long-tailed Image Classification}
{\bf Dataset}. 
The original CIFAR-10 (CIFAR-100) dataset contains 50,000 training images and 10,000 test images of size 32x32 uniformly falling into 10 (100) classes~\cite{2009cifar}. Cui et al.~\cite{cui2019class} created long-tailed versions by randomly removing training examples. In particular, $\mu = n_{t} / n_h$ controls the imbalance factor of the dataset, where $n_h$ is the number of examples in the head class and $n_t$ is the number of examples from the tailed class. By varying $\mu$, we arrive at 5 training sets, respectively, with the imbalance factors of 1, 0.2, 0.1, 0.02, and 0.01, where $\mu=1$ corresponds to the
original datasets. 

{\bf Biased Models}.
We test two settings of biased models in the experiments. The first biased model directly uses the statistical distribution of the training set as biased predictions, noted as GGD$^{d}$.
The second is the self-ensemble GGD$^{se}$

{\bf Experimental Setups}. 
We choose ResNet-32 as our baseline model. All experiments are conducted with the Adam optimizer for 250 epochs and the learning rate is initialized as 0.01. 

{\bf Experimental Results}.
With the prior knowledge of the unbalanced distribution, both GGD$_{gs}$ and GGD$_{cr}$ can promisingly improve the performance of the long-tailed training data. However, in self-ensemble version without prior knowledge, GGD$_{gs}$ will fail to estimate the biases of data. The baseline model itself cannot correctly reflect the data distribution.
On the other hand, GGD$_{cr}$ is much more robust. It at least keep the performance of the base model at all imbalanced levels without prior knowledge, even when the imbalanced factor $\mu=1$.

%% file: bare_jrnl_compsoc.bbl
\begin{thebibliography}{10}
\providecommand{\url}[1]{#1}
\csname url@samestyle\endcsname
\providecommand{\newblock}{\relax}
\providecommand{\bibinfo}[2]{#2}
\providecommand{\BIBentrySTDinterwordspacing}{\spaceskip=0pt\relax}
\providecommand{\BIBentryALTinterwordstretchfactor}{4}
\providecommand{\BIBentryALTinterwordspacing}{\spaceskip=\fontdimen2\font plus
\BIBentryALTinterwordstretchfactor\fontdimen3\font minus
  \fontdimen4\font\relax}
\providecommand{\BIBforeignlanguage}[2]{{%
\expandafter\ifx\csname l@#1\endcsname\relax
\typeout{** WARNING: IEEEtran.bst: No hyphenation pattern has been}%
\typeout{** loaded for the language `#1'. Using the pattern for}%
\typeout{** the default language instead.}%
\else
\language=\csname l@#1\endcsname
\fi
#2}}
\providecommand{\BIBdecl}{\relax}
\BIBdecl

\bibitem{9356220}
Z.~Qi, S.~Wang, C.~Su, L.~Su, Q.~Huang, and Q.~Tian, ``Self-regulated learning
  for egocentric video activity anticipation,'' \emph{IEEE Transactions on
  Pattern Analysis and Machine Intelligence}, 2021.

\bibitem{2011unbiased}
A.~Torralba and A.~Efros, ``Unbiased look at dataset bias,'' in
  \emph{Proceedings of the 2011 IEEE Conference on Computer Vision and Pattern
  Recognition}, 2011, pp. 1521--1528.

\bibitem{2017generalization}
C.~Zhang, S.~Bengio, M.~Hardt, B.~Recht, and O.~Vinyals, ``Understanding deep
  learning requires rethinking generalization,'' in \emph{Proceedings of the
  International Conference on Learning Representations}, 2017.

\bibitem{2020shortcut}
R.~Geirhos, J.-H. Jacobsen, C.~Michaelis, R.~Zemel, W.~Brendel, M.~Bethge, and
  F.~A. Wichmann, ``Shortcut learning in deep neural networks,'' \emph{Nature
  Machine Intelligence}, vol.~2, no.~11, pp. 665--673, 2020.

\bibitem{2020overpara}
S.~Sagawa, A.~Raghunathan, P.~W. Koh, and P.~Liang, ``An investigation of why
  overparameterization exacerbates spurious correlations,'' in
  \emph{International Conference on Machine Learning}.\hskip 1em plus 0.5em
  minus 0.4em\relax PMLR, 2020, pp. 8346--8356.

\bibitem{2016squad}
P.~Rajpurkar, J.~Zhang, K.~Lopyrev, and P.~Liang, ``Squad: 100,000+ questions
  for machine comprehension of text,'' in \emph{Proceedings of the 2016
  Conference on Empirical Methods in Natural Language Processing}, 2016, pp.
  2383--2392.

\bibitem{2017adqa}
R.~Jia and P.~Liang, ``Adversarial examples for evaluating reading
  comprehension systems,'' in \emph{Proceedings of the 2017 Conference on
  Empirical Methods in Natural Language Processing}, 2017, pp. 2021--2031.

\bibitem{2019-compositional}
\BIBentryALTinterwordspacing
S.~Min, E.~Wallace, S.~Singh, M.~Gardner, H.~Hajishirzi, and L.~Zettlemoyer,
  ``Compositional questions do not necessitate multi-hop reasoning,'' in
  \emph{Proceedings of the 57th Annual Meeting of the Association for
  Computational Linguistics}.\hskip 1em plus 0.5em minus 0.4em\relax Florence,
  Italy: Association for Computational Linguistics, Jul. 2019, pp. 4249--4257.
  [Online]. Available: \url{https://www.aclweb.org/anthology/P19-1416}
\BIBentrySTDinterwordspacing

\bibitem{2017mfh}
Y.~Goyal, T.~Khot, D.~Summers-Stay, D.~Batra, and D.~Parikh, ``Making the v in
  vqa matter: Elevating the role of image understanding in visual question
  answering,'' in \emph{Proceedings of the IEEE Conference on Computer Vision
  and Pattern Recognition}, 2017, pp. 6904--6913.

\bibitem{2017analysis}
K.~Kafle and C.~Kanan, ``An analysis of visual question answering algorithms,''
  in \emph{Proceedings of the IEEE International Conference on Computer
  Vision}, 2017, pp. 1965--1973.

\bibitem{2018vqacp}
A.~Agrawal, D.~Batra, D.~Parikh, and A.~Kembhavi, ``Don't just assume; look and
  answer: Overcoming priors for visual question answering,'' in
  \emph{Proceedings of the IEEE Conference on Computer Vision and Pattern
  Recognition}, 2018, pp. 4971--4980.

\bibitem{2020rebias}
H.~Bahng, S.~Chun, S.~Yun, J.~Choo, and S.~J. Oh, ``Learning de-biased
  representations with biased representations,'' in \emph{International
  Conference on Machine Learning}.\hskip 1em plus 0.5em minus 0.4em\relax PMLR,
  2020, pp. 528--539.

\bibitem{2015CelebA}
Z.~Liu, P.~Luo, X.~Wang, and X.~Tang, ``Deep learning face attributes in the
  wild,'' in \emph{Proceedings of the IEEE international conference on computer
  vision}, 2015, pp. 3730--3738.

\bibitem{2021NICO}
Y.~He, Z.~Shen, and P.~Cui, ``Towards non-iid image classification: A dataset
  and baselines,'' \emph{Pattern Recognition}, vol. 110, p. 107383, 2021.

\bibitem{2019LDAM}
K.~Cao, C.~Wei, A.~Gaidon, N.~Arechiga, and T.~Ma, ``Learning imbalanced
  datasets with label-distribution-aware margin loss,'' in \emph{Advances in
  Neural Information Processing Systems}, 2019.

\bibitem{2019class}
Y.~Cui, M.~Jia, T.-Y. Lin, Y.~Song, and S.~Belongie, ``Class-balanced loss
  based on effective number of samples,'' in \emph{Proceedings of the IEEE/CVF
  Conference on Computer Vision and Pattern Recognition}, 2019, pp. 9268--9277.

\bibitem{2019don}
C.~Clark, M.~Yatskar, and L.~Zettlemoyer, ``Don’t take the easy way out:
  Ensemble based methods for avoiding known dataset biases,'' in
  \emph{Proceedings of the 2019 Conference on Empirical Methods in Natural
  Language Processing and the 9th International Joint Conference on Natural
  Language Processing}, 2019, pp. 4060--4073.

\bibitem{2019rubi}
R.~Cadene, C.~Dancette, M.~Cord, D.~Parikh \emph{et~al.}, ``Rubi: Reducing
  unimodal biases for visual question answering,'' in \emph{Advances in neural
  information processing systems}, 2019, pp. 841--852.

\bibitem{2019GDRO}
S.~Sagawa, P.~W. Koh, T.~B. Hashimoto, and P.~Liang, ``Distributionally robust
  neural networks for group shifts: On the importance of regularization for
  worst-case generalization,'' in \emph{Proceedings of the International
  Conference on Learning Representations}, 2019.

\bibitem{2019lnl}
B.~Kim, H.~Kim, K.~Kim, S.~Kim, and J.~Kim, ``Learning not to learn: Training
  deep neural networks with biased data,'' in \emph{Proceedings of the IEEE/CVF
  Conference on Computer Vision and Pattern Recognition}, 2019, pp. 9012--9020.

\bibitem{2021end}
E.~Tartaglione, C.~A. Barbano, and M.~Grangetto, ``End: Entangling and
  disentangling deep representations for bias correction,'' in
  \emph{Proceedings of the IEEE/CVF Conference on Computer Vision and Pattern
  Recognition (CVPR)}, June 2021, pp. 13\,508--13\,517.

\bibitem{2020counterfactual}
L.~Chen, X.~Yan, J.~Xiao, H.~Zhang, S.~Pu, and Y.~Zhuang, ``Counterfactual
  samples synthesizing for robust visual question answering,'' in
  \emph{Proceedings of the IEEE/CVF Conference on Computer Vision and Pattern
  Recognition}, 2020, pp. 10\,800--10\,809.

\bibitem{2020counterfactualgs}
D.~Teney, E.~Abbasnedjad, and A.~van~den Hengel, ``Learning what makes a
  difference from counterfactual examples and gradient supervision,'' in
  \emph{Proceedings of the European conference on computer vision}, 2020, pp.
  580--599.

\bibitem{2021investigation}
R.~Shrestha, K.~Kafle, and C.~Kanan, ``An investigation of critical issues in
  bias mitigation techniques,'' \emph{arXiv preprint arXiv:2104.00170}, 2021.

\bibitem{2020lff}
J.~H. Nam, H.~Cha, S.~Ahn, J.~Lee, and J.~Shin, ``Learning from failure:
  De-biasing classifier from biased classifier,'' in \emph{34th Conference on
  Neural Information Processing Systems (NeurIPS) 2020}, vol.~33, 2020, pp.
  20\,673--20\,684.

\bibitem{2020GS}
M.~Pezeshki, S.-O. Kaba, Y.~Bengio, A.~Courville, D.~Precup, and G.~Lajoie,
  ``Gradient starvation: A learning proclivity in neural networks,''
  \emph{arXiv preprint arXiv:2011.09468}, 2020.

\bibitem{2020rsc}
Z.~Huang, H.~Wang, E.~P. Xing, and D.~Huang, ``Self-challenging improves
  cross-domain generalization,'' in \emph{Proceedings of the European
  Conference on Computer Vision}, 2020.

\bibitem{2021biaswap}
E.~Kim, J.~Lee, and J.~Choo, ``Biaswap: Removing dataset bias with
  bias-tailored swapping augmentation,'' in \emph{Proceedings of the IEEE/CVF
  International Conference on Computer Vision (ICCV)}, October 2021, pp.
  14\,992--15\,001.

\bibitem{han2021greedy}
X.~Han, S.~Wang, C.~Su, Q.~Huang, and Q.~Tian, ``Greedy gradient ensemble for
  robust visual question answering,'' in \emph{Proceedings of the IEEE/CVF
  International Conference on Computer Vision}, 2021, pp. 1584--1593.

\bibitem{2021ebd}
R.~Xiong, Y.~Chen, L.~Pang, X.~Cheng, Z.-M. Ma, and Y.~Lan, ``Uncertainty
  calibration for ensemble-based debiasing methods,'' \emph{Advances in Neural
  Information Processing Systems}, vol.~34, 2021.

\bibitem{2021introspective}
Y.~Niu and H.~Zhang, ``Introspective distillation for robust question
  answering,'' in \emph{Advances in Neural Information Processing Systems},
  vol.~34, 2021.

\bibitem{kumar2022calibrated}
A.~Kumar, T.~Ma, P.~Liang, and A.~Raghunathan, ``Calibrated ensembles can
  mitigate accuracy tradeoffs under distribution shift,'' in \emph{The 38th
  Conference on Uncertainty in Artificial Intelligence}, 2022.

\bibitem{2009curriculum}
Y.~Bengio, J.~Louradour, R.~Collobert, and J.~Weston, ``Curriculum learning,''
  in \emph{Proceedings of the 26th annual international conference on machine
  learning}, 2009, pp. 41--48.

\bibitem{2009imagenet}
J.~Deng, W.~Dong, R.~Socher, L.-J. Li, K.~Li, and L.~Fei-Fei, ``Imagenet: A
  large-scale hierarchical image database,'' in \emph{2009 IEEE conference on
  computer vision and pattern recognition}.\hskip 1em plus 0.5em minus
  0.4em\relax Ieee, 2009, pp. 248--255.

\bibitem{2009cifar}
A.~Krizhevsky, G.~Hinton \emph{et~al.}, ``Learning multiple layers of features
  from tiny images,'' 2009.

\bibitem{xie2020n}
S.~M. Xie, A.~Kumar, R.~Jones, F.~Khani, T.~Ma, and P.~Liang, ``In-n-out:
  Pre-training and self-training using auxiliary information for
  out-of-distribution robustness,'' in \emph{International Conference on
  Learning Representations}, 2021.

\bibitem{kumar2021fine}
A.~Kumar, A.~Raghunathan, R.~M. Jones, T.~Ma, and P.~Liang, ``Fine-tuning can
  distort pretrained features and underperform out-of-distribution,'' in
  \emph{International Conference on Learning Representations}, 2021.

\bibitem{2017ad_squad}
\BIBentryALTinterwordspacing
R.~Jia and P.~Liang, ``Adversarial examples for evaluating reading
  comprehension systems,'' in \emph{Proceedings of the 2017 Conference on
  Empirical Methods in Natural Language Processing}.\hskip 1em plus 0.5em minus
  0.4em\relax Copenhagen, Denmark: Association for Computational Linguistics,
  Sep. 2017, pp. 2021--2031. [Online]. Available:
  \url{https://aclanthology.org/D17-1215}
\BIBentrySTDinterwordspacing

\bibitem{2021gqa-ood}
C.~Kervadec, G.~Antipov, M.~Baccouche, and C.~Wolf, ``Roses are red, violets
  are blue... but should vqa expect them to?'' in \emph{Proceedings of the
  IEEE/CVF Conference on Computer Vision and Pattern Recognition}, 2021, pp.
  2776--2785.

\bibitem{cui2019class}
Y.~Cui, M.~Jia, T.-Y. Lin, Y.~Song, and S.~Belongie, ``Class-balanced loss
  based on effective number of samples,'' in \emph{Proceedings of the IEEE/CVF
  conference on computer vision and pattern recognition}, 2019, pp. 9268--9277.

\bibitem{2019coqa}
S.~Reddy, D.~Chen, and C.~D. Manning, ``Coqa: A conversational question
  answering challenge,'' \emph{Transactions of the Association for
  Computational Linguistics}, vol.~7, pp. 249--266, 2019.

\bibitem{2018quac}
\BIBentryALTinterwordspacing
E.~Choi, H.~He, M.~Iyyer, M.~Yatskar, W.-t. Yih, Y.~Choi, P.~Liang, and
  L.~Zettlemoyer, ``{Q}u{AC}: Question answering in context,'' in
  \emph{Proceedings of the 2018 Conference on Empirical Methods in Natural
  Language Processing}.\hskip 1em plus 0.5em minus 0.4em\relax Brussels,
  Belgium: Association for Computational Linguistics, Oct.-Nov. 2018, pp.
  2174--2184. [Online]. Available: \url{https://aclanthology.org/D18-1241}
\BIBentrySTDinterwordspacing

\bibitem{2016yin}
P.~Zhang, Y.~Goyal, D.~Summers-Stay, D.~Batra, and D.~Parikh, ``Yin and yang:
  Balancing and answering binary visual questions,'' in \emph{Proceedings of
  the IEEE Conference on Computer Vision and Pattern Recognition}, 2016, pp.
  5014--5022.

\bibitem{2015vqa}
S.~Antol, A.~Agrawal, J.~Lu, M.~Mitchell, D.~Batra, C.~Lawrence~Zitnick, and
  D.~Parikh, ``Vqa: Visual question answering,'' in \emph{Proceedings of the
  IEEE international conference on computer vision}, 2015, pp. 2425--2433.

\bibitem{2021towards}
Z.~Shen, J.~Liu, Y.~He, X.~Zhang, R.~Xu, H.~Yu, and P.~Cui, ``Towards
  out-of-distribution generalization: A survey,'' \emph{arXiv preprint
  arXiv:2108.13624}, 2021.

\bibitem{2018overcoming}
S.~Ramakrishnan, A.~Agrawal, and S.~Lee, ``Overcoming language priors in visual
  question answering with adversarial regularization,'' in \emph{Advances in
  Neural Information Processing Systems}, 2018, pp. 1541--1551.

\bibitem{2020end}
R.~K. Mahabadi, Y.~Belinkov, and J.~Henderson, ``End-to-end bias mitigation by
  modelling biases in corpora,'' in \emph{Proceedings of the 58th Annual
  Meeting of the Association for Computational Linguistics}, 2020.

\bibitem{2005hsic}
A.~Gretton, O.~Bousquet, A.~Smola, and B.~Sch{\"o}lkopf, ``Measuring
  statistical dependence with hilbert-schmidt norms,'' in \emph{International
  conference on algorithmic learning theory}.\hskip 1em plus 0.5em minus
  0.4em\relax Springer, 2005, pp. 63--77.

\bibitem{2021A-INLP}
P.~P. Liang, C.~Wu, L.-P. Morency, and R.~Salakhutdinov, ``Towards
  understanding and mitigating social biases in language models,'' in
  \emph{International Conference on Machine Learning}.\hskip 1em plus 0.5em
  minus 0.4em\relax PMLR, 2021, pp. 6565--6576.

\bibitem{2019hex}
H.~Wang, Z.~He, Z.~C. Lipton, and E.~P. Xing, ``Learning robust representations
  by projecting superficial statistics out,'' in \emph{Proceedings of the
  International Conference on Learning Representations}, 2019.

\bibitem{2020mfe}
I.~Gat, I.~Schwartz, A.~Schwing, and T.~Hazan, ``Removing bias in multi-modal
  classifiers: Regularization by maximizing functional entropies,''
  \emph{Advances in Neural Information Processing Systems}, vol.~33, 2020.

\bibitem{2021CSAD}
W.~Zhu, H.~Zheng, H.~Liao, W.~Li, and J.~Luo, ``Learning bias-invariant
  representation by cross-sample mutual information minimization,'' in
  \emph{Proceedings of the IEEE/CVF International Conference on Computer Vision
  (ICCV)}, October 2021, pp. 15\,002--15\,012.

\bibitem{2021bcbb}
Y.~Hong and E.~Yang, ``Unbiased classification through bias-contrastive and
  bias-balanced learning,'' \emph{Advances in Neural Information Processing
  Systems}, vol.~34, 2021.

\bibitem{2020stable}
Z.~Shen, P.~Cui, T.~Zhang, and K.~Kunag, ``Stable learning via sample
  reweighting,'' in \emph{Proceedings of the AAAI Conference on Artificial
  Intelligence}, vol.~34, no.~04, 2020, pp. 5692--5699.

\bibitem{kuang2020stable}
K.~Kuang, R.~Xiong, P.~Cui, S.~Athey, and B.~Li, ``Stable prediction with model
  misspecification and agnostic distribution shift,'' in \emph{Proceedings of
  the AAAI Conference on Artificial Intelligence}, vol.~34, no.~04, 2020, pp.
  4485--4492.

\bibitem{2021lfm}
V.~Sanh, T.~Wolf, Y.~Belinkov, and A.~M. Rush, ``Learning from others'
  mistakes: Avoiding dataset biases without modeling them,'' in
  \emph{Proceedings of the International Conference on Learning
  Representations}, 2021.

\bibitem{2020nlu}
P.~A. Utama, N.~S. Moosavi, and I.~Gurevych, ``Towards debiasing nlu models
  from unknown biases,'' in \emph{Proceedings of the 2020 Conference on
  Empirical Methods in Natural Language Processing (EMNLP)}, 2020.

\bibitem{2020mce}
C.~Clark, M.~Yatskar, and L.~Zettlemoyer, ``Learning to model and ignore
  dataset bias with mixed capacity ensembles,'' \emph{arXiv preprint
  arXiv:2011.03856}, 2020.

\bibitem{2021stablenet}
X.~Zhang, P.~Cui, R.~Xu, L.~Zhou, Y.~He, and Z.~Shen, ``Deep stable learning
  for out-of-distribution generalization,'' in \emph{Proceedings of the
  IEEE/CVF Conference on Computer Vision and Pattern Recognition}, 2021, pp.
  5372--5382.

\bibitem{blanchard2021domain}
G.~Blanchard, A.~A. Deshmukh, U.~Dogan, G.~Lee, and C.~Scott, ``Domain
  generalization by marginal transfer learning,'' \emph{Journal of Machine
  Learning Research}, vol.~22, pp. 1--55, 2021.

\bibitem{christiansen2021causal}
R.~Christiansen, N.~Pfister, M.~E. Jakobsen, N.~Gnecco, and J.~Peters, ``A
  causal framework for distribution generalization,'' \emph{IEEE Transactions
  on Pattern Analysis and Machine Intelligence}, 2021.

\bibitem{2018bottomup}
P.~Anderson, X.~He, C.~Buehler, D.~Teney, M.~Johnson, S.~Gould, and L.~Zhang,
  ``Bottom-up and top-down attention for image captioning and visual question
  answering,'' in \emph{Proceedings of the IEEE Conference on Computer Vision
  and Pattern Recognition}, 2018, pp. 6077--6086.

\bibitem{2000gradient}
L.~Mason, J.~Baxter, P.~L. Bartlett, and M.~R. Frean, ``Boosting algorithms as
  gradient descent,'' in \emph{Advances in neural information processing
  systems}, 2000, pp. 512--518.

\bibitem{2021disentangling}
A.~S. Rawat, A.~K. Menon, W.~Jitkrittum, S.~Jayasumana, F.~Yu, S.~Reddi, and
  S.~Kumar, ``Disentangling sampling and labeling bias for learning in
  large-output spaces,'' in \emph{Proceedings of the 38th International
  Conference on Machine Learning}, ser. Proceedings of Machine Learning
  Research, M.~Meila and T.~Zhang, Eds., vol. 139.\hskip 1em plus 0.5em minus
  0.4em\relax PMLR, 18--24 Jul 2021, pp. 8890--8901.

\bibitem{2020decoupling}
B.~Kang, S.~Xie, M.~Rohrbach, Z.~Yan, A.~Gordo, J.~Feng, and Y.~Kalantidis,
  ``Decoupling representation and classifier for long-tailed recognition,''
  \emph{Proceedings of the International Conference on Learning
  Representations}, 2020.

\bibitem{2021disalign}
S.~Zhang, Z.~Li, S.~Yan, X.~He, and J.~Sun, ``Distribution alignment: A unified
  framework for long-tail visual recognition,'' in \emph{Proceedings of the
  IEEE/CVF Conference on Computer Vision and Pattern Recognition}, 2021, pp.
  2361--2370.

\bibitem{pagliardini2022agree}
M.~Pagliardini, M.~Jaggi, F.~Fleuret, and S.~P. Karimireddy, ``Agree to
  disagree: Diversity through disagreement for better transferability,''
  \emph{arXiv preprint arXiv:2202.04414}, 2022.

\bibitem{1995boosting}
Y.~Freund, ``Boosting a weak learning algorithm by majority,''
  \emph{Information and computation}, vol. 121, no.~2, pp. 256--285, 1995.

\bibitem{schapire1990strength}
R.~E. Schapire, ``The strength of weak learnability,'' \emph{Machine learning},
  vol.~5, no.~2, pp. 197--227, 1990.

\bibitem{2006some}
P.~J. Bickel, Y.~Ritov, and A.~Zakai, ``Some theory for generalized boosting
  algorithms,'' \emph{Journal of Machine Learning Research}, vol.~7, no. May,
  pp. 705--732, 2006.

\bibitem{2021calibration}
Y.~Wald, A.~Feder, D.~Greenfeld, and U.~Shalit, ``On calibration and
  out-of-domain generalization,'' in \emph{Advances in neural information
  processing systems}, vol.~34, 2021, pp. 2215--2227.

\bibitem{2022improved}
M.~Yi, R.~Wang, J.~Sun, Z.~Li, and Z.-M. Ma, ``Improved ood generalization via
  conditional invariant regularizer,'' \emph{arXiv preprint arXiv:2207.06687},
  2022.

\bibitem{hendrycks2019using}
D.~Hendrycks, K.~Lee, and M.~Mazeika, ``Using pre-training can improve model
  robustness and uncertainty,'' in \emph{International Conference on Machine
  Learning}.\hskip 1em plus 0.5em minus 0.4em\relax PMLR, 2019, pp. 2712--2721.

\bibitem{1998mnist}
Y.~LeCun, L.~Bottou, Y.~Bengio, and P.~Haffner, ``Gradient-based learning
  applied to document recognition,'' \emph{Proceedings of the IEEE}, vol.~86,
  no.~11, pp. 2278--2324, 1998.

\bibitem{2019hint}
R.~R. Selvaraju, S.~Lee, Y.~Shen, H.~Jin, S.~Ghosh, L.~Heck, D.~Batra, and
  D.~Parikh, ``Taking a hint: Leveraging explanations to make vision and
  language models more grounded,'' in \emph{Proceedings of the IEEE
  International Conference on Computer Vision}, 2019, pp. 2591--2600.

\bibitem{2019SCR}
J.~Wu and R.~Mooney, ``Self-critical reasoning for robust visual question
  answering,'' in \emph{Advances in Neural Information Processing Systems},
  2019, pp. 8604--8614.

\bibitem{2020DLP}
C.~Jing, Y.~Wu, X.~Zhang, Y.~Jia, and Q.~Wu, ``Overcoming language priors in
  vqa via decomposed linguistic representations.'' in \emph{AAAI}, 2020, pp.
  11\,181--11\,188.

\bibitem{2020reducing}
G.~KV and A.~Mittal, ``Reducing language biases in visual question answering
  with visually-grounded question encoder,'' \emph{arXiv preprint
  arXiv:2007.06198}, 2020.

\bibitem{2020cf-vqa}
Y.~Niu, K.~Tang, H.~Zhang, Z.~Lu, X.-S. Hua, and J.-R. Wen, ``Counterfactual
  vqa: A cause-effect look at language bias,'' \emph{arXiv preprint
  arXiv:2006.04315}, 2020.

\bibitem{2021SAR}
Q.~Si, Z.~Lin, M.~Zheng, P.~Fu, and W.~Wang, ``Check it again: Progressive
  visual question answering via visual entailment,'' \emph{arXiv preprint
  arXiv:2106.04605}, 2021.

\bibitem{2017bidaf}
M.~Seo, A.~Kembhavi, A.~Farhadi, and H.~Hajishirzi, ``Bidirectional attention
  flow for machine comprehension,'' in \emph{Proceedings of the International
  Conference on Learning Representations}, 2017.

\bibitem{gao2019multi}
P.~Gao, H.~You, Z.~Zhang, X.~Wang, and H.~Li, ``Multi-modality latent
  interaction network for visual question answering,'' \emph{arXiv preprint
  arXiv:1908.04289}, 2019.

\bibitem{2017n2nmn}
R.~Hu, J.~Andreas, M.~Rohrbach, T.~Darrell, and K.~Saenko, ``Learning to
  reason: End-to-end module networks for visual question answering,'' in
  \emph{Proceedings of the IEEE International Conference on Computer Vision},
  2017.

\bibitem{2019lcgn}
R.~Hu, A.~Rohrbach, T.~Darrell, and K.~Saenko, ``Language-conditioned graph
  networks for relational reasoning,'' in \emph{Proceedings of the IEEE
  International Conference on Computer Vision}, 2019, pp. 10\,294--10\,303.

\bibitem{2019nscl}
J.~Mao, C.~Gan, P.~Kohli, J.~B. Tenenbaum, and J.~Wu, ``The neuro-symbolic
  concept learner: Interpreting scenes, words, and sentences from natural
  supervision,'' \emph{arXiv preprint arXiv:1904.12584}, 2019.

\bibitem{han2020interpretable}
X.~Han, S.~Wang, C.~Su, W.~Zhang, Q.~Huang, and Q.~Tian, ``Interpretable visual
  reasoning via probabilistic formulation under natural supervision,'' in
  \emph{European Conference on Computer Vision}.\hskip 1em plus 0.5em minus
  0.4em\relax Springer, 2020, pp. 553--570.

\bibitem{2017clevr}
J.~Johnson, B.~Hariharan, L.~van~der Maaten, L.~Fei-Fei, C.~Lawrence~Zitnick,
  and R.~Girshick, ``Clevr: A diagnostic dataset for compositional language and
  elementary visual reasoning,'' in \emph{Proceedings of the IEEE Conference on
  Computer Vision and Pattern Recognition}, 2017, pp. 2901--2910.

\bibitem{2019gqa}
D.~A. Hudson and C.~D. Manning, ``Gqa: A new dataset for real-world visual
  reasoning and compositional question answering,'' in \emph{Proceedings of the
  IEEE Conference on Computer Vision and Pattern Recognition}, 2019, pp.
  6700--6709.

\bibitem{2017frcnn}
S.~Ren, K.~He, R.~Girshick, and J.~Sun, ``Faster r-cnn: Towards real-time
  object detection with region proposal networks,'' \emph{IEEE Transactions on
  Pattern Analysis and Machine Intelligence}, vol.~39, no.~6, pp. 1137--1149,
  2017.

\bibitem{teney2020value}
D.~Teney, E.~Abbasnejad, K.~Kafle, R.~Shrestha, C.~Kanan, and A.~Van
  Den~Hengel, ``On the value of out-of-distribution testing: An example of
  goodhart's law,'' in \emph{Advances in Neural Information Processing
  Systems}, vol.~33, 2020, pp. 407--417.

\bibitem{2015bn}
S.~Ioffe and C.~Szegedy, ``Batch normalization: Accelerating deep network
  training by reducing internal covariate shift,'' in \emph{International
  conference on machine learning}.\hskip 1em plus 0.5em minus 0.4em\relax PMLR,
  2015, pp. 448--456.

\bibitem{2018BAN}
J.-H. Kim, J.~Jun, and B.-T. Zhang, ``Bilinear attention networks,'' in
  \emph{Advances in Neural Information Processing Systems}, 2018, pp.
  1564--1574.

\bibitem{2014glove}
J.~Pennington, R.~Socher, and C.~Manning, ``Glove: Global vectors for word
  representation,'' in \emph{Proceedings of the 2014 conference on empirical
  methods in natural language processing}, 2014, pp. 1532--1543.

\bibitem{2019pytorch}
A.~Paszke, S.~Gross, F.~Massa, A.~Lerer, J.~Bradbury, G.~Chanan, T.~Killeen,
  Z.~Lin, N.~Gimelshein, L.~Antiga \emph{et~al.}, ``Pytorch: An imperative
  style, high-performance deep learning library,'' in \emph{Advances in neural
  information processing systems}, 2019, pp. 8026--8037.

\bibitem{2017gradcam}
R.~R. Selvaraju, M.~Cogswell, A.~Das, R.~Vedantam, D.~Parikh, and D.~Batra,
  ``Grad-cam: Visual explanations from deep networks via gradient-based
  localization,'' in \emph{Proceedings of the IEEE international conference on
  computer vision}, 2017, pp. 618--626.

\end{thebibliography}
